\documentclass[11pt,column]{article}
\usepackage{amsfonts,amssymb,amsmath,mathtools,bm,amsthm}
\usepackage{amssymb,amsmath,amsthm,mathrsfs,ulem,tikz, caption,titlesec}

\title{%Mean Field Theory on 
SGD Distributional Dynamics of Three Layer Neural Networks}
\author{Victor Luo and Yazhen Wang  \\ Department of Statistics, University of Wisconsin-Madison \\ Madison, WI 53706, USA. 
Email: vluo@wisc.edu, yzwang@stat.wisc.edu \\
Glenn Fung \\
American Family Insurance, 6000 American Parkway \\ Madison, WI  53783, USA. Email: gfung@amfam.com 
} 

\usepackage[margin=0.7in]{geometry}

\newtheorem{thm}{Theorem}[section]
\newtheorem{cor}{Corollary}
\newtheorem{lemma}{Lemma}
\newtheorem{proposition}{Proposition}[section]

\newtheorem{assump}{Assumption}

\begin{document}
\maketitle
\begin{abstract}
With the rise of big data analytics, multi-layer neural networks have surfaced as one of the most powerful machine learning methods. However, their theoretical mathematical properties are still not fully understood. Training a neural network requires optimizing a non-convex objective function, typically done using stochastic gradient descent (SGD). In this paper, we seek to extend the mean field results of \cite{MMN2018} from two-layer neural networks with one hidden layer to three-layer neural networks with two hidden layers. We will show that the SGD dynamics is captured by a set of non-linear partial differential equations, and prove that the distributions of weights in the two hidden layers are independent. We will also detail exploratory work done based on simulation and real-world data.
\end{abstract}

\tableofcontents

\section{Introduction}
%https://arxiv.org/pdf/1902.06720.pdf
%file:///Users/victorluo/Desktop/1502.04608v6.pdf

Multi-layer neural networks have been an integral part of statistical machine learning since the 1960s. With advancements in technology and large data capacity, they have surfaced as a ``jack of all trades" in a broad array of learning tasks. In spite of the success multi-layer neural networks 
lack deep theoretical understanding based on rigorous mathematical justifications. Training neural networks requires to solve non-convex high-dimensional optimization problems, common approaches rely on the use of stochastic gradient descent (SGD), and the obtained neural networks
heavily depend on the SGD performance (\cite{Goodfellow2016}, \cite{Luo2020}, \cite{MMN2018},  \cite{Wang2020}). 
It has been shown in \cite{MMN2018} via mean field theory that the SGD distributional dynamics of two-layers neural networks can be described 
by a non-linear partial differential equation (PDE). This paper extends the mean field results of \cite{MMN2018} from two-layer neural networks with one hidden layer to three-layer neural networks with two hidden layers. We will show that the SGD dynamics is captured by a set of non-linear PDEs, and prove that the distributions of weights in the two hidden layers are independent. We will also detail exploratory work done based on simulation and real-world data.

The rest of the paper is organized as follows. Section 2 describes neutral networks and SGD. Section 3 features the theoretical development of 
SGD distributional dynamics by mean field theory. Section 4 provides simulations and applications to illustrate the numerical performance of 
the developed methods. 

\section{Neural networks and SGD}

For this paper, we will focus on the supervised learning setting. 
Assume we are given data points $({\bf x}_i, y_i) \in \mathbb{R}^d \times \mathbb{R}$, indexed by $i \in \mathbb{N}$, which are assumed to be i.i.d. from unknown distribution $\mathbb{P}$ on $\mathbb{R}^d \times \mathbb{R}$. ${\bf x}_i \in \mathbb{R}^d$ is a feature vector and $y_i \in \mathbb{R}$ is a label. Denote $\mathbb{R}^d$ as $\mathbb{R}^{\prescript{}{0} D}$ for simplicity. Our goal is to model the dependence of the label $y_i$ on the feature vector of ${\bf x}_i$ to assign labels to unlabeled examples. In a three layer neural network (two hidden layers), the dependence is modeled as

\begin{align}
\hat{y}({\bf x} ; {\bm \theta})=\frac{1}{\prescript{}{2} N}\sum_{m=1}^{\prescript{}{2} N} \prescript{}{2} \sigma_* (\prescript{}{1} \sigma_* ({\bf x}, \prescript{}{1} {\bm \theta}),\prescript{}{2} {\bm \theta}_m) \label{yhat}
\end{align}
where $\prescript{}{i} N$ is the number of hidden units in the $i$th hidden layer, $\prescript{}{i} \sigma_*:\mathbb{R}^{\prescript{}{i-1} D}\times \mathbb{R}^{\prescript{}{i} D} \to \mathbb{R}$ is the activation function for the $i$th layer, and $\prescript{}{i} {\bm \theta} \in \mathbb{R}^{\prescript{}{i} D}$ are parameters in the $i$th layer, collectively denoted as $\prescript{}{i} {\bm \theta}=(\prescript{}{i} {\bm \theta}_1,\dots,\prescript{}{i} {\bm \theta}_{\prescript{}{i} N})$ for $i=1,2$. We will also denote ${\bm \theta}=(\prescript{}{1} {\bm \theta},\prescript{}{2} {\bm \theta})$ as a vector of all the parameters. Typically, $\prescript{}{i} {\bm \theta}_j=(\prescript{}{i} a_j,\prescript{}{i} b_j, \prescript{}{i} {\bf w}_j)$, with

$$\prescript{}{i} \sigma_* (\prescript{}{i-1}{\bf z};\prescript{}{i} {\bm \theta}_j) =\prescript{}{i} a_j *\sigma(\prescript{}{i} {\bf w}_j \circ \prescript{}{i-1}{\bf z}+\prescript{}{i} b_j)$$
for $\prescript{}{i-1}{\bf z}\in \mathbb{R}^{\prescript{}{i-1} D}$ and some $\sigma:\mathbb{R}\to\mathbb{R}$, where $\circ$ represents the Hadamard product. In (\ref{yhat}), $\prescript{}{1}\sigma_*({\bf x},\prescript{}{1}{\bm \theta})$ represents the vector of size $\prescript{}{1}D$ of values produced by the activation function $\prescript{}{1}\sigma_*$ from the input layer towards the first hidden layer. Typically, the parameters ${\bm \theta}$ are chosen to minimize the risk $R_{\prescript{}{2} N} ({\bm \theta})=\mathbb{E}\{\ell(y, \hat{y}({\bf x};{\bm \theta}))\}$, where $\ell:\mathbb{R} \times \mathbb{R} \to \mathbb{R}$ is a certain loss function. We will focus on the squared loss $\ell(y,\hat{y})=(y-\hat{y})^2$. 

We can rewrite the population risk $R_{\prescript{}{2} N}({\bm \theta})=\mathbb{E}\{(y-\hat{y}({\bf x};{\bm \theta}))^2\}$ as

\begin{align}
R_{\prescript{}{2} N}({\bm \theta}) &=\mathbb{E}\{(y-\hat{y}({\bf x},{\bm \theta}))^2\} \nonumber \\
&= \mathbb{E}\{y^2-2y\hat{y}({\bf x};{\bm \theta})+[\hat{y}({\bf x};{\bm \theta})]^2\}\nonumber \\
&= \mathbb{E}\{y^2\}+2\mathbb{E}\{-y\hat{y}({\bf x};{\bm \theta})\}+\mathbb{E}\{[\hat{y}({\bf x};{\bm \theta})]^2 \}\nonumber\\
&=R_{\#}+\frac{2}{\prescript{}{2} N}\sum_{i=1}^{\prescript{}{2} N} V(\prescript{}{1} {\bm \theta},\prescript{}{2} {\bm \theta}_i)+\frac{1}{\prescript{}{2} N^2}\sum_{i,j=1}^{\prescript{}{2} N} U(\prescript{}{1} {\bm \theta},\prescript{}{2} {\bm \theta}_i,\prescript{}{2} {\bm \theta}_j)
\end{align}
where the constant $R_{\#}=\mathbb{E}\{y^2\}$ is the risk of the trivial predictor $\hat{y}=0$, and we define potentials $V(\prescript{}{1}{\bm \theta},\prescript{}{2} {\bm \theta})=-\mathbb{E}\{y\prescript{}{2} \sigma_*(\prescript{}{1} \sigma_*({\bf x},\prescript{}{1} {\bm \theta}),\prescript{}{2} {\bm \theta})\}$, $U(\prescript{}{1} {\bm \theta},\prescript{}{2} {\bm \theta}_1,\prescript{}{2} {\bm \theta}_2)=\mathbb{E}\{\prescript{}{2} \sigma_*(\prescript{}{1} \sigma_*({\bf x},\prescript{}{1} {\bm \theta}),\prescript{}{2} {\bm \theta}_1)\prescript{}{2} \sigma_*(\prescript{}{1} \sigma_*({\bf x},\prescript{}{1} {\bm \theta}),\prescript{}{2} {\bm \theta}_2) \}$. We note that $U(\cdot,\cdot)$ is a symmetric positive semidefinite kernel. 
%http://web.iitd.ac.in/~sumeet/CLT2008S-lecture18.pdf see proof of Mercer's Theorem for proof on positive semidefinite

$R_{\prescript{}{2} N}({\bm \theta})$ only depends on $\prescript{}{2} {\bm \theta}_1,\dots,\prescript{}{2} {\bm \theta}_{\prescript{}{2} N}$ through their empirical distribution $\hat{\rho}^{(\prescript{}{2} N)} ({\bm \eta})=\frac{1}{\prescript{}{2} N} \sum_{i=1}^{\prescript{}{2} N} \delta_{\prescript{}{2} {\bm \theta}_i}({\bm \eta})$, where ${\bm\eta}=({\bm \eta}_i)_{1\leq i\leq \prescript{}{2}N}$ with  

\[
  \delta_{\prescript{}{2} {\bm \theta}_i}({\bm \eta}) =
  \begin{cases}
                                   1 & if \text{ ${\bm \eta}_i \leq \prescript{}{2} {\bm \theta}_i$ }component-wise \\
                                   0 & otherwise
  \end{cases}
\]
\noindent and ``passively" also depends on $\prescript{}{1} {\bm \theta}$, the parameters from the first hidden layer.

We then consider a risk function defined for $\prescript{}{2}\rho \in P(\mathbb{R}^{\prescript{}{2}D})$, where $P(\Omega)$ is the space of probability distributions on $\Omega$:

\begin{align}
R(\prescript{}{2}\rho) = &R_\# +2\int V(\prescript{}{1} {\bm \theta},\prescript{}{2} {\bm \theta})\prescript{}{2}\rho (d\prescript{}{2} {\bm \theta}) + \int U(\prescript{}{1} {\bm \theta},\prescript{}{2} {\bm \theta},\prescript{}{2} {\bm \theta}')\prescript{}{2}\rho(d\prescript{}{2} {\bm \theta})\prescript{}{2}\rho(d\prescript{}{2} {\bm \theta}')
\end{align}

\noindent Essentially, we have that $R(\prescript{}{2}\rho)$ corresponds to the population risk when the number of hidden units in the second hidden layer goes to infinity, and the empirical distribution of parameters $\hat{\rho}^{(\prescript{}{2} N)}$ converges to $\prescript{}{2}\rho$, respectively. Due to $U(\cdot,\cdot,\cdot)$ being positive semidefinite, the risk becomes convex in the limit. 

The parameters of neural networks are learned using stochastic gradient descent (SGD) or some variation. For layer $j$, this leads to the iteration

\begin{align}
\prescript{}{j} {\bm \theta}_i^{k+1} = \prescript{}{j} {\bm \theta}_i^{k}+2s_k (y_k-\hat{y}({\bf x}_k; {\bm \theta}^k))\nabla_{\prescript{}{j} {\bm \theta}_i} \prescript{}{j} \sigma_* (\prescript{}{i-1}{\bf z}_k ; \prescript{}{j} {\bm \theta}_i^{k}).
\end{align}

\noindent Here, $\prescript{}{j} {\bm \theta}^k=(\prescript{}{j} {\bm \theta}_i^{k})_{i\leq \prescript{}{j} N}$ denotes the parameters in hidden layer $j$ after $k$ iterations, ${\bm \theta}^{k}=(\prescript{}{1} {\bm \theta}^{k},\prescript{}{2} {\bm \theta}^{k})$, $s_k$ as the step size, and $(\prescript{}{i-1}{\bf z}_k, y_k)$ as the $k$th example. We make the \emph{One-pass Assumption}, which assumes that training examples are never revisited. This is equivalent to $\{({\bf x}_k, y_k)\}_{k\geq 1}$ are i.i.d. $({\bf x}_k,y_k) \sim \mathbb{P}$.  

Our goal is to reduce learning in three-layer neural networks to analyzing the PDE systems in the next section. The structure of the differential dynamics in the next section follows very closely to those of the one and two dimensional diffusions partial differential equations, of which there is much mathematical literature. More specifically, we can view the system as gradient flows where the cost function is $R(\prescript{}{2} \rho;\prescript{}{1}\rho)$ in the space $(P(\mathbb{R}^{\prescript{}{2}D}),W_2)$, probability measures on $\mathbb{R}^{\prescript{}{2} D}$ with the Wasserstein metric, where the Wasserstein metric is defined as 

\begin{align}
W_2(\rho_1,\rho_2) = \bigg( \inf_{\gamma \in \Gamma(\rho_1,\rho_2)} \int \|{\bm \theta}_1 - {\bm \theta}_2\|_2^2 \gamma(d{\bm \theta}_1,d{\bm \theta}_2) \bigg)^{1/2}
\end{align}
where $\Gamma(\rho_1,\rho_2)$ represents the set of all couplings of $\rho_1$ and $\rho_2$ and $\| \cdot \|_2$ is the $\ell_2$-norm.

%https://www-dimat.unipv.it/savare/Ravello2010/JKO.pdf FOKKER-PLANCK Equation
%http://www2.stat.duke.edu/~sayan/ambrosio.pdf
%https://eprints.lib.hokudai.ac.jp/dspace/bitstream/2115/5878/1/AMO42-2.pdf
%https://arxiv.org/pdf/1704.00607.pdf

\section{Theoretical Work}
In this section, our interest is in defining a theoretical framework for the three layer neural network. More specifically, we are interested in establishing marginal distributions for each of the layers, as well as an overarching joint distribution for the entire network. We assume that the step size in the SGD is given by $s_k=\epsilon \xi(k\epsilon)$ where $\xi :\mathbb{R}_{\geq 0} \to \mathbb{R}_{\geq 0}$ is a sufficiently regular function. 
We will also denote $\hat{\prescript{}{1}\rho}_k^{(\prescript{}{1} N)}=\frac{1}{\prescript{}{1} N} \sum_{i=1}^{\prescript{}{1} N} \delta_{\prescript{}{1} {\bm \theta}_i^k}$ and $\hat{\prescript{}{2}\rho}_k^{(\prescript{}{2} N)}=\frac{1}{\prescript{}{2} N} \sum_{i=1}^{\prescript{}{2} N} \delta_{\prescript{}{2} {\bm \theta}_i^k}$ as the empirical distribution of parameters in the first and second hidden layer, 
respectively, after $k$ SGD steps. 

%\begin{align}
%\hat{\rho}_{t/\epsilon}^{(\prescript{}{2} N)} \Rightarrow \prescript{}{2}\rho_t
%\end{align}
%as $N\to \infty$ and $\epsilon \to 0$, where $\Rightarrow$ denotes weak convergence. 
%If we think of $\prescript{}{2} {\bm \theta}_1, \dots,\prescript{}{2} {\bm \theta}_{\prescript{}{2} N}$ as the positions of $\prescript{}{2} N$ particles in a $\prescript{}{2} D$-dimensional space, then when $\prescript{}{2}N$ is large enough, the density $\prescript{}{2}\rho_t (\prescript{}{2}{\bm \theta})$ describes the behavior of such a `cloud' of particles, where $t$ indexes time. We can define the asymptotic dynamics of $\prescript{}{2}\rho_t$ using the following PDE equations:
%\begin{align}
%\partial_t \prescript{}{2} \rho_t &= 2 \xi (t)\text{div}_{\prescript{}{2}{\bm \theta}}\{ \prescript{}{2}\rho_t \circ  \nabla_{\prescript{}{2}{\bm \theta}} \Psi ({\bm \theta}; \prescript{}{1} \rho_t, \prescript{}{2}\rho_t ) \}\\
%\partial_t \prescript{}{1} \rho_t &= 2 \xi (t)\text{div}_{\prescript{}{1}{\bm \theta}}\{ \prescript{}{2}\rho_t \circ  \nabla_{\prescript{}{1}{\bm \theta}} \Psi ({\bm \theta}; \prescript{}{1} \rho_t, \prescript{}{2}\rho_t ) \} \\
%\Psi({\bm \theta};\prescript{}{1}\rho,\prescript{}{2}\rho) &\equiv V(\prescript{}{1}{\bm \theta}, \prescript{}{2}{\bm \theta}|\prescript{}{1}{\bm \theta} )\prescript{}{1}\rho(\prescript{}{1}{\bm \theta})+ \int U(\prescript{}{1} {\bm \theta},\prescript{}{2} {\bm \theta},\prescript{}{2} {\bm \theta}'|\prescript{}{1} {\bm \theta}) \prescript{}{1}\rho(\prescript{}{1}{\bm \theta})\prescript{}{2}\rho(d\prescript{}{2} {\bm \theta}')
%\end{align}

\subsection{Theoretical Framework}

\subsubsection{Joint Distribution}

From our background section, we had that

\begin{align}
R(\prescript{}{2}\rho) = &R_\# +2\int V(\prescript{}{1} {\bm \theta},\prescript{}{2} {\bm \theta})\prescript{}{2}\rho (d\prescript{}{2} {\bm \theta}) + \int U(\prescript{}{1} {\bm \theta},\prescript{}{2} {\bm \theta},\prescript{}{2} {\bm \theta}')\prescript{}{2}\rho(d\prescript{}{2} {\bm \theta})\prescript{}{2}\rho(d\prescript{}{2} {\bm \theta}')
\end{align}

%https://math.stackexchange.com/questions/3217489/what-is-the-difference-between-pdx-and-pxdx
We can rewrite this in terms of conditionals by using the Law of Total Expectation (Tower Property) on our potentials $V(\cdot,\cdot)$ and $U(\cdot,\cdot,\cdot)$:

\begin{align}
R(\prescript{}{2}\rho;\prescript{}{1}\rho) = &R_\# +2\int \int V(\prescript{}{1} {\bm \theta},\prescript{}{2} {\bm \theta}|\prescript{}{1} {\bm \theta})\prescript{}{1}\rho (d\prescript{}{1} {\bm \theta}) \prescript{}{2}\rho (d\prescript{}{2} {\bm \theta}) \nonumber \\
&+ \int\int U(\prescript{}{1} {\bm \theta},\prescript{}{2} {\bm \theta},\prescript{}{2} {\bm \theta}'|\prescript{}{1} {\bm \theta}) \prescript{}{1}\rho (d\prescript{}{1} {\bm \theta})\prescript{}{2}\rho(d\prescript{}{2} {\bm \theta})\prescript{}{2}\rho(d\prescript{}{2} {\bm \theta}')
\end{align}

\noindent where $\prescript{}{1}\rho\in P(\mathbb{R}^{\prescript{}{1} D})$.

\begin{figure}[h]
\caption{$R(\prescript{}{2}\rho;\prescript{}{1}\rho)$ visualized in neural network.}
\label{JointDist}
\centering
\includegraphics[scale=0.5]{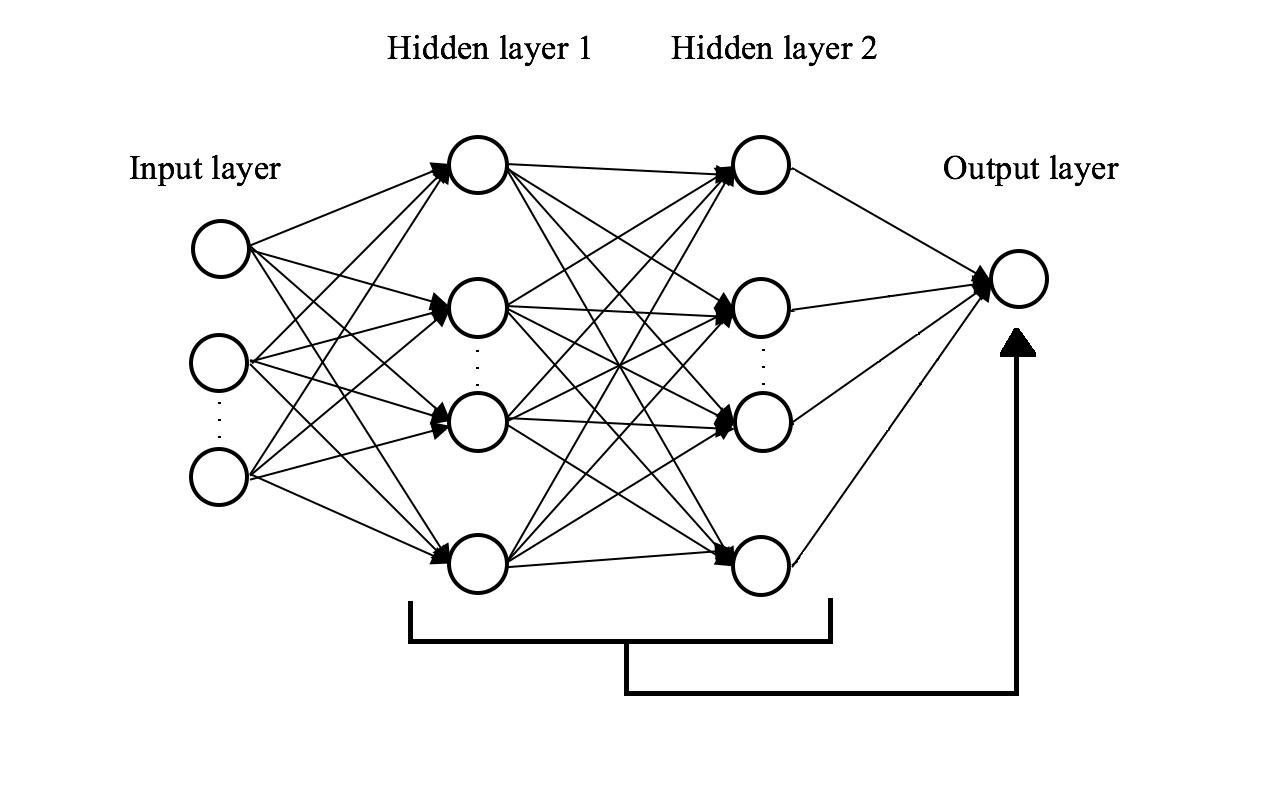}
\end{figure}

Thus, we can loosely think of $R(\prescript{}{2}\rho;\prescript{}{1}\rho)$ as the joint distribution of the two hidden layers in the neural network towards the output. This is visualized in Figure \ref{JointDist}. By taking the partial derivative of $R(\prescript{}{2}\rho ; \prescript{}{1}\rho)$ with respect to $\prescript{}{1}\rho$ and $\prescript{}{2}\rho$, we obtain

\begin{align}
\label{diffusionA} \prescript{}{1}R &\equiv \frac{\partial R(\prescript{}{2}\rho;\prescript{}{1}\rho)}{\partial \prescript{}{2}\rho}= V(\prescript{}{1}{\bm \theta}, \prescript{}{2}{\bm \theta})+ \int U(\prescript{}{1} {\bm \theta},\prescript{}{2} {\bm \theta},\prescript{}{2} {\bm \theta}') \prescript{}{2}\rho(d\prescript{}{2} {\bm \theta}')\\
\label{diffusionB} \prescript{}{2}R & \equiv \frac{\partial R(\prescript{}{2}\rho;\prescript{}{1}\rho)}{\partial \prescript{}{1}\rho}=  \int V(\prescript{}{1} {\bm \theta},\prescript{}{2} {\bm \theta}|\prescript{}{1} {\bm \theta})\prescript{}{2}\rho (d\prescript{}{2} {\bm \theta})+ \int U(\prescript{}{1} {\bm \theta},\prescript{}{2} {\bm \theta},\prescript{}{2} {\bm \theta}'|\prescript{}{1} {\bm \theta}) \prescript{}{2}\rho(d\prescript{}{2} {\bm \theta})\prescript{}{2}\rho(d\prescript{}{2} {\bm \theta}')
\end{align}

Consider $\prescript{}{1}R$ and $\prescript{}{2}R$ as the risks in the marginals for layer 1 and 2, respectively. In addition, let $\prescript{}{1,2}\rho$ denote the joint density function of $\prescript{}{1}{\bm \theta}$ and $\prescript{}{2}{\bm \theta}$. Assume that the marginal densities $\prescript{}{1} \rho_t$ and $\prescript{}{2} \rho_t$ are independent ($\star$). Then, we expect the asymptotic dynamics of $\prescript{}{1,2}\rho_t$ to follow 

\begin{align}
\partial_t \prescript{}{1,2} \rho_t &= \frac{\partial \prescript{}{1}\rho_t}{\partial t}\prescript{}{2}\rho_t +  \frac{\partial \prescript{}{2}\rho_t}{\partial t}\prescript{}{1}\rho_t
\end{align}

by the product rule. We will return to this after specifying the marginals. 

%%%%%% diffusion equations
%https://www.uni-muenster.de/imperia/md/content/physik_tp/lectures/ws2016-2017/num_methods_i/heat.pdf
%https://scipython.com/book/chapter-7-matplotlib/examples/the-two-dimensional-diffusion-equation/
%%%%%%% incompressible vs compressible flow
%https://web.stanford.edu/~cantwell/AA200_Course_Material/AA200_Ch1_Intro_to_fluid_flow.pdf
%https://www.chegg.com/homework-help/definitions/incompressible-and-compressible-flow-5
%%%these next two are most important 
%http://www.columbia.edu/itc/ldeo/lackner/E4900/Themelis5.pdf (5.2.16)
%https://ocw.mit.edu/courses/materials-science-and-engineering/3-185-transport-phenomena-in-materials-engineering-fall-2003/study-materials/3_21.pdf (61)

%RISKS ARE VIEWED AS ENERGIES...
\subsubsection{Marginal Distribution of Second Layer}

Turning towards the marginals, we first look at the marginal of the second layer. We can see that the representation in (\ref{diffusionB}) closely resembles the risk function from \cite{MMN2018}. Our risk is as follows: 

\begin{align}
\label{marg2}
\prescript{}{2}R(\prescript{}{2}\rho|\prescript{}{1} {\bm \theta}) \equiv \int V(\prescript{}{1} {\bm \theta},\prescript{}{2} {\bm \theta}|\prescript{}{1} {\bm \theta})\prescript{}{2}\rho (d\prescript{}{2} {\bm \theta}) +\int U(\prescript{}{1} {\bm \theta},\prescript{}{2} {\bm \theta},\prescript{}{2} {\bm \theta}'|\prescript{}{1} {\bm \theta}) \prescript{}{2}\rho(d\prescript{}{2} {\bm \theta})\prescript{}{2}\rho(d\prescript{}{2} {\bm \theta}').
\end{align}

The similarities to the risk function from \cite{MMN2018} seem reasonable as we are essentially dealing with a single hidden layer that is directly feeding towards the output layer, which is the type of neural network that \cite{MMN2018} was analyzing. It also makes sense for there to be conditioning on $\prescript{}{1}{\bm \theta}$ as we are essentially treating the first hidden layer as our input layer in this marginal representation. 

\begin{figure}[h]
\caption{$\prescript{}{2}R(\prescript{}{2}\rho)$ visualized in neural network.}
\label{Marg2Dist}
\centering
\includegraphics[scale=0.5]{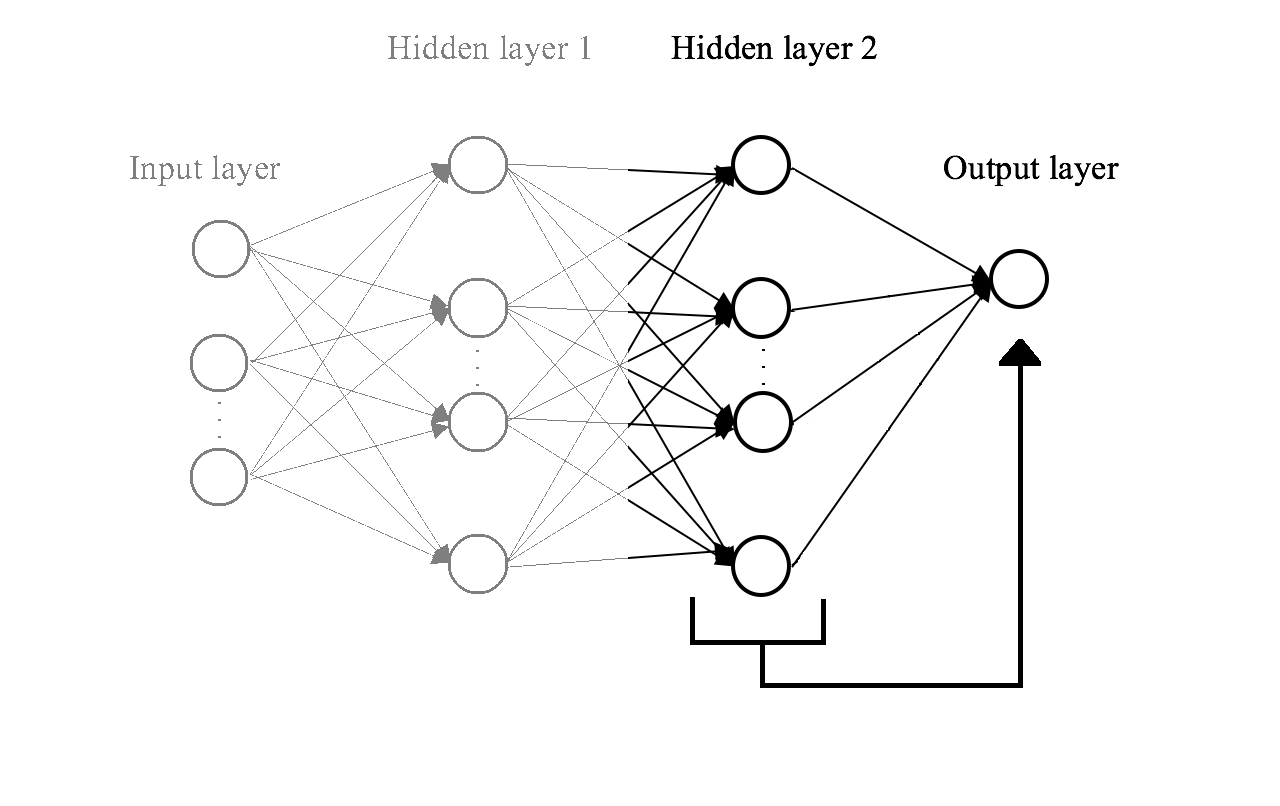}
\end{figure} 

Given the conditioning on $\prescript{}{1}{\bm \theta}$, we would then expect the asymptotic dynamics of $\prescript{}{2|1}{\rho}_t$, where $\prescript{}{2|1}{ \rho}_t$ represents the density of $\prescript{}{2}{\bm \theta}|\prescript{}{1}{\bm \theta}$, to follow similar dynamics to those in \cite{MMN2018}:

\begin{align}
\partial_t \prescript{}{2|1} \rho_t &=  2\xi (t)\text{div}_{\prescript{}{2}{\bm \theta}}\{ \prescript{}{2}\rho_t   \nabla_{\prescript{}{2}{\bm \theta}} \Psi_{2|1} ({\bm \theta};\prescript{}{2}\rho_t ) \} \\
\Psi_{2|1}({\bm \theta},\prescript{}{2}\rho) &\equiv V(\prescript{}{1}{\bm \theta}, \prescript{}{2}{\bm \theta}|\prescript{}{1}{\bm \theta} )+ \int U(\prescript{}{1} {\bm \theta},\prescript{}{2} {\bm \theta},\prescript{}{2} {\bm \theta}'|\prescript{}{1} {\bm \theta}) \prescript{}{2}\rho(d\prescript{}{2} {\bm \theta}') 
\end{align} 

\noindent where $\text{div}_{{\bm \theta}}\{ {\bf v} ({\bm \theta})\}$ represents the divergence of the vector field ${\bf v} ({\bm \theta})$. Then, since $\int \prescript{}{2|1} \rho \prescript{}{1}\rho(d\prescript{}{1}{\bm \theta})=\prescript{}{2}\rho$, we have that the asymptotic dynamics of $\prescript{}{2} \rho_t$ would follow

\begin{align}
\partial_t \prescript{}{2} \rho_t &=  2\xi (t)\text{div}_{\prescript{}{2}{\bm \theta}}\{ \prescript{}{2}\rho_t   \nabla_{\prescript{}{2}{\bm \theta}} \Psi_2 ({\bm \theta};\prescript{}{2}\rho_t ) \} \label{PDE2Layera}\\
\Psi_2({\bm \theta},\prescript{}{2}\rho) &\equiv V(\prescript{}{1}{\bm \theta}, \prescript{}{2}{\bm \theta})+ \int U(\prescript{}{1} {\bm \theta},\prescript{}{2} {\bm \theta},\prescript{}{2} {\bm \theta}') \prescript{}{2}\rho(d\prescript{}{2} {\bm \theta}')  \label{PDE2Layerb}
\end{align}

We are interested in showing that $\prescript{}{2}\rho_t$ is a good approximation of $\prescript{}{2}{\hat{\rho}}_k^{(\prescript{}{2} N)}$, $k=t/\epsilon$, as soon as $\epsilon << 1/\prescript{}{2}D$ and $\prescript{}{2} N >> \prescript{}{2}D$. 

%%%%%%%%%%%%%%%%%%%%%%%%%%%%%%%%%%%%%%%%%
\subsubsection{Marginal Distribution of First Layer}
%https://math.stackexchange.com/questions/836335/fundamental-theorem-of-calculus-for-double-integrals-over-a-single-variable
We identify the marginal risk of the first layer as similar to that in (\ref{diffusionA}):

\begin{align}
\prescript{}{1}R(\prescript{}{1}\rho)\equiv V(\prescript{}{1}{\bm \theta}, \prescript{}{2}{\bm \theta})+ \int U(\prescript{}{1} {\bm \theta},\prescript{}{2} {\bm \theta},\prescript{}{2} {\bm \theta}') \prescript{}{2}\rho(d\prescript{}{2} {\bm \theta}')
\end{align}

\begin{figure}[h]
\caption{$\prescript{}{1}R(\prescript{}{1}\rho)$ visualized in neural network.}
\label{Marg1Dist}
\centering
\includegraphics[scale=0.5]{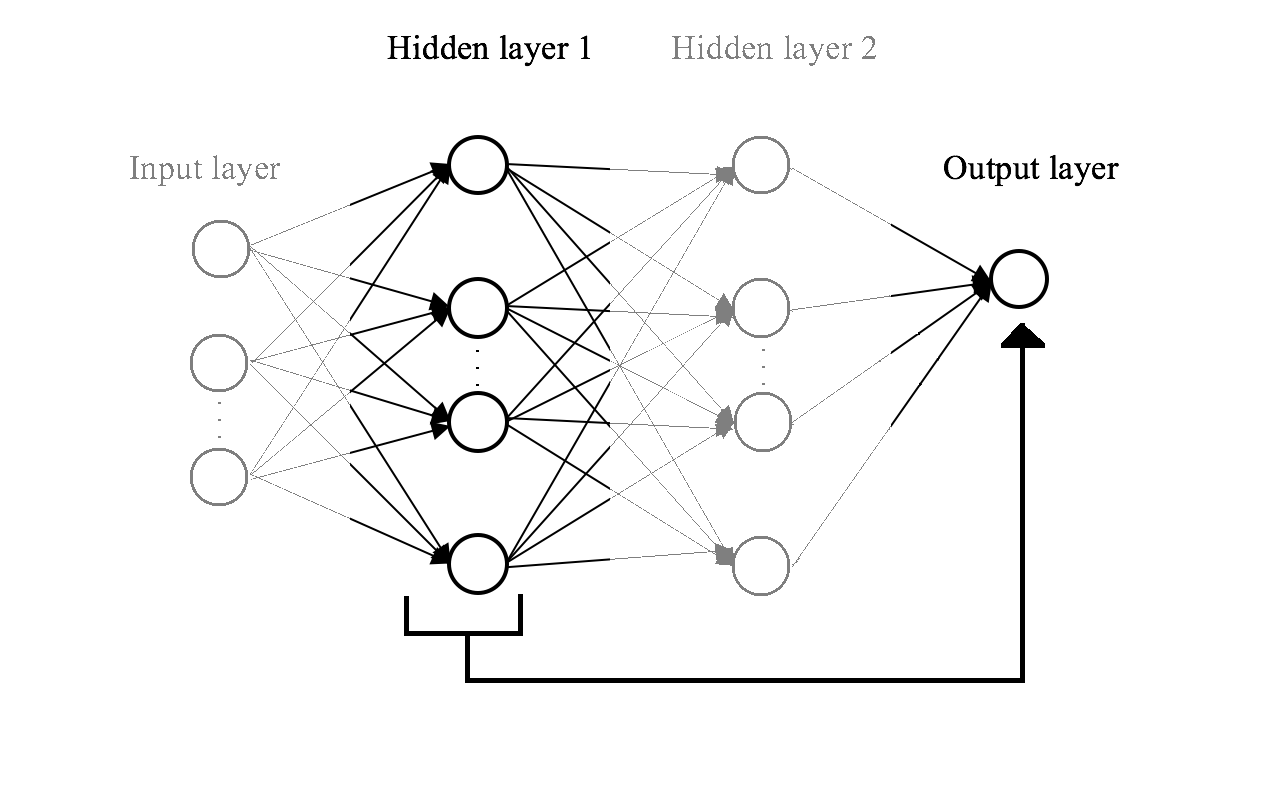}
\end{figure}

Again, we just have a single hidden layer flowing to an output layer; however, we are still ``passively" moving through the second hidden layer (pictured in Figure \ref{Marg1Dist}) to get to the final output layer, so we expect the asymptotic dynamics of $\prescript{}{1|2}\rho_t$, where $\prescript{}{1|2}{ \rho}_t$ represents the density of $\prescript{}{1}{\bm \theta}|\prescript{}{2}{\bm \theta}$, to follow asymptotic dynamics

\begin{align}
\partial_t \prescript{}{1|2} \rho_t &= 2\xi (t)\text{div}_{\prescript{}{1}{\bm \theta}}\{ \prescript{}{1}\rho_t  \nabla_{\prescript{}{1}{\bm \theta}} \Psi_{1|2} ({\bm \theta};\prescript{}{2}\rho_t ) \} \\
\Psi_{1|2}({\bm \theta},\prescript{}{2}\rho) &\equiv V(\prescript{}{1}{\bm \theta}, \prescript{}{2}{\bm \theta}|\prescript{}{2}{\bm \theta} )+ \int U(\prescript{}{1} {\bm \theta},\prescript{}{2} {\bm \theta},\prescript{}{2} {\bm \theta}'|\prescript{}{2} {\bm \theta}) \prescript{}{2}\rho (\prescript{}{2} {\bm \theta} )   
\end{align} 

\noindent where $\text{div}_{{\bm \theta}}\{ {\bf v} ({\bm \theta})\}$ represents the divergence of the vector field ${\bf v} ({\bm \theta})$. Then, since $\int \prescript{}{1|2} \rho \prescript{}{2}\rho(d\prescript{}{2}{\bm \theta})=\prescript{}{1}\rho$, we have that the asymptotic dynamics of $\prescript{}{1} \rho_t$ would follow

\begin{align}
\partial_t \prescript{}{1} \rho_t &= 2\xi (t)\text{div}_{\prescript{}{1}{\bm \theta}}\{ \prescript{}{1}\rho_t  \nabla_{\prescript{}{1}{\bm \theta}} \Psi_{1} ({\bm \theta};\prescript{}{2}\rho_t ) \} \label{PDE1Layera}\\
\Psi_{1}({\bm \theta},\prescript{}{2}\rho) &\equiv V(\prescript{}{1}{\bm \theta}, \prescript{}{2}{\bm \theta})+ \int U(\prescript{}{1} {\bm \theta},\prescript{}{2} {\bm \theta},\prescript{}{2} {\bm \theta}') \prescript{}{2}\rho (\prescript{}{2} {\bm \theta} )   \label{PDE1Layerb}
\end{align} 

We are interested in showing that $\prescript{}{1}\rho_t$ is a good approximation of $\prescript{}{1}{\hat{\rho}}_k^{(\prescript{}{1} N)}$, $k=t/\epsilon$, as soon as $\epsilon << 1/\prescript{}{1}D$ and $\prescript{}{1} N >> \prescript{}{1}D$.

%%%%%%%%%%%%%%%%%%%%%%%%%%%%%%%%%%%%%%%%%
\subsubsection{Revisiting the Joint Distribution}

From before, under $(\star)$, $\partial_t \prescript{}{1,2} \rho_t = \frac{\partial \prescript{}{1}\rho_t}{\partial t}\prescript{}{2}\rho_t +  \frac{\partial \prescript{}{2}\rho_t}{\partial t}\prescript{}{1}\rho_t$. Combining the results from the previous section, we then have that 

\begin{align}
\partial_t \prescript{}{1,2}\rho_t &= 2\xi(t)\text{div}_{\bm \theta} \{\prescript{}{1,2}\rho_t \nabla_{\bm \theta} \Psi({\bm \theta},\prescript{}{2}\rho_t) \}\\
\Psi({\bm \theta},\prescript{}{2} \rho) &\equiv V(\prescript{}{1}{\bm \theta}, \prescript{}{2}{\bm \theta})+ \int U(\prescript{}{1} {\bm \theta},\prescript{}{2} {\bm \theta},\prescript{}{2} {\bm \theta}') \prescript{}{2}\rho (\prescript{}{2} {\bm \theta} )
\end{align}

\subsubsection{Simulation Test for Independence of Marginals}
%https://stats.stackexchange.com/questions/73646/how-do-i-test-that-two-continuous-variables-are-independent

In order to get a sense of if $(\star)$ has weight to be proven, we first run some simulations to see if there is indeed independence in the weights from the two hidden layers. To test for the independence between the two distributions, we will use the Hoeffding non-parametric independence test, where $H_0$ is that the two distributions are independent, and $H_A$ is that they are not independent (see \cite{H1948}). The Hoeffding $D$ statistic is calculated as

\begin{align}
D &= \frac{A-2(n-2)B+(n-2)(n-3)C}{n(n-1)(n-2)(n-3)(n-4)} \\
A &= \sum_{\alpha =1}^n a_\alpha (a_\alpha -1)b_\alpha (b_\alpha -1) \\
B &= \sum_{\alpha=1}^n (a_\alpha -1)(b_\alpha-1) c_\alpha \\
C &= \sum_{\alpha=1}^n c_\alpha (c_\alpha -1) \\
a_\alpha &= \sum_{\beta=1}^n \gamma (X_\alpha - X_\beta)-1 \\
b_\alpha &= \sum_{\beta=1}^n \gamma(Y_\alpha - Y_\beta)-1 \\
c_\alpha &= \sum_{\beta=1}^n \gamma (X_\alpha-X_\beta)\gamma(Y_\alpha - Y_\beta)-1
\end{align}

\noindent where $\gamma(u)=1$ if $u\geq 0$, and 0 otherwise and $n$ is the number of data points we have for each density. P-values on the test statistic $D$ are approximated by linear interpolation on the table in Hollander and Wolfe, which uses the asymptotically equivalent Blum-Kiefer-Rosenblatt statistic (see \cite{HolWolf1973}).

\begin{table}[h!]
  \begin{center}
    \caption{Table comparing p-values from Hoeffding tests.}
    \label{tab:hoeffding}
    \begin{tabular}{c|c|c|c|c|c} 
      $N$ & $\lambda$ & \textbf{Hidden Nodes per Layer} & $d$ & \textbf{mean p-value} &\textbf{\% of p-values less than 0.05}\\
      \hline
      5000 & 0.001 & 100 & 100 & 0.396 & $5 \%$ \\
      5000 & 0.01 & 100 & 100& 0.420& $ 6\%$ \\
      1000 & 0.001 & 100 & 100 &0.364 & $9 \%$ \\
      1000 & 0.01 & 100 & 100& 0.407& $ 6\%$ \\
      5000 & 0.001 & 100 & 50& 0.375 & $ 2\%$ \\
      5000 & 0.01 & 100 & 50&0.408 & $ 4\%$ \\
      1000 & 0.001 & 100 & 50&0.403 & $1 \%$ \\
      1000 & 0.01 & 100 & 50&0.406 & $6 \%$ \\
      5000 & 0.001 & 50 & 100&0.413 & $ 5\%$ \\
      5000 & 0.01 & 50 & 100&0.405 & $ 2\%$ \\
      1000 & 0.001 & 50 & 100&0.410 & $7 \%$ \\
      1000 & 0.01 & 50 & 100&0.391 & $ 5\%$ \\
      5000 & 0.001 & 50 & 50&0.437 & $ 3\%$ \\
      5000 & 0.01 & 50 & 50 &0.388 & $ 3\%$ \\
      1000 & 0.001 & 50 & 50 &0.409 & $6 \%$ \\
      1000 & 0.01 & 50 & 50 &0.410 & $11 \%$  \\
      250 & 0.001 & 100 & 100 & 0.395 & $ 13\%$ \\
      250 & 0.01 & 100 & 100 & 0.407 & $7 \%$ \\
      250 & 0.001 & 100 & 50 & 0.436 & $ 7\%$ \\
      250 & 0.01 & 100 & 50 &0.403  & $3 \%$ 
    \end{tabular}
  \end{center}
\end{table}

We create a target vector $y$ of size $N$, where $y_i=1$ with probability $1/2$ and $0$ with probability $1/2$. Next, we create our covariance matrix, ${\bm \Sigma}$ of size $d$ by $d$, where the diagonal entries are $(1+\Delta)^2$ and the off diagonals are $0.001$ or $0.01$ ($\lambda$), and $d$ is the number of parameters/predictor variables. Then, sample $x\sim N(0,{\bf \Sigma})$. We will vary $N$, $\lambda$, and the number of weights in each hidden layer (the hidden layers will use the same number of nodes). The standard neural network produced in the neuralnet package will be used, only specifying the number of hidden nodes in each layer and the classification parameter (lin.output = TRUE). For each specification of the parameters $N$, $\lambda$, $d$, and number of hidden nodes in each layer, we run the simulation 100 times and calculate the Hoeffding test each time and record the p-value from each test. The results are produced in Table \ref{tab:hoeffding}.

From the results, it appears that the two weight distributions are independent in the varying simulations, and our Type I Error for the most part falls in line with our chosen significance level of $0.05$. 

%%%%%%%%%%%%%%%%%%%%%%%%%%%%%%%%%%%%%%%%%
\subsection{Convergence Properties}

We first want to show that the minimum of the asymptotic risk $R(\prescript{}{2}\rho)$ is a good approximation to minimizing the finite-$\prescript{}{2} N$ risk $R_{\prescript{}{2}N}( {\bm \theta})$. 
%%%
\begin{proposition}\label{prop1}
Assume one of the following conditions hold:
%https://math.stackexchange.com/questions/53794/simple-question-the-double-supremum

(a) $\inf_{\prescript{}{2} \rho} R(\prescript{}{2}\rho)$ is achieved by some distribution $\prescript{}{2} \rho_*$ s.t.  $ \int U(\prescript{}{1} {\bm \theta},\prescript{}{2} {\bm \theta},\prescript{}{2} {\bm \theta}') \prescript{}{2}\rho_*(d\prescript{}{2} {\bm \theta}) \leq K$;

(b) $\exists \epsilon_0>0$ s.t. for any $\prescript{}{2} \rho \in P(\mathbb{R}^{\prescript{}{2} D})$ s.t. $R(\prescript{}{2}\rho) \leq \inf_{\prescript{}{2}\rho} R(\prescript{}{2}\rho)+\epsilon_0$ we have $\int U(\prescript{}{1} {\bm \theta},\prescript{}{2} {\bm \theta},\prescript{}{2} {\bm \theta}')\prescript{}{2}\rho(d\prescript{}{2} {\bm \theta}) \leq K$.

Then,
\begin{align}
| \inf_{\prescript{}{2}{\bm \theta}} R_{\prescript{}{2} N} ({\bm \theta}) - \inf_{\prescript{}{2} \rho} R({\prescript{}{2} \rho})|\leq K/\prescript{}{2}N.
\end{align}

%In addition, assume that $\prescript{}{2}{\bm \theta} \mapsto V(\prescript{}{1}{\bm \theta},\prescript{}{2} {\bm \theta}|\prescript{}{1}{\bm \theta})$ and $(\prescript{}{2} {\bm \theta},\prescript{}{2} {\bm \theta}')\mapsto U(\prescript{}{1} {\bm \theta},\prescript{}{2} {\bm \theta},\prescript{}{2} {\bm \theta}'|\prescript{}{1} {\bm \theta}) $ are continuous, with $U$ bounded from below. If $\inf_{\prescript{}{2}{\bm \theta} \in \mathbb{R}^{\prescript{}{2}D}} \Psi ({\bm \theta}, \prescript{}{2} \rho_*) > -\infty$ and $\text{supp} (\prescript{}{2}\rho_*)\subseteq \arg\min_{\prescript{}{2}{\bm \theta} \in \mathbb{R}^{\prescript{}{2}D}} \Psi ({\bm \theta}, \prescript{}{2} \rho_*)$, then the probability measure $\prescript{}{2}\rho_*$ is a global minimum of $R$. 
\end{proposition} 

\begin{proof}
We see that for any $\prescript{}{2}{\bm \theta} = (\prescript{}{2}{\bm \theta}_i)_{i\leq \prescript{}{2}N}$, we have that $R_{\prescript{}{2}N}({\bm \theta})\geq \inf_{\prescript{}{2}\rho} R(\prescript{}{2}\rho)$ where ${\bm \theta}=(\prescript{}{1}{\bm \theta},\prescript{}{2}{\bm \theta})$. Equality holds when $\rho_2=\frac{1}{\prescript{}{2}N} \sum_{i=1}^n \delta_{\prescript{}{2}{\bm \theta}_i}$. 

Let $\prescript{}{2}\rho_*\in P(\mathbb{R}^{\prescript{}{2}D})$ satisfy assumption (a), i.e. $\prescript{}{2}\rho_*$ achieves $\inf_{\prescript{}{2}\rho} R(\prescript{}{2}\rho)$, or $R(\prescript{}{2}\rho)\leq R(\prescript{}{2}\rho_*)+\epsilon$ for any $\prescript{}{2}\rho \in P(R^{\prescript{}{2}D})$ under assumption (b). Assume $(\prescript{}{2}{\bm \theta}_i)_{i \leq \prescript{}{2}N} \sim_{i.i.d.} \prescript{}{2}\rho_*$.  Since $\int U(\prescript{}{1}{\bm \theta},\prescript{}{2}{\bm \theta}, \prescript{}{2}{\bm \theta}' )\prescript{}{2}\rho_*(d\prescript{}{2}{\bm \theta})\prescript{}{2}\rho_*(d\prescript{}{2}{\bm \theta}')=E\{y({\bf x})^2 \}\geq 0$ for $y({\bf x})=\int \prescript{}{2}\sigma_*(\prescript{}{1}\sigma_*({\bf x},\prescript{}{1}{\bm \theta}),\prescript{}{2}{\bm \theta})\prescript{}{2}\rho_*(d \prescript{}{2}{\bm \theta})$, we have that

\begin{align}
\mathbb{E}_{\prescript{}{2}{\bm \theta}}[R_{\prescript{}{2}N}({\bm \theta})]-R(\prescript{}{2}\rho_*) &= \frac{1}{\prescript{}{2}N}\bigg\{\int U(\prescript{}{1}{\bm \theta},\prescript{}{2}{\bm \theta},\prescript{}{2}{\bm \theta}' )\prescript{}{2}\rho_*(d\prescript{}{2}{\bm \theta})\nonumber \\
&\hspace{1cm} -\int U(\prescript{}{1}{\bm \theta},\prescript{}{2}{\bm \theta},\prescript{}{2}{\bm \theta}' )\prescript{}{2}\rho_*(d\prescript{}{2}{\bm \theta})\prescript{}{2}\rho_*(d\prescript{}{2}{\bm \theta}') \bigg\} \\
&\leq \frac{1}{\prescript{}{2}N} \int U(\prescript{}{1}{\bm \theta},\prescript{}{2}{\bm \theta},\prescript{}{2}{\bm \theta}' )\prescript{}{2}\rho_*(d\prescript{}{2}{\bm \theta}) \leq \frac{K}{\prescript{}{2}N}, \label{prop1inequality}
\end{align}

\noindent where the last inequality follows from assumption. Thus, 

\begin{align}
\inf_{\prescript{}{2}{\bm \theta}} R_{\prescript{}{2}N}({\bm \theta}) \leq R(\prescript{}{2}\rho_*)+\frac{K}{\prescript{}{2}N}+\epsilon.
\end{align}

Since $\epsilon$ is arbitrary, our claim follows.
\end{proof}

\noindent Looking towards the PDE systems, we make the following assumptions to establish that the PDE systems describe the limit of the SGD dynamics.

\begin{assump}\label{assump1}
$t \mapsto \xi(t)$ is bounded Lipschitz: $\|\xi\|_\infty, \|\xi\|_{\text{Lip}} \leq K_1$, with $\int_0^\infty \xi(t)dt=\infty$. 
\end{assump}
\begin{assump}\label{assump2}
The activation functions $({\bf z},\prescript{}{i}{\bm \theta}) \mapsto \prescript{}{i} \sigma_* ({\bf z}; \prescript{}{i} {\bm \theta})$ are bounded, with sub-Gaussian gradient: $\|\prescript{}{i}\sigma_*\|_\infty \leq K_2$, $\|\nabla_{\prescript{}{i} {\bm \theta}} \prescript{}{i} \sigma_* (\cdot;\prescript{}{i} {\bm \theta})\|_{\psi_2} \leq K_2$ for $i=1,2$, where $\|\cdot \|_{\psi_2}$ is the Orlicz norm. Labels are also bounded $|y_k| \leq K_2$. 
\end{assump}
\begin{assump}\label{assump4}
The gradients $\prescript{}{2}{\bm \theta} \mapsto \nabla_{\prescript{}{2} {\bm \theta}} V(\prescript{}{1}{\bm \theta},\prescript{}{2} {\bm \theta})$ and $(\prescript{}{2} {\bm \theta},\prescript{}{2} {\bm \theta}')\mapsto \nabla_{\prescript{}{2} {\bm \theta}} U(\prescript{}{1} {\bm \theta},\prescript{}{2} {\bm \theta},\prescript{}{2} {\bm \theta}') $ are bounded, Lipschitz continuous, i.e. we have that $\|\nabla_{\prescript{}{2} {\bm \theta}} V(\prescript{}{1}{\bm \theta},\prescript{}{2} {\bm \theta})\|_2,\|\nabla_{\prescript{}{2} {\bm \theta}} U(\prescript{}{1} {\bm \theta},\prescript{}{2} {\bm \theta},\prescript{}{2} {\bm \theta}') \|_2 \leq K_3$, $\| \nabla_{\prescript{}{2} {\bm \theta}} V(\prescript{}{1}{\bm \theta},\prescript{}{2} {\bm \theta}) - \nabla_{\prescript{}{2} {\bm \theta}} V(\prescript{}{1}{\bm \theta},\prescript{}{2} {\bm \theta}')\|_2\leq K_3 \|\prescript{}{2} {\bm \theta}-\prescript{}{2}{\bm \theta}'\|_2$, and $\| \nabla_{\prescript{}{2} {\bm \theta}} U(\prescript{}{1} {\bm \theta},\prescript{}{2} {\bm \theta}_1,\prescript{}{2} {\bm \theta}_2) -\nabla_{\prescript{}{2} {\bm \theta}} U(\prescript{}{1} {\bm \theta},\prescript{}{2} {\bm \theta}_1',\prescript{}{2} {\bm \theta}_2') \|_2 \leq K_3\|(\prescript{}{2}{\bm \theta}_1,\prescript{}{2}{\bm \theta}_2)-(\prescript{}{2}{\bm \theta}_1',\prescript{}{2}{\bm \theta}_2') \|_2$.
\end{assump}

With these assumptions, we state the convergence of the SGD process to the PDE model in the second hidden layer and the first hidden layer:
%%%
\begin{thm}\label{layer2convergence}
Assume Assumptions \ref{assump1}, \ref{assump2}, and \ref{assump4} hold. For $\prescript{}{i} \rho_0 \in P(\mathbb{R}^{\prescript{}{i}D})$ for $i=1,2$, consider SGD with initialization $(\prescript{}{i} {\bm \theta}_j^0)_{j \leq {\prescript{}{i} N}} \sim_{i.i.d.} \prescript{}{i} \rho_0$ and step size $s_k = \epsilon \xi(k\epsilon)$. For $t\geq 0$, let $\prescript{}{2}\rho_t$ be the solution of the PDE system from (\ref{PDE2Layera}) and (\ref{PDE2Layerb}). Then, for any fixed $t\geq 0$, $\prescript{}{2}{\hat \rho}_{\lfloor t/\epsilon \rfloor}^{(\prescript{}{2} N)} \Rightarrow \prescript{}{2} \rho_t$ almost surely along any sequence $(\prescript{}{2}N, \epsilon=\epsilon_{\prescript{}{2}N})$ s.t. $N\to \infty$, $\epsilon_{\prescript{}{2}N}\to 0$, $\prescript{}{2}N/\log (\prescript{}{2} N /\epsilon_{\prescript{}{2}N}) \to \infty$ and $\epsilon_{\prescript{}{2} N} \log(\prescript{}{2} N/\epsilon_{\prescript{}{2}N})\to 0$. 
%The extra terms $\prescript{}{2}N/\log (\prescript{}{2} N /\epsilon_{\prescript{}{2}N}) \to \infty$ and $\epsilon_{\prescript{}{2} N} \log(\prescript{}{2} N/\epsilon_{\prescript{}{2}N})\to 0$ are used to make sure that the bound doesn't get out of control.
\end{thm}

\begin{proof}
For this proof, let $K$ denote a generic constant depending on the constants $K_1$, $K_2$, and $K_3$ from Assumptions \ref{assump1}, \ref{assump2}, and \ref{assump3}. From (\ref{yhat}), $\prescript{}{1}\sigma_*({\bf x},\prescript{}{1}{\bm \theta})$ represents the vector of size $\prescript{}{1}D$ of values produced by the activation function $\prescript{}{1}\sigma_*$ from the input layer towards the first hidden layer. Letting ${\bm \beta}_k = (\prescript{}{1}\sigma_*({\bf x}_k,\prescript{}{1}{\bm \theta}),y_k)$ denote the $k$th example, we define

\begin{align}
\prescript{}{2}{\bf F}_i({\bm \theta};{\bm \beta}_k)&=\big(y_k-{\hat y}({\bf x}_k;{\bm \theta})\big)\nabla_{\prescript{}{2}{\bm \theta}_i} \prescript{}{2}\sigma_*(\prescript{}{1}\sigma_*({\bf x}_k,\prescript{}{1}{\bm \theta}),\prescript{}{2}{\bm \theta}_i ), \hspace{2cm} \prescript{}{2}{\bm \theta}=(\prescript{}{2}{\bm \theta}_i)_{i\leq \prescript{}{2}N} \in \mathbb{R}^{\prescript{}{2} D\times \prescript{}{2}N} \\
\prescript{}{2}{\bf G}({\bm \theta};\prescript{}{2}\rho) &= - \nabla_{\prescript{}{2}{\bm \theta}} \Psi_2({\bm \theta},\prescript{}{2}\rho)=-\nabla_{\prescript{}{2}{\bm \theta}} V(\prescript{}{1}{\bm \theta},\prescript{}{2}{\bm \theta})-\int \nabla_{\prescript{}{2}{\bm \theta}} U(\prescript{}{1}{\bm \theta},\prescript{}{2}{\bm \theta},\prescript{}{2}{\bm \theta}')\prescript{}{2}\rho(d\prescript{}{2}{\bm \theta}'), \hspace{1cm} \prescript{}{2}{\bm \theta} \in \mathbb{R}^{\prescript{}{2}D}.
\end{align}

By the assumption of bounded Lipschitz for $\nabla_{\prescript{}{2}{\bm \theta}} V$ and $\nabla_{\prescript{}{2}{\bm \theta}} U$, we have that $\|\prescript{}{2}{\bf G}({\bm \theta};\prescript{}{2}\rho)\|_2 \leq K$ and $\|\prescript{}{2}{\bf G}({\bm \theta}_1;\prescript{}{2}\rho)-\prescript{}{2}{\bf G}({\bm \theta}_2;\prescript{}{2}\rho)\|_2 \leq K\|{\bm \theta}_1-{\bm \theta}_2\|_2$. In addition,

\begin{align}
\|\prescript{}{2}{\bf G}({\bm \theta};\prescript{}{2}\rho_1)-\prescript{}{2}{\bf G}({\bm \theta};\prescript{}{2}\rho_2)\|_2=\bigg\|\int \nabla_{\prescript{}{2}{\bm \theta}} U(\prescript{}{1}{\bm \theta},\prescript{}{2}{\bm \theta},\prescript{}{2}{\bm \theta}' ) (\prescript{}{2}\rho_1-\prescript{}{2}\rho_2)(d\prescript{}{2}{\bm \theta}')\bigg\|_2 \leq K d_{\text{BL}}(\prescript{}{2}\rho_1,\prescript{}{2}\rho_2), \label{GLip}
\end{align} 

\noindent where $d_{\text{BL}}(\cdot,\cdot)$ is the bounded Lipschitz distance between probability measures, i.e. 

$$d_{\text{BL}}(\mu,\eta)=\sup\{|\int f({\bf x})\mu(d{\bf x})-\int f({\bf x})\eta(d{\bf x})|: \|f\|_\infty \leq 1, \|f\|_{\text{Lip}} \leq 1 \}$$

and $\|f \|_{\text{Lip}}\equiv \sup_{x\not= y}|f({\bf x})-f({\bf y})|/\|{\bf x}-{\bf y}\|_2$.

We can now rewrite the SGD dynamics in the second layer as

\begin{align}
\prescript{}{2}{\bm \theta}_i^{k+1}&=\prescript{}{2}{\bm \theta}_i^{k}+2\epsilon \xi(k\epsilon)\prescript{}{2}{\bf F}_i({\bm \theta}_i^k,{\bm \beta}_{k+1}) \\
\Rightarrow \prescript{}{2}{\bm \theta}_i^{k}&=\prescript{}{2}{\bm \theta}_i^{0}+2\epsilon \sum_{\ell=0}^{k-1} \xi(\ell\epsilon)\prescript{}{2}{\bf F}_i({\bm \theta}_i^\ell,{\bm \beta}_{\ell+1}). \label{newSGD}
\end{align}

In order to define nonlinear dynamics to analyze the PDE system for the first layer, we introduce trajectories $(\prescript{}{1}{\bar{\bm \theta}}_i^t,\prescript{}{2}{\bar{\bm \theta}}_i^t)_{1,\leq i, \leq \prescript{}{1}N,1\leq j \leq \prescript{}{2}N,t\in \mathbb{R}_{\geq 0}}$ and letting ${\bar{\bm \theta}}_i^0={\bm \theta}_i^0$ be the same initialization as for the SGD. Then, for $t\geq 0$, the nonlinear dynamics follow

\begin{align}
\prescript{}{2}{\bar{\bm \theta}}_i^t&=\prescript{}{2}{\bar{\bm \theta}}_i^0-2\int_0^t \xi(s) \nabla_{\prescript{}{2}{\bar{\bm \theta}}}\Psi_2({\bar{\bm \theta}}_i^s,\prescript{}{2}\rho_s)ds \label{nonlineardynamics}\\
\prescript{}{2}\rho_s&=P_{\prescript{}{2}{\bar{\bm \theta}}_i^s}, 
\end{align}

\noindent where $P_X$ is the law of the random variable $X$. From \cite{Szn91}, under Assumptions \ref{assump1} and \ref{assump4}, the nonlinear dynamics has a unique solution, with $\prescript{}{2}\rho_t$ satisfying (\ref{PDE2Layera}). We notice that under the nonlinear dynamics, the trajectories $(\prescript{}{2}{\bar{\bm \theta}}_i^t)_{1\leq i \leq \prescript{}{2}N, t\in \mathbb{R}_{\geq 0}}$ are i.i.d., which implies that 

\begin{align}
\frac{1}{\prescript{}{2}N}\sum_{i=1}^{\prescript{}{2}N} \delta_{\prescript{}{2}{\bar{\bm \theta}}_i^t } \Rightarrow \prescript{}{2}\rho_t \label{nonlineardynamicsrhot}
\end{align}

\noindent in distribution. We rewrite (\ref{nonlineardynamics}) as 

\begin{align}
\prescript{}{2}{\bar{\bm \theta}}_i^t&=\prescript{}{2}{\bar{\bm \theta}}_i^0+2\int_0^t \xi(s) \prescript{}{2}{\bf G}({\bar{\bm \theta}}_i^s;\prescript{}{2}\rho_s)ds \label{newnonlineardynamics}
\end{align}

\noindent and compare it to (\ref{newSGD}). We will define $[t]=\epsilon\lfloor t/\epsilon\rfloor$ for $t\in \mathbb{R}_{\geq 0}$. We next detail some Lemmas that bound the difference between the nonlinear dynamics and original dynamics.

\begin{lemma}\label{Lemma1Lip}
Assume Assumptions \ref{assump1} and \ref{assump4} hold. Let $(\prescript{}{2}\rho_t)_{t\geq0}$ be the solution of (\ref{PDE2Layera}). Let $(\prescript{}{2}{\bar{\bm \theta}}_i^t)_{t\geq 0}$ be the solution of (\ref{nonlineardynamics}). Then $t\mapsto \prescript{}{2}{\bar{\bm \theta}}_i^t$ is $K_1K_3$-Lipschitz continuous, and $t\mapsto \prescript{}{2}\rho_t$ is $K_1K_3$-Lipschitz continuous in $W_2$ Wasserstein distance. 
\end{lemma}

\begin{proof}
Since $\xi$ is $K_1$ bounded and $\nabla_{\prescript{}{2}{{\bm \theta}}}\Psi_2$ is $K_3$ bounded, $t\mapsto \prescript{}{2}{\bar{\bm \theta}}_i^t$ is $K_1K_3$-Lipschitz continuous. In addition, since 

$$d_{\text{BL}}(\prescript{}{2}\rho_t,\prescript{}{2}\rho_s) \leq W_2(\prescript{}{2}\rho_t,\prescript{}{2}\rho_s) \leq  (\mathbb{E}[\|\prescript{}{2}{\bar{\bm \theta}}_i^t-\prescript{}{2}{\bar{\bm \theta}}_i^s\|_2^2])^{1/2} \leq K_1K_3|t-s|,$$

\noindent $t\mapsto \prescript{}{2}\rho_t$ is Lipschitz continuous in $W_2$ Wasserstein distance.
\end{proof}

\begin{lemma} \label{Lemma2Lip}
Under Assumptions \ref{assump1}, \ref{assump2}, \ref{assump4}, $\exists$ constant $K$, dependent on $K_1$, $K_2$, and $K_3$, s.t. for any $T\geq 0$, 

\begin{align}
\max_{i\leq \prescript{}{2}N} \sup_{\ell\in [0,T/\epsilon]\cap \mathbb{N}} \|\prescript{}{2}{\bm \theta}_i^\ell -\prescript{}{2}{\bar{\bm \theta}}_i^{\ell\epsilon} \|_2
\leq KTe^{KT} \bigg[\epsilon+\frac{1}{\prescript{}{2}N}+\sqrt{1/\prescript{}{2}N\vee \epsilon}\big(\sqrt{\prescript{}{2}D+\log(\prescript{}{2}N(t/\epsilon \vee 1))}+z \big) \bigg]
%\leq K e^{KT}\sqrt{1/\prescript{}{2}N \vee \epsilon}\bigg[\sqrt{\prescript{}{2}D+\log(\prescript{}{2}N(T/\epsilon\vee 1))}+ z \bigg]
\end{align}

with probability at least $1-e^{-z^2}$. 
\end{lemma}
\begin{proof}
We consider $t\in \mathbb{N}\epsilon \cap [0,T]$ rather than $\ell\in [0,T/\epsilon]\cap \mathbb{N}$. If we take the difference in (\ref{newnonlineardynamics}) and (\ref{newSGD}), we have

\begin{align}
\|\prescript{}{2}{\bar{\bm \theta}}_i^t-\prescript{}{2}{\bm \theta}_i^{t/\epsilon}\|_2&=2\bigg\|\int_0^t \xi(s)\prescript{}{2}{\bf G}({\bar{\bm \theta}}_i^s;\prescript{}{2}\rho_s)ds - \epsilon\sum_{\ell=0}^{t/\epsilon-1} \xi(\ell \epsilon)\prescript{}{2}{\bf F}_i({{\bm \theta}}_i^\ell ; {\bm \beta}_{\ell+1})ds\bigg\|_2 \\
&\leq 2\int_0^t \bigg\| \xi(s)\prescript{}{2}{\bf G}({\bar{\bm \theta}}_i^s;\prescript{}{2}\rho_s)-\xi([s])\prescript{}{2}{\bf G}({\bar{\bm \theta}}_i^{[s]};\prescript{}{2}\rho_{[s]})\bigg\|_2 ds \nonumber \\
& \hspace{1cm} + 2\int_0^t \bigg\|\xi([s])\prescript{}{2}{\bf G}({\bar{\bm \theta}}_i^{[s]};\prescript{}{2}\rho_{[s]})-\xi([s])\prescript{}{2}{\bf G}({{\bm \theta}}_i^{\lfloor s/\epsilon\rfloor};\prescript{}{2}\rho_{[s]})\bigg\|_2 ds \nonumber \\
&\hspace{1cm}+ 2\bigg\| \epsilon \sum_{\ell=0}^{t/\epsilon-1} \xi(k\epsilon) \big\{\prescript{}{2}{\bf F}_i ({{\bm \theta}}_i^\ell ; {\bm \beta}_{\ell+1})-\prescript{}{2}{\bf G}({{\bm \theta}}_i^\ell;\prescript{}{2}\rho_{\ell \epsilon})\big\} \bigg\|_2 \\
&\equiv 2Q_1^i(t)+2Q_2^i(t)+2Q_3^i(t). \label{Qbound}
\end{align}

For the first term $Q_1^i(t)$, we have 

\begin{align}
Q_1^i(t) &\leq t \sup_{s\in [0,t]}\bigg\{ \|\xi(s)\prescript{}{2}{\bf G}({\bar{\bm \theta}}_i^s;\prescript{}{2}\rho_s)-\xi([s])\prescript{}{2}{\bf G}({\bar{\bm \theta}}_i^s;\prescript{}{2}\rho_s)\|_2 \nonumber \\
&\hspace{1cm} +\|\xi([s])\prescript{}{2}{\bf G}({\bar{\bm \theta}}_i^s;\prescript{}{2}\rho_s)-\xi([s])\prescript{}{2}{\bf G}({\bar{\bm \theta}}_i^{[s]};\prescript{}{2}\rho_s)\|_2 \nonumber \\
&\hspace{1cm} + \|\xi([s])\prescript{}{2}{\bf G}({\bar{\bm \theta}}_i^{[s]};\prescript{}{2}\rho_s)-\xi([s])\prescript{}{2}{\bf G}({\bar{\bm \theta}}_i^{[s]};\prescript{}{2}\rho_{[s]})\|_2 \bigg\} \nonumber \\
&\leq Kt\epsilon \label{Q1bound}
\end{align}

\noindent since $\prescript{}{2}{\bf G}({\bm \theta};\prescript{}{2}\rho)$ is Lipschitz continuous with respect to ${\bm \theta}$ and $\prescript{}{2}\rho$ from (\ref{GLip}) and by Assumption \ref{assump1} and Lemma \ref{Lemma1Lip} (together implying that $\xi$, $\prescript{}{2}\rho_s$, and $\prescript{}{2}{\bar{\bm \theta}}_i^t$ are Lipschitz continuous).

To bound $Q_2^i(t)$, we use Assumption \ref{assump1} and the Lipschitz continuity of $\prescript{}{2}{\bf G}$ with respect to ${\bm \theta}$:

\begin{align}
Q_2^i(t) \leq K \int_0^t \|\prescript{}{2}{\bf G}({\bar{\bm \theta}}_i^{[s]};\prescript{}{2}\rho_{[s]})-\prescript{}{2}{\bf G}({{\bm \theta}}_i^{\lfloor s/\epsilon \rfloor};\prescript{}{2}\rho_{[s]})\|_2 ds \leq K^2 \int_0^t \|\prescript{}{2}{\bar{\bm \theta}}_i^{[s]}-\prescript{}{2}{{\bm \theta}}_i^{\lfloor s/\epsilon \rfloor} \|_2 ds. \label{Q2bound}
\end{align}

To bound $Q_3^i(t)$, we introduce $\mathcal{F}_\ell$, the sigma-algebra generated by $(\prescript{}{2}{{\bm \theta}}_i^0)_{i\leq N}$ and ${\bm \beta}_1,\dots,{\bm \beta}_\ell$ for $\ell\in \mathbb{N}$. We see that

\begin{align}
\mathbb{E}\big\{ \prescript{}{2}{\bf F}_i({{\bm \theta}}_i^\ell;{\bm \beta}_{\ell+1})|\mathcal{F}_\ell\big\}=-\nabla_{\prescript{}{2}{\bm \theta}_i} V(\prescript{}{1}{\bm  \theta}_i^\ell,\prescript{}{2}{\bm  \theta}_i^\ell)-\frac{1}{\prescript{}{2}N}\sum_{j=1}^{\prescript{}{2}N} \nabla_{\prescript{}{2}{\bm \theta}_i} U(\prescript{}{1}{\bm  \theta}_i^\ell,\prescript{}{2}{\bm  \theta}_i^\ell,\prescript{}{2}{\bm  \theta}_j^\ell )=\prescript{}{2}{\bf G}({\bm  \theta}_i^\ell;\prescript{}{2}{\hat \rho}_\ell^{(\prescript{}{2}N)}),
\end{align}

\noindent where $\prescript{}{2}{\hat \rho}_\ell^{(\prescript{}{2}N)}\equiv \frac{1}{\prescript{}{2}N}\sum_{i\leq \prescript{}{2}N}\delta_{\prescript{}{2} {\bm \theta}_i^\ell}$. Therefore,

\begin{align}
Q_3^i(t) &=\bigg\| \epsilon \sum_{\ell=0}^{t/\epsilon-1} \xi(k\epsilon) \big\{\prescript{}{2}{\bf F}_i ({{\bm \theta}}_i^\ell ; {\bm \beta}_{\ell+1})-\prescript{}{2}{\bf G}({{\bm \theta}}_i^\ell;\prescript{}{2}\rho_{\ell \epsilon})\big\} \bigg\|_2 \\
&\leq \bigg\|\epsilon \sum_{\ell=0}^{t/\epsilon-1} \xi(\ell \epsilon) \big\{\prescript{}{2}{\bf G}({\bm \theta}_i^\ell;\prescript{}{2}{\hat \rho}_\ell^{(\prescript{}{2}N)})-\prescript{}{2}{\bf G}({\bm \theta}_i^\ell;\prescript{}{2}\rho_{\ell \epsilon})\big\} \bigg\|_2 + \bigg\|\epsilon \sum_{\ell=0}^{t/\epsilon-1} \xi(\ell \epsilon) \prescript{}{2}{\bf D}_i^\ell \bigg\|_2 \nonumber\\
&\equiv Q_{3,0}^i(t)+O_{3,1}^i(t), \label{q3}
\end{align}

\noindent with the $\prescript{}{2}{\bf D}^\ell_i\equiv \prescript{}{2}{\bf F}_i({\bm \theta}_i^\ell ; {\bm \beta}_{\ell+1})-\mathbb{E}\{\prescript{}{2}{\bf F}_i({\bm \theta}_i^\ell ; {\bm \beta}_{\ell+1})|\mathcal{F}_\ell \}$ being martingale differences. Since a sum of martingale differences is martingale, using the Azuma-Hoeffding inequality (Lemma A.1 in \cite{MMN2018}), we see that since each $\xi(\ell \epsilon)\prescript{}{2}{\bf D}_i^\ell$ is $K^2$-sub-Gaussian due to $\prescript{}{2}\sigma_*(\cdot;\prescript{}{2}{\bm \theta})$ being bounded and $\nabla_{\prescript{}{2}{\bm \theta}} \prescript{}{2}\sigma_*(\cdot;\prescript{}{2}{\bm \theta})$ being sub-Gaussian, we have
%it's \sqrt{t\epsilon} due to ell \in [0,t/\epsilon] and multiplying the sum by \epsilon, which is of order \ell\epsilon. same with 1/prescript{}{2}N for second union bound. Basically length of interval we are considering.
\begin{align}
\mathbb{P}\bigg(\max_{\ell \in [0,t/\epsilon]\cap \mathbb{N}} O_{3,1}^i(\ell\epsilon)\geq K \sqrt{t\epsilon}\big(\sqrt{\prescript{}{2}D}+u\big)\bigg) \leq e^{-u^2}.
\end{align}

\noindent Taking the union bound over $i\leq \prescript{}{2}N$ gives us

\begin{align}
&\mathbb{P}\bigg(\max_{i\leq \prescript{}{2}N} \max_{\ell \in [0,t/\epsilon]\cap \mathbb{N}} O_{3,1}^i(\ell\epsilon)\geq K \sqrt{t\epsilon}\big(\sqrt{\prescript{}{2}D}+u\big)\bigg) \leq \prescript{}{2}Ne^{-u^2}=e^{-u^2+\log\prescript{}{2}N} \\
&\hspace{2cm}\text{setting }-z^2=-u^2+\log\prescript{}{2}N\Rightarrow u=\sqrt{z^2+\log\prescript{}{2}N} \text{ gives} \nonumber \\
&\mathbb{P}\bigg(\max_{i\leq \prescript{}{2}N} \max_{\ell \in [0,t/\epsilon]\cap \mathbb{N}} O_{3,1}^i(\ell\epsilon)\geq K \sqrt{t\epsilon}\big(\sqrt{\prescript{}{2}D}+\sqrt{z^2+\log\prescript{}{2}N}\big)\bigg) \leq e^{-z^2} \\
&\Rightarrow \mathbb{P}\bigg(\max_{i\leq \prescript{}{2}N} \max_{\ell \in [0,t/\epsilon]\cap \mathbb{N}} O_{3,1}^i(\ell\epsilon)\geq K \sqrt{t\epsilon}\big(\sqrt{\prescript{}{2}D+\log\prescript{}{2}N}+z\big)\bigg) \leq e^{-z^2}\\
&\hspace{2cm} \text{since }\sqrt{\prescript{}{2}D}+\sqrt{z^2+\log\prescript{}{2}N}\geq \sqrt{\prescript{}{2}D+\log\prescript{}{2}N}+z \text{ for }\prescript{}{2}D \geq z^2 \nonumber\\
&\Rightarrow \mathbb{P}\bigg(\max_{i\leq \prescript{}{2}N}\max_{\ell \in [0,t/\epsilon]\cap \mathbb{N}} O_{3,1}^i(\ell \epsilon) \leq K \sqrt{t \epsilon}\big(\sqrt{\prescript{}{2}D+\log \prescript{}{2}N}+z\big) \bigg) \geq 1-e^{-z^2}, \label{o31}
\end{align}

\noindent For $Q_{3,0}^i(t)$, we again utilize (\ref{GLip}), obtaining

\begin{align}
\|\prescript{}{2}{\bf G}({\bm \theta}_i^\ell \prescript{}{2}{\hat \rho}_\ell^{(\prescript{}{2}N)})-\prescript{}{2}{\bf G}({\bm \theta}_i^\ell ;\prescript{}{2}\rho_{\ell \epsilon})\|_2 &\leq\bigg\|\frac{1}{\prescript{}{2}N}\sum_{j=1}^{\prescript{}{2}N}\big[\nabla_{\prescript{}{2}{\bm \theta}_i} U(\prescript{}{1}{\bm \theta}_i^\ell,\prescript{}{2}{\bm \theta}_i^\ell,\prescript{}{2}{\bm \theta}_j^\ell)-\nabla_{\prescript{}{2}{\bm \theta}_i} U(\prescript{}{1}{\bm \theta}_i^\ell,\prescript{}{2}{\bm \theta}_i^\ell,\prescript{}{1}{\bar{\bm \theta}}_j^{\ell\epsilon})\big]\bigg\|_2 \nonumber \\
&\hspace{1cm} +\bigg\| \frac{1}{\prescript{}{2}N}\sum_{j=1}^{\prescript{}{2}N}\big[\nabla_{\prescript{}{2}{\bm \theta}_i} U(\prescript{}{1}{\bm \theta}_i^\ell,\prescript{}{2}{\bm \theta}_i^\ell,\prescript{}{2}{\bar{\bm \theta}}_j^{\ell \epsilon})-\mathbb{E}_{\prescript{}{2}{\bar {\bm \theta}}} \nabla_{\prescript{}{2}{\bm \theta}_i} U(\prescript{}{1}{\bm \theta}_i^\ell,\prescript{}{2}{\bm \theta}_i^\ell,\prescript{}{2}{\bar{\bm \theta}}_j^{\ell\epsilon})\big]\bigg\|_2 \nonumber \\
&\leq \frac{K}{\prescript{}{2}N} \sum_{j=1}^{\prescript{}{2}N}\|\prescript{}{2}{\bm \theta}_j^\ell-\prescript{}{2}{\bar{\bm \theta}}_j^{\ell\epsilon}\|_2 +\bigg[O_{3,2}^i(\ell \epsilon)+\frac{K}{\prescript{}{2}N}\bigg],
\end{align}

\noindent where

\begin{align}
O_{3,2}^i(\ell \epsilon)\equiv \bigg\| \frac{1}{\prescript{}{2}N}\sum_{j\leq \prescript{}{2}N,j\not=i}\big[\nabla_{\prescript{}{2}{\bm \theta}_i} U(\prescript{}{1}{\bm \theta}_i^\ell,\prescript{}{2}{\bm \theta}_i^\ell,\prescript{}{2}{\bar{\bm \theta}}_j^{\ell \epsilon})-\mathbb{E}_{\prescript{}{2}{\bar {\bm \theta}}} \nabla_{\prescript{}{2}{\bm \theta}_i} U(\prescript{}{1}{\bm \theta}_i^\ell,\prescript{}{2}{\bm \theta}_i^\ell,\prescript{}{2}{\bar{\bm \theta}}_j^{\ell\epsilon})\big]\bigg\|_2
\end{align}

\noindent for $\ell\in \mathbb{N}$. For any fixed $\ell$, $(\prescript{}{2}{\bar{\bm \theta}_j^{\ell \epsilon}})_{j \leq \prescript{}{2}N,j\not=i}$ are i.i.d. and independent of $\prescript{}{2}{\bm \theta}_i^\ell$, and $\nabla_{\prescript{}{2}{\bm \theta}_i} U$ is bounded. Then, by applying the Azuma-Hoeffding inequality (specifically (A.7) of Lemma A.1 in \cite{MMN2018} since we are dealing with a fixed $\ell$), 

\begin{align}
\mathbb{P}\bigg(O_{3,2}^i(\ell \epsilon) \geq K\sqrt{1/\prescript{}{2}N}(\sqrt{\prescript{}{2}D}+u)\bigg)\leq e^{-u^2}.
\end{align}

\noindent Taking the union bound over $\ell \in [0,t/\epsilon]\cap \mathbb{N}$ and $i\leq \prescript{}{2}N$ leaves us with

\begin{align}
\mathbb{P}\bigg(\max_{i \leq \prescript{}{2}N}\max_{\ell \in [0,t/\epsilon]\cap \mathbb{N}} O_{3,2}^i(\ell\epsilon)\leq K\sqrt{1/\prescript{}{2}N}\big(\sqrt{\prescript{}{2}D+\log(\prescript{}{2}N(t/\epsilon \vee 1))}+z\big)\bigg)\geq 1-e^{-z^2}, \label{o32}
\end{align}

\noindent Conditioning on the good events in (\ref{o31}) and (\ref{o32}), (\ref{q3}) becomes

\begingroup
\allowdisplaybreaks
\begin{align}
Q_3^i(t)&\leq Q_{3,0}^i(t)+O_{3,1}^i(t) \\
&\leq \bigg[\sum_{\ell=0}^{t/\epsilon-1}\epsilon \xi(\ell \epsilon)\big\{\frac{K}{\prescript{}{2}N}\sum_{j=1}^{\prescript{}{2}N} \| \prescript{}{2}{\bm \theta}_j^\ell-\prescript{}{2}{\bar{\bm \theta}}_j^{\ell\epsilon}\|_2 +O_{3,2}^i(\ell \epsilon)+\frac{K}{\prescript{}{2}N} \big\} \bigg] +\bigg[ \max_{i\leq \prescript{}{2}N}\max_{\ell \in [0,t/\epsilon]\cap \mathbb{N}} O_{3,1}^i(\ell \epsilon) \bigg]\\
&=\bigg[\int_{s=0}^t\big\{\frac{K}{\prescript{}{2}N}\sum_{j=1}^{\prescript{}{2}N} \| \prescript{}{2}{\bm \theta}_j^{\lfloor s/\epsilon\rfloor}-\prescript{}{2}{\bar{\bm \theta}}_j^{[s]}\|_2 +O_{3,2}^i([s])+\frac{K}{\prescript{}{2}N} \big\}ds \bigg] +\bigg[ \max_{i\leq \prescript{}{2}N}\max_{\ell \in [0,t/\epsilon]\cap \mathbb{N}} O_{3,1}^i(\ell \epsilon) \bigg] \\
&\leq \bigg[\frac{K}{\prescript{}{2}N}\sum_{j=1}^{\prescript{}{2}N}\int_0^t \| \prescript{}{2}{\bm \theta}_j^{\lfloor s/\epsilon\rfloor}-\prescript{}{2}{\bar{\bm \theta}}_j^{[s]}\|_2 ds +t*\max_{i\leq \prescript{}{2}N}\max_{\ell \in [0,t/\epsilon]\cap \mathbb{N}} O_{3,2}^i(\ell \epsilon)+\frac{tK}{\prescript{}{2}N} \bigg]+\bigg[ \max_{i\leq \prescript{}{2}N}\max_{\ell \in [0,t/\epsilon]\cap \mathbb{N}} O_{3,1}^i(\ell \epsilon) \bigg] \\
&=\frac{K}{\prescript{}{2}N}\sum_{j=1}^{\prescript{}{2}N}\int_0^t \| \prescript{}{2}{\bm \theta}_j^{\lfloor s/\epsilon\rfloor}-\prescript{}{2}{\bar{\bm \theta}}_j^{[s]}\|_2 ds +\frac{tK}{\prescript{}{2}N} +t*\max_{i\leq \prescript{}{2}N}\max_{\ell \in [0,t/\epsilon]\cap \mathbb{N}} O_{3,2}^i(\ell \epsilon)+ \max_{i\leq \prescript{}{2}N}\max_{\ell \in [0,t/\epsilon]\cap \mathbb{N}} O_{3,1}^i(\ell \epsilon)\\
&\leq \frac{K}{\prescript{}{2}N}\sum_{j=1}^{\prescript{}{2}N}\int_0^t \| \prescript{}{2}{\bm \theta}_j^{\lfloor s/\epsilon\rfloor}-\prescript{}{2}{\bar{\bm \theta}}_j^{[s]}\|_2 ds +\frac{tK}{\prescript{}{2}N} \nonumber\\
&\hspace{1cm} +Kt\sqrt{1/\prescript{}{2}N}\big(\sqrt{\prescript{}{2}D+\log(\prescript{}{2}N(t/\epsilon \vee 1))}+z\big)+K\sqrt{t \epsilon}\big(\sqrt{\prescript{}{2}D+\log\prescript{}{2}N}+z\big)\nonumber \\
&\leq \frac{K}{\prescript{}{2}N}\sum_{j=1}^{\prescript{}{2}N}\int_0^t \| \prescript{}{2}{\bm \theta}_j^{\lfloor s/\epsilon\rfloor}-\prescript{}{2}{\bar{\bm \theta}}_j^{[s]}\|_2 ds +\frac{tK}{\prescript{}{2}N}+K(\sqrt{t}\vee t)\sqrt{1/\prescript{}{2}N\vee \epsilon}\big(\sqrt{\prescript{}{2}D+\log(\prescript{}{2}N(t/\epsilon \vee 1))}+z \big) \\
&\equiv \frac{K}{\prescript{}{2}N}\sum_{j=1}^{\prescript{}{2}N}\int_0^t \| \prescript{}{2}{\bm \theta}_j^{\lfloor s/\epsilon\rfloor}-\prescript{}{2}{\bar{\bm \theta}}_j^{[s]}\|_2 ds +\frac{tK}{\prescript{}{2}N} +O_3(t) \label{Q3bound}
\end{align}
\endgroup

\noindent with probability at least $1-e^{-z^2}$. 

We were interested in bounding $\Delta(t;\prescript{}{2}N,\epsilon)\equiv\max_{i\leq \prescript{}{2}N} \sup_{\ell \in [0,t/\epsilon]\cap\mathbb{N}}\|\prescript{}{2}{\bm\theta}_i^\ell - \prescript{}{2} {\bar{\bm \theta}}_i^{\ell\epsilon}\|_2$. From (\ref{Qbound}), we had $\|\prescript{}{2}{\bar{\bm \theta}}_i^t-\prescript{}{2}{\bm \theta}_i^{t/\epsilon}\|_2\leq 2Q_1^i(t)+2Q_2^i(t)+2Q_3^i(t)$. Using (\ref{Q1bound}), (\ref{Q2bound}), and (\ref{Q3bound}), we have

\begin{align}
\Delta(t;\prescript{}{2}N,\epsilon) &\leq \max_{i\leq \prescript{}{2}N} \sup_{\ell \in [0,t/\epsilon]\cap\mathbb{N}} \bigg[\big\{Kt\epsilon\big\}+\big\{K^2 \int_0^t \|\prescript{}{2}{\bar{\bm \theta}}_i^{[s]}-\prescript{}{2}{{\bm \theta}}_i^{\lfloor s/\epsilon \rfloor} \|_2 ds\big\} \nonumber \\ 
&\hspace{1cm}+ \big\{ \frac{K}{\prescript{}{2}N}\sum_{j=1}^{\prescript{}{2}N}\int_0^t \| \prescript{}{2}{\bm \theta}_j^{\lfloor s/\epsilon\rfloor}-\prescript{}{2}{\bar{\bm \theta}}_j^{[s]}\|_2 ds +\frac{tK}{\prescript{}{2}N} +O_3(t) \big\} \bigg]\\
&\leq \max_{i\leq \prescript{}{2}N} \sup_{\ell \in [0,t/\epsilon]\cap\mathbb{N}} \bigg[K^2 \int_0^t \|\prescript{}{2}{\bar{\bm \theta}}_i^{[s]}-\prescript{}{2}{{\bm \theta}}_i^{\lfloor s/\epsilon \rfloor} \|_2 ds+\frac{K}{\prescript{}{2}N}\sum_{j=1}^{\prescript{}{2}N}\int_0^t \| \prescript{}{2}{\bm \theta}_j^{\lfloor s/\epsilon\rfloor}-\prescript{}{2}{\bar{\bm \theta}}_j^{[s]}\|_2 ds\bigg] \nonumber\\
&\hspace{1cm}+ Kt\epsilon+\frac{tK}{\prescript{}{2}N}+O_3(t) \\
&\leq K\int_0^t \Delta(s;\prescript{}{2}N,\epsilon)+Kt\epsilon+\frac{Kt}{\prescript{}{2}N}+O_3(t)
\end{align}

\noindent We utilize Gronwall's Inequality to further bound $\Delta(t;\prescript{}{2}N,\epsilon)$. Gronwall's Inequality states that if $\beta$ is non-negative, $\alpha$ is nondecreasing, and $u$ satisfies $u(t)\leq \alpha(t)+\int_0^t \beta(s)u(s)ds$, then $u(t)\leq \alpha(t)\exp\big(\int_0^t \beta(s)ds\big)$. In our case, we have $\beta(t)=K$, $\alpha(t)=Kt\epsilon+\frac{Kt}{\prescript{}{2}N}+O_3(t)$, and $u(t)=\Delta(t;\prescript{}{2}N,\epsilon)$. Thus, Gronwall's gives us

\begin{align}
\Delta(t;\prescript{}{2}N,\epsilon) &\leq \bigg(Kt\epsilon+\frac{Kt}{\prescript{}{2}N}+O_3(t) \bigg)\exp\big\{ \int_0^t K ds \big\}=\bigg(Kt\epsilon+\frac{Kt}{\prescript{}{2}N}+O_3(t) \bigg)e^{Kt} \\
&\leq e^{Kt} \bigg(Kt\epsilon+\frac{Kt}{\prescript{}{2}N}+K(\sqrt{t}\vee t)\sqrt{1/\prescript{}{2}N\vee \epsilon}\big(\sqrt{\prescript{}{2}D+\log(\prescript{}{2}N(t/\epsilon \vee 1))}+z \big) \bigg) \\
&\leq Kte^{Kt} \bigg(\epsilon+\frac{1}{\prescript{}{2}N}+\sqrt{1/\prescript{}{2}N\vee \epsilon}\big(\sqrt{\prescript{}{2}D+\log(\prescript{}{2}N(t/\epsilon \vee 1))}+z \big) \bigg) 
%&\to Kte^{Kt} \bigg(\sqrt{1/\prescript{}{2}N\vee \epsilon}\big(\sqrt{\prescript{}{2}D+\log(\prescript{}{2}N(t/\epsilon \vee 1))}+z \big) \bigg) 
\end{align}

\noindent as required.

\end{proof}

By bounding the difference in the nonlinear dynamics and original dynamics in Lemma \ref{Lemma1Lip}, coupled with the observation in (\ref{nonlineardynamicsrhot}), we see that as a result, for any sequence $(\prescript{}{2}N,\epsilon=\epsilon_{\prescript{}{2}N})$ s.t. $\prescript{}{2}N \to \infty$ and $\epsilon_{\prescript{}{2}N} \to 0$ with $\prescript{}{2}N/\log(\prescript{}{2}N/\epsilon_{\prescript{}{2}N})\to \infty$ and $\epsilon_{\prescript{}{2}N}\log(\prescript{}{2}N/\epsilon_{\prescript{}{2}N}) \to 0$, we have $\prescript{}{2}{\hat\rho}_{\lfloor t/\epsilon \rfloor}^{(\prescript{}{2}N)}$ converges weakly to $\prescript{}{2}\rho_t$ almost surely.

\end{proof}

We introduce another assumption that will help to bound terms in the first hidden layer and provide a similar result to Theorem \ref{layer2convergence} in the first hidden layer.

\begin{assump}\label{assump3}
The gradients $\prescript{}{1}{\bm \theta} \mapsto \nabla_{\prescript{}{1} {\bm \theta}} V(\prescript{}{1}{\bm \theta},\prescript{}{2} {\bm \theta})$ and $(\prescript{}{1} {\bm \theta})\mapsto \nabla_{\prescript{}{1} {\bm \theta}} U(\prescript{}{1} {\bm \theta},\prescript{}{2} {\bm \theta},\prescript{}{2} {\bm \theta}') $ are bounded, Lipschitz continuous, i.e. we have that $\|\nabla_{\prescript{}{1} {\bm \theta}} V(\prescript{}{1}{\bm \theta},\prescript{}{2} {\bm \theta})\|_2,\|\nabla_{\prescript{}{1} {\bm \theta}} U(\prescript{}{1} {\bm \theta},\prescript{}{2} {\bm \theta},\prescript{}{2} {\bm \theta}') \|_2 \leq K_4$, $\| \nabla_{\prescript{}{1} {\bm \theta}} V(\prescript{}{1}{\bm \theta},\prescript{}{2} {\bm \theta}) - \nabla_{\prescript{}{1} {\bm \theta}} V(\prescript{}{1}{\bm \theta}',\prescript{}{2} {\bm \theta})\|_2\leq K_4 \|\prescript{}{1} {\bm \theta}-\prescript{}{1}{\bm \theta}'\|_2$, and $\| \nabla_{\prescript{}{1} {\bm \theta}} U(\prescript{}{1} {\bm \theta},\prescript{}{2} {\bm \theta},\prescript{}{2} {\bm \theta}') -\nabla_{\prescript{}{1} {\bm \theta}} U(\prescript{}{1} {\bm \theta}',\prescript{}{2} {\bm \theta},\prescript{}{2} {\bm \theta}') \|_2 \leq K_4\|\prescript{}{1}{\bm \theta}-\prescript{}{1}{\bm \theta}' \|_2$.
\end{assump}

%%%%%%%%%%%%%%%%%%
\begin{thm} \label{layer1convergence}
Assume Assumptions \ref{assump1}, \ref{assump2}, \ref{assump4}, and \ref{assump3} hold. For $\prescript{}{i} \rho_0 \in P(\mathbb{R}^{\prescript{}{i}D})$ for $i=1,2$, consider SGD with initialization $(\prescript{}{i} {\bm \theta}_j^0)_{j \leq {\prescript{}{i} N}} \sim_{i.i.d.} \prescript{}{i} \rho_0$ and step size $s_k = \epsilon \xi(k\epsilon)$. For $t\geq 0$, let $\prescript{}{1}\rho_t$ be the solution of the PDE system from (\ref{PDE1Layera}) and (\ref{PDE1Layerb}). Then, for any fixed $t\geq 0$, $\prescript{}{1}{\hat \rho}_{\lfloor t/\epsilon \rfloor}^{\prescript{}{1} N} \Rightarrow \prescript{}{1} \rho_t$ almost surely along any sequence $(\prescript{}{1}N, \epsilon=\epsilon_{\prescript{}{1}N})$ s.t. $N\to \infty$, $\epsilon_{\prescript{}{1}N}\to 0$, $\prescript{}{1}N/\log (\prescript{}{1} N /\epsilon_{\prescript{}{1}N}) \to \infty$ and $\epsilon_{\prescript{}{1} N} \log(\prescript{}{1} N/\epsilon_{\prescript{}{1}N})\to 0$. 
\end{thm}

\begin{proof}
This proof will follow similarly to the proof for Theorem \ref{layer2convergence}. For this proof, let $K$ denote a generic constant depending on the constants $K_1$, $K_2$, and $K_4$ from Assumptions \ref{assump1}, \ref{assump2}, and \ref{assump3}. Letting ${\bm \beta}_k = ({\bf x}_k,y_k)$ denote the $k$th example, we define

\begin{align}
\prescript{}{1}{\bf F}_i({\bm \theta};{\bm \beta}_k)&=\big(y_k-{\hat y}({\bf x}_k;{\bm \theta})\big)\nabla_{\prescript{}{1}{\bm \theta}_i} \prescript{}{1}\sigma_*({\bf x}_k,\prescript{}{1}{\bm \theta}), \hspace{2cm} \prescript{}{1}{\bm \theta}=(\prescript{}{1}{\bm \theta}_i)_{i\leq \prescript{}{1}N} \in \mathbb{R}^{\prescript{}{1} D\times \prescript{}{1}N} \\
\prescript{}{1}{\bf G}({\bm \theta};\prescript{}{2}\rho) &= - \nabla_{\prescript{}{1}{\bm \theta}} \Psi_1({\bm \theta},\prescript{}{2}\rho)=-\nabla_{\prescript{}{1}{\bm \theta}} V(\prescript{}{1}{\bm \theta},\prescript{}{2}{\bm \theta})-\int \nabla_{\prescript{}{1}{\bm \theta}} U(\prescript{}{1}{\bm \theta},\prescript{}{2}{\bm \theta},\prescript{}{2}{\bm \theta}')\prescript{}{2}\rho(d\prescript{}{2}{\bm \theta}'), \hspace{1cm} \prescript{}{1}{\bm \theta} \in \mathbb{R}^{\prescript{}{1}D}.
\end{align}

By the assumption of bounded Lipschitz for $\nabla_{\prescript{}{1}{\bm \theta}} V$ and $\nabla_{\prescript{}{1}{\bm \theta}} U$, we have that $\|\prescript{}{1}{\bf G}({\bm \theta};\prescript{}{2}\rho)\|_2 \leq K$ and $\|\prescript{}{1}{\bf G}({\bm \theta}_1;\prescript{}{2}\rho)-\prescript{}{1}{\bf G}({\bm \theta}_2;\prescript{}{2}\rho)\|_2 \leq K\|{\bm \theta}_1-{\bm \theta}_2\|_2$. In addition,

\begin{align}
\|\prescript{}{1}{\bf G}({\bm \theta};\prescript{}{2}\rho_1)-\prescript{}{1}{\bf G}({\bm \theta};\prescript{}{2}\rho_2)\|_2=\bigg\|\int \nabla_{\prescript{}{1}{\bm \theta}} U(\prescript{}{1}{\bm \theta},\prescript{}{2}{\bm \theta},\prescript{}{2}{\bm \theta}' ) (\prescript{}{2}\rho_1-\prescript{}{2}\rho_2)(d\prescript{}{2}{\bm \theta}')\bigg\|_2 \leq K d_{\text{BL}}(\prescript{}{2}\rho_1,\prescript{}{2}\rho_2), \label{GLip1}
\end{align} 

We can now rewrite the SGD dynamics in the first layer as

\begin{align}
\prescript{}{1}{\bm \theta}_i^{k+1}&=\prescript{}{1}{\bm \theta}_i^{k}+2\epsilon \xi(k\epsilon)\prescript{}{1}{\bf F}_i({\bm \theta}_i^k,{\bm \beta}_{k+1}) \\
\Rightarrow \prescript{}{1}{\bm \theta}_i^{k}&=\prescript{}{1}{\bm \theta}_i^{0}+2\epsilon \sum_{\ell=0}^{k-1} \xi(\ell\epsilon)\prescript{}{1}{\bf F}_i({\bm \theta}_i^\ell,{\bm \beta}_{\ell+1}). \label{newSGD1}
\end{align}

In order to define nonlinear dynamics to analyze the PDE system for the first layer, we introduce trajectories $(\prescript{}{1}{\bar{\bm \theta}}_i^t,\prescript{}{2}{\bar{\bm \theta}}_i^t)_{1,\leq i, \leq \prescript{}{1}N,1\leq j \leq \prescript{}{2}N,t\in \mathbb{R}_{\geq 0}}$ and letting ${\bar{\bm \theta}}_i^0={\bm \theta}_i^0$ be the same initialization as for the SGD. Then, for $t\geq 0$, the nonlinear dynamics follow

\begin{align}
\prescript{}{1}{\bar{\bm \theta}}_i^t&=\prescript{}{1}{\bar{\bm \theta}}_i^0-2\int_0^t \xi(s) \nabla_{\prescript{}{1}{\bar{\bm \theta}}}\Psi_1({\bar{\bm \theta}}_i^s,\prescript{}{2}\rho_s)ds \label{nonlineardynamics1}\\
\prescript{}{1}\rho_s&=P_{\prescript{}{1}{\bar{\bm \theta}}_i^s}, 
\end{align}

\noindent From \cite{Szn91}, under Assumptions \ref{assump1} and \ref{assump3}, the nonlinear dynamics has a unique solution, with $\prescript{}{1}\rho_t$ satisfying (\ref{PDE1Layera}). We notice that under the nonlinear dynamics, the trajectories $(\prescript{}{1}{\bar{\bm \theta}}_i^t)_{1\leq i \leq \prescript{}{1}N, t\in \mathbb{R}_{\geq 0}}$ are i.i.d., which implies that 

\begin{align}
\frac{1}{\prescript{}{1}N}\sum_{i=1}^{\prescript{}{1}N} \delta_{\prescript{}{1}{\bar{\bm \theta}}_i^t } \Rightarrow \prescript{}{1}\rho_t \label{nonlineardynamicsrhot1}
\end{align}

\noindent in distribution. We rewrite (\ref{nonlineardynamics1}) as 

\begin{align}
\prescript{}{1}{\bar{\bm \theta}}_i^t&=\prescript{}{1}{\bar{\bm \theta}}_i^0+2\int_0^t \xi(s) \prescript{}{1}{\bf G}({\bar{\bm \theta}}_i^s;\prescript{}{2}\rho_s)ds \label{newnonlineardynamics1}
\end{align}

\noindent and compare it to (\ref{newSGD1}). We will define $[t]=\epsilon\lfloor t/\epsilon\rfloor$ for $t\in \mathbb{R}_{\geq 0}$. We next detail some Lemmas that bound the difference between the nonlinear dynamics and original dynamics.

\begin{lemma}\label{Lemma1Lip1}
Assume Assumptions \ref{assump1} and \ref{assump3} hold. Let $(\prescript{}{1}\rho_t)_{t\geq0}$ be the solution of (\ref{PDE1Layera}). Let $(\prescript{}{1}{\bar{\bm \theta}}_i^t)_{t\geq 0}$ be the solution of (\ref{nonlineardynamics1}). Then $t\mapsto \prescript{}{1}{\bar{\bm \theta}}_i^t$ is $K_1K_4$-Lipschitz continuous, and $t\mapsto \prescript{}{1}\rho_t$ is $K_1K_4$-Lipschitz continuous in $W_2$ Wasserstein distance. 
\end{lemma}

\begin{proof}
Since $\xi$ is $K_1$ bounded and $\nabla_{\prescript{}{1}{{\bm \theta}}}\Psi_1$ is $K_4$ bounded, $t\mapsto \prescript{}{1}{\bar{\bm \theta}}_i^t$ is $K_1K_4$-Lipschitz continuous. In addition, since 

$$d_{\text{BL}}(\prescript{}{1}\rho_t,\prescript{}{1}\rho_s) \leq W_2(\prescript{}{1}\rho_t,\prescript{}{1}\rho_s) \leq  (\mathbb{E}[\|\prescript{}{1}{\bar{\bm \theta}}_i^t-\prescript{}{1}{\bar{\bm \theta}}_i^s\|_2^2])^{1/2} \leq K_1K_4|t-s|,$$

$t\mapsto \prescript{}{1}\rho_t$ is Lipschitz continuous in $W_2$ Wasserstein distance.
\end{proof}

\begin{lemma}\label{Lemma2Lip1}
Under Assumptions \ref{assump1}, \ref{assump2}, \ref{assump4}, \ref{assump3}, $\exists$ constant $K$, dependent on $K_1$, $K_2$, and $K_4$, s.t. for any $T\geq 0$, 

\begin{align}
\max_{i\leq \prescript{}{1}N} \sup_{\ell\in [0,T/\epsilon]\cap \mathbb{N}} \|\prescript{}{1}{\bm \theta}_i^\ell -\prescript{}{1}{\bar{\bm \theta}}_i^{\ell\epsilon} \|_2
\leq KTe^{KT} \bigg[\epsilon+\frac{1}{\prescript{}{1}N}+\sqrt{1/\prescript{}{1}N\vee \epsilon}\big(\sqrt{\prescript{}{1}D+\log(\prescript{}{1}N(t/\epsilon \vee 1))}+z \big) \bigg]
%\leq K e^{KT}\sqrt{1/\prescript{}{1}N \vee \epsilon}\bigg[\sqrt{\prescript{}{1}D+\log(\prescript{}{1}N(T/\epsilon\vee 1))}+ z \bigg]
\end{align}

with probability at least $1-e^{-z^2}$. 
\end{lemma}
\begin{proof}
The proof follows the same method of propogation of chaos as the proof of Lemma \ref{Lemma2Lip}. 
\end{proof}

By bounding the difference in the nonlinear dynamics and original dynamics in Lemma \ref{Lemma1Lip1}, coupled with the observation in (\ref{nonlineardynamicsrhot1}), we see that as a result, for any sequence $(\prescript{}{1}N,\epsilon=\epsilon_{\prescript{}{1}N})$ s.t. $\prescript{}{1}N \to \infty$ and $\epsilon_{\prescript{}{1}N} \to 0$ with $\prescript{}{1}N/\log(\prescript{}{1}N/\epsilon_{\prescript{}{1}N})\to \infty$ and $\epsilon_{\prescript{}{1}N}\log(\prescript{}{1}N/\epsilon_{\prescript{}{1}N}) \to 0$, we have $\prescript{}{1}{\hat\rho}_{\lfloor t/\epsilon \rfloor}^{(\prescript{}{1}N)}$ converges weakly to $\prescript{}{1}\rho_t$ almost surely.

\end{proof}

%%%%%%%%%%%%%%%%%%

We now move on to the convergence properties of the joint distribution. We introduce the following Theorem below as stated in \cite{Kevei2016}, as it will help to reduce proving $(\star)$ to a simpler proof.

\begin{thm}[Grincevicius] \label{grin}
Assume the following conditions hold:

1. $A\geq 0$ a.s., $\mathbb{E} A^\alpha =1$ for some $\alpha >0$, $\mathbb{E} A^\alpha log_+ A < \infty$ and the law of $\log A$ conditioned on $\{A > 0\}$ is arithmetic (a random variable $Y$, or its distribution, is called arithmetic if $Y\in h\mathbb{Z} = \{0,\pm h, \pm 2h, \dots\}$ a.s. for some $h>0$.

2. There exists $\alpha>0$ such that $\mathbb{E}[|B|^\alpha]<\infty$.

3. $\mathbb{P}(Ax+B=x)<1$ for every $x\in \mathbb{R}$.

Then the equation $X \stackrel{d}{=} AX+B$ has a solution $X$ which is independent of $(A,B)$ and there exist functions $q_+,q_-\in \mathbb{Q}$ s.t. $q_+(x)+q_-(x)>0$ and
\begin{align}
\lim_{n\to \infty} \mathbb{P}(X>xe^{nh})=\frac{q_+(x)}{x^\alpha e^{\alpha nh}} \\
\lim_{n\to \infty} \mathbb{P}(X<-xe^{nh})=\frac{q_-(x)}{x^\alpha e^{\alpha nh}},
\end{align}
for all $x\in \mathbb{R}$ if $B\geq 0$ a.s., where 
\begin{align}
\mathbb{Q}=\bigg\{ q: (0,\infty) \to [0,\infty) : &x^{-\alpha} q(x)\text{ is nonincreasing for some }\alpha>0, \\ &q(xe^h)=q(x),\forall x>0, \text{ for some }h>0\bigg\}.
\end{align}
We note that $q\in \mathbb{Q}$ is either strictly positive or identically 0.
\end{thm}

We next are interested in showing that $(\star)$ is true. 

\begin{thm}\label{independence}
Under Assumptions $\ref{assump1}$, $\ref{assump2}$, $\ref{assump4}$, $\ref{assump3}$, and $\ref{assump5}$ the marginal densities $\prescript{}{1}\rho_t$ and $\prescript{}{2}\rho_t$ are independent asymptotically, e.g. $\prescript{}{1,2}\rho=\prescript{}{1}\rho\prescript{}{2}\rho$. $(\star)$
\end{thm}

\begin{proof}
Assumption $\ref{assump5}$ is stated in the following proof.

To show asymptotic independence, we refer to Sibuya's Condition, which states that a bivariate d.f. $F$ with marginal distributions $X_1$ and $X_2$ is asymptotically independent if and only if $\mathbb{P}(X_1>q_1(u) | X_2 >q_2(u))\to 0$ as $u\to 1$, where $q_i(u)$ is the $u$-quantile of the $i$-th marginal distribution.

In the context of our problem, we are interested in showing 
\begin{align} \label{goallimit}
\lim_{u\to \infty} \mathbb{P}\big(|\prescript{}{2}{\bar{\bm \theta}}|>u^{1/\alpha_2} \big||\prescript{}{1}{\bar{\bm \theta}}|>u^{1/\alpha_1} \big) = 0,
\end{align}

\noindent where $\prescript{}{j}{\bar{\bm \theta}}\equiv \lim_{t\to\infty} \prescript{}{j}{\bar{\bm \theta}}^t$ for $j=1,2$. Under Theorem \ref{grin} and simple Bayesian statistics, it is enough to prove

\begin{align} \label{goallimit2}
\lim_{u\to \infty} u\mathbb{P}\big(|\prescript{}{2}{\bar{\bm \theta}}|>u^{1/\alpha_2} ,|\prescript{}{1}{\bar{\bm \theta}}|>u^{1/\alpha_1} \big) = 0,
\end{align}

\noindent where $u$ represents $(xe^{nh})^{\alpha}$ from Theorem \ref{grin}. Under this, and properties of $\mathbb{Q}$, we see that $u^{-1} q_\pm (x)$ is decreasing or identically $0$. Recall that 

\begin{align}
\prescript{}{j}{\bar{\bm \theta}}^t&=\prescript{}{j}{\bar{\bm \theta}}^0+2\int_0^t \xi(s) \prescript{}{j}{\bf G}({\bar{\bm \theta}}^s;\prescript{}{2}\rho_s)ds \\
\prescript{}{1}{\bf G}({\bm \theta};\prescript{}{2}\rho) &= - \nabla_{\prescript{}{1}{\bm \theta}} \Psi_1({\bm \theta},\prescript{}{2}\rho)=-\nabla_{\prescript{}{1}{\bm \theta}} V(\prescript{}{1}{\bm \theta},\prescript{}{2}{\bm \theta})-\int \nabla_{\prescript{}{1}{\bm \theta}} U(\prescript{}{1}{\bm \theta},\prescript{}{2}{\bm \theta},\prescript{}{2}{\bm \theta}')\prescript{}{2}\rho(d\prescript{}{2}{\bm \theta}'), \hspace{1cm} \prescript{}{1}{\bm \theta} \in \mathbb{R}^{\prescript{}{1}D}\\
\prescript{}{2}{\bf G}({\bm \theta};\prescript{}{2}\rho) &= - \nabla_{\prescript{}{2}{\bm \theta}} \Psi_2({\bm \theta},\prescript{}{2}\rho)=-\nabla_{\prescript{}{2}{\bm \theta}} V(\prescript{}{1}{\bm \theta},\prescript{}{2}{\bm \theta})-\int \nabla_{\prescript{}{2}{\bm \theta}} U(\prescript{}{1}{\bm \theta},\prescript{}{2}{\bm \theta},\prescript{}{2}{\bm \theta}')\prescript{}{2}\rho(d\prescript{}{2}{\bm \theta}'), \hspace{1cm} \prescript{}{2}{\bm \theta} \in \mathbb{R}^{\prescript{}{2}D}.
\end{align}

\noindent We can reduce the problem to studying the dominating integrals

\begin{align}
|\prescript{}{j}{\bar{\bm \theta}}^t| \leq \prescript{*}{j}{\bar{\bm \theta}}^t \equiv \int_0^t \big| 2\xi(s) \prescript{}{j}{\bf G}({\bar{\bm \theta}}^s;\prescript{}{2}\rho_s)\big|ds, \hspace{2cm} j=1,2.
\end{align}
% and $\prescript{}{j} Q_i \equiv \int_0^t 2\xi(s) \prescript{}{j}{\bf G}({\bar{\bm \theta}}_i^s;\prescript{}{2}\rho_s)ds$

\noindent Define $\prescript{}{j} Q^s\equiv 2\xi(s) \prescript{}{j}{\bf G}({\bar{\bm \theta}}^s;\prescript{}{2}\rho_s)$. We see that $\prescript{*}{j}{\bar{\bm \theta}} \equiv \lim_{t\to \infty} \prescript{*}{j}{\bar{\bm \theta}}^t$ satisfies the fixed point equation in distribution (i.e. $\prescript{*}{j}{\bar{\bm \theta}} \stackrel{d}{=} \prescript{*}{j}{\bar{\bm \theta}} +|\prescript{}{j} Q |$ for $j=1,2$, where $\prescript{}{j} Q=\lim_{s\to \infty} \prescript{}{j} Q^s$). Applying Theorem \ref{grin} gives us

\begin{align}\label{KGTstar}
\lim_{u \to \infty} u\mathbb{P}( \prescript{*}{2}{\bar{\bm \theta}}>u^{1/\alpha_2})=q_2^*\geq0 \text{ and } \lim_{u \to \infty} u\mathbb{P}( \prescript{*}{1}{\bar{\bm \theta}}>u^{1/\alpha_1})=q_1^*\geq0.
\end{align}

\noindent Since $| \prescript{}{j}{\bar{\bm \theta}}| \leq \prescript{*}{j}{\bar{\bm \theta}}$ for $j=1,2$, asymptotic independence will follow if we show

\begin{align}
\lim_{u\to \infty} u\mathbb{P} \big( \prescript{*}{2}{\bar{\bm \theta}}>u^{1/\alpha_2},\prescript{*}{1}{\bar{\bm \theta}}>u^{1/\alpha_1}\big) =0.
\end{align}

WLOG, assume $\alpha_2 \leq \alpha_1$. Then, by introducing an exit time in the first layer for $\prescript{*}{1}{\bar{\bm \theta}}^t$

\begin{align} \label{lemma33cond1}
\prescript{}{1}T_u \equiv \inf \big\{t\in \mathbb{R}: \prescript{*}{1}{\bar{\bm \theta}}^t > u^{1/\alpha_1} \big\},
\end{align}

\noindent where this represents the first time a component of $\prescript{*}{1}{\bar{\bm \theta}}^t$ is greater than $u^{1/\alpha_1}$, we can see that $\{\prescript{*}{1}{\bar{\bm \theta}}^t > u^{1/\alpha_1}\}=\{\prescript{}{1}T_u < \infty \}$, since $\prescript{*}{1}{\bar{\bm \theta}} = \sup_{t\geq 0} \prescript{*}{1}{\bar{\bm \theta}}^t$. By (\ref{KGTstar}), 

\begin{align} \label{lemma33cond2}
\lim_{u\to \infty} u\mathbb{P} (\prescript{}{1}T_u <\infty)>0,
\end{align}
 and we are left with showing 

\begin{align}
\lim_{u\to \infty} \mathbb{P}\big(  \prescript{*}{2}{\bar{\bm \theta}}>u^{1/\alpha_2}|\prescript{}{1}T_u <\infty \big)=0.
\end{align}

\noindent Considering the partial integrals

\begin{align}
 \prescript{*}{j}{\bar{\bm \theta}}^{m:n} \equiv \int_m^n |\prescript{}{j}Q^s| ds
\end{align}

\noindent for $j=1,2$, we have on the set $\{\prescript{}{1}T_u <\infty\}$

\begin{align}
 \prescript{*}{2}{\bar{\bm \theta}}= \prescript{*}{2}{\bar{\bm \theta}}^{\prescript{}{1}T_u}+ \prescript{*}{2}{\bar{\bm \theta}}^{\prescript{}{1}T_u:\infty}.
\end{align}

\noindent Noticing the inclusion

\begin{align}
\{  \prescript{*}{2}{\bar{\bm \theta}}> d\}\subset \big\{\prescript{*}{2}{\bar{\bm \theta}}^{\prescript{}{1}T_u}> d/2 \big\} \cup \big\{ \prescript{*}{2}{\bar{\bm \theta}}^{\prescript{}{1}T_u:\infty}>d/2 \big\} \equiv I \cup II 
\end{align}

\noindent allows us to separately analyze each part's contribution. We introduce the following Lemma to evaluate and bound the contributions of $I$ and $II$. We state the following assumption to help in the bounding in Lemma \ref{hulemma}.

\begin{assump}\label{assump5}
For any $i$,

\begin{align}
\lim_{u \to \infty} \log(u)\mathbb{P}\bigg( \frac{\prescript{}{2}Q_i}{\prescript{}{1}Q_i} >u^\epsilon\bigg) =0
\end{align}

for all $\epsilon>0$, where $\prescript{}{j}Q_i$ is the $i$-th component of $\prescript{}{j}Q$ for $j=1,2$.
\end{assump}

\begin{lemma} \label{hulemma}
For any $\epsilon>0$, define 

\begin{align}
H_u(\epsilon)\equiv  \bigg\{\prescript{*}{1}{\bar{\bm \theta}}^{\prescript{}{1}T_u}\leq u^{\frac{1+\epsilon}{\alpha_1}} \bigg\} \cap \bigg\{\max_i \max_{0\leq k \leq \prescript{}{1}T_u} \frac{|\prescript{}{2}Q_i^k |}{|\prescript{}{1}Q_i^k |}\leq u^{\frac{\epsilon}{\alpha_1}} \bigg\}
\end{align}

where $\{\prescript{*}{1}{\bar{\bm \theta}}^{\prescript{}{1}T_u}\leq u^{\frac{1+\epsilon}{\alpha_1}} \}$ represents the event that all components of $\prescript{*}{1}{\bar{\bm \theta}}^{\prescript{}{1}T_u}$ are less than or equal to $u^{\frac{1+\epsilon}{\alpha_1}}$. Then, 

\begin{align}
\lim_{u\to \infty} \mathbb{P} \bigg( \{ \prescript{*}{2}{\bar{\bm \theta}}>u^{1/\alpha_2}\} \cap H_u(\epsilon) \bigg| \prescript{}{1}T_u <\infty\bigg) = \lim_{u\to \infty} \mathbb{P} \bigg( \prescript{*}{2}{\bar{\bm \theta}}> u^{1/\alpha_2} \bigg| \prescript{}{1}T_u <\infty\bigg)
\end{align}
as long as the limits exist.

\end{lemma}

\begin{proof}
Consider $\epsilon>0$ and let $H_u\equiv H_u(\epsilon)$. To prove the Lemma, we have to show that $\lim_{u\to \infty} \mathbb{P} (H_u^c |T_u <\infty)=0$, where $H_u^c$ is the complement set of $H_u$, since

\begin{align}
&\mathbb{P}\bigg(\prescript{*}{2}{\bar{\bm \theta}} >u^{1/\alpha_2}\bigg| \prescript{}{1}T_u < \infty \bigg) \geq \mathbb{P}\bigg(\big\{ \prescript{*}{2}{\bar{\bm \theta}}> u^{1/\alpha_2} \big\} \cap H_u \bigg| \prescript{}{1}T_u < \infty \bigg) \\
&= \mathbb{P}\bigg(\prescript{*}{2}{\bar{\bm \theta}} >u^{1/\alpha_2}\bigg| \prescript{}{1}T_u < \infty \bigg)-\mathbb{P}\bigg(\big\{ \prescript{*}{2}{\bar{\bm \theta}}> u^{1/\alpha_2} \big\} \cap H_u^c \bigg| \prescript{}{1}T_u < \infty \bigg) \\
&= \mathbb{P}\bigg(\prescript{*}{2}{\bar{\bm \theta}} >u^{1/\alpha_2}\bigg| \prescript{}{1}T_u < \infty \bigg)-\mathbb{P}\bigg(H_u^c \bigg| \prescript{}{1}T_u < \infty \bigg).
\end{align}

\noindent Define

\begin{align}
H_u^c&=\bigg\{\prescript{*}{1}{\bar{\bm \theta}}_i^{\prescript{}{1}T_u}> u^{\frac{1+\epsilon}{\alpha_1}} \bigg\} \cup \bigg\{\max_i \max_{0\leq k \leq \prescript{}{1}T_u} \frac{|\prescript{}{2}Q_i^k |}{|\prescript{}{1}Q_i^k |}> u^{\frac{\epsilon}{\alpha_1}} \bigg\} \\
&\equiv R_1 \cup R_2.
\end{align}

\noindent It is enough to show that $\lim_{u\to \infty} u \mathbb{P} (H_u^c \cap \{\prescript{}{1}T_u <\infty\}) \leq \lim_{u \to \infty} u \mathbb{P}(H_u^c)=0$ by (\ref{lemma33cond2}). 

For $R_1$, by (\ref{KGTstar}), $\lim_{u\to\infty} u \mathbb{P} ( \prescript{*}{1}{\bar{\bm \theta}}>u^{(1+\epsilon)/\alpha_1} )=0$, which leads to $\lim_{u\to\infty} u \mathbb{P} (\prescript{*}{1}{\bar{\bm \theta}}^{\prescript{}{1}T_u}>u^{(1+\epsilon)/\alpha_1} )=0$ since $\prescript{*}{1}{\bar{\bm \theta}}=\sup_{t\geq 0} \prescript{*}{1}{\bar{\bm \theta}}^t$.

For $R_2$, for an arbitrary $i$, using a union bound gives that

\begin{align}
\mathbb{P} \bigg(\max_{0\leq k \leq \prescript{}{1}T_u} \frac{|\prescript{}{2}Q_i^k |}{|\prescript{}{1}Q_i^k |}> u^{\frac{\epsilon}{\alpha_1}},\prescript{}{1}T_u <\infty \bigg) \leq \sum_{k=0}^{\prescript{}{1}T_u}  \mathbb{P} \bigg( u^{\epsilon/\alpha_1} |\prescript{}{1}Q_i^k |<|\prescript{}{2}Q_i^k | ,\prescript{}{1}T_u < \infty\bigg).
\end{align} 

\noindent Decomposing the term in the summand gives

\begin{align}
 &\mathbb{P} \bigg( u^{\epsilon/\alpha_1} |\prescript{}{1}Q_i^k |<|\prescript{}{2}Q_i^k | ,\prescript{}{1}T_u < \infty\bigg) \leq  \mathbb{P} \bigg( u^{\epsilon/\alpha_1} |\prescript{}{1}Q_i^k |<|\prescript{}{2}Q_i^k | ,\prescript{*}{1}{\bar{\bm \theta}}_i^t > u^{1/\alpha_1}\bigg)\\
&\leq   \mathbb{P} \bigg( u^{\epsilon/\alpha_1} |\prescript{}{1}Q_i^k |<|\prescript{}{2}Q_i^k | ,|\prescript{}{1}Q_i^k |+\sum_{\ell\not=k} |\prescript{}{1}Q_i^\ell | > u^{1/\alpha_1}\bigg) \\
&\leq  \mathbb{P} \bigg( u^{\epsilon/\alpha_1} |\prescript{}{1}Q_i^k |<|\prescript{}{2}Q_i^k | ,|\prescript{}{1}Q_i^k | > u^{1/\alpha_1}/2\bigg) \nonumber \\
&\hspace{1cm} +  \mathbb{P} \bigg( u^{\epsilon/\alpha_1} |\prescript{}{1}Q_i^k |<|\prescript{}{2}Q_i^k | ,\sum_{\ell\not=k} |\prescript{}{1}Q_i^\ell | > u^{1/\alpha_1}/2\bigg). \label{decompprob}
\end{align}

\noindent If we can show that both of the probabilities, when summed over $k$ from $0$ to $\prescript{}{1}T_u$, are of order $o(u^{-1})$, then as $u\to\infty$, we will have our result.

\noindent For the second term of (\ref{decompprob}), we have by independence that
%since the bar theta are independent and functions of independent are independent

\begin{align}
&\mathbb{P} \bigg( u^{\epsilon/\alpha_1} |\prescript{}{1}Q_i^k |<|\prescript{}{2}Q_i^k | ,\sum_{\ell\not=k} |\prescript{}{1}Q_i^\ell | > u^{1/\alpha_1}/2\bigg) \\
&=\mathbb{P} \bigg( u^{\epsilon/\alpha_1} |\prescript{}{1}Q_i |<|\prescript{}{2}Q_i |\bigg) \mathbb{P}\bigg(\sum_{\ell\not=k} |\prescript{}{1}Q_i^\ell | > u^{1/\alpha_1}/2\bigg) \\
&\leq \mathbb{P} \bigg( u^{\epsilon/\alpha_1} |\prescript{}{1}Q_i |<|\prescript{}{2}Q_i |\bigg) \mathbb{P}\bigg( \prescript{*}{1}{\bar{\bm \theta}}_i> u^{1/\alpha_1}/2\bigg) \\
&= o\bigg(\frac{1}{\log(u)u}\bigg)
\end{align}

%RETHINK THIS... need to add $\{T_u > L_u\}$ TO FIND BIG O FOR GROWTH.

\noindent under Assumption \ref{assump5} and by (\ref{KGTstar}). Since $\prescript{}{1}T_u$ is bounded above by $O(\log(u))$ by Theorem 2.1 of \cite{Buretal2014}, summing over $k$ from $0$ to $\prescript{}{1}T_u$ gives the necessary contribution of $o(u^{-1})$.

\noindent For the first term of (\ref{decompprob}), we have by Markov's Inequality on order $\frac{\alpha_1}{1+\epsilon} < \kappa < \alpha_1$ that 

\begin{align}
&\mathbb{P} \bigg( u^{\epsilon/\alpha_1} |\prescript{}{1}Q_i^k |<|\prescript{}{2}Q_i^k | ,|\prescript{}{1}Q_i^k | > u^{1/\alpha_1}/2\bigg) \\
&\leq  \mathbb{P} \bigg(|\prescript{}{2}Q_i^k | > u^{(1+\epsilon)/\alpha_1}/2\bigg) \\
&\leq \frac{2^\kappa\mathbb{E}\| \prescript{}{2}Q_i^k \|^\kappa}{u^{\kappa(1+\epsilon)/\alpha_1}}.
\end{align}

\noindent Summing over $k$ from $0$ to $\infty$, by applying Assumption \ref{assump4} and ${\bf G}$ being Lipschitz continuous, we have

\begin{align}
&\sum_{k=0}^\infty \mathbb{P} \bigg( u^{\epsilon/\alpha_1} |\prescript{}{1}Q_i^k |<|\prescript{}{2}Q_i^k | ,|\prescript{}{1}Q_i^k | > u^{1/\alpha_1}/2\bigg)\\
&\leq \sum_{k=0}^\infty \frac{2^\kappa\mathbb{E}\| \prescript{}{2}Q_i^k \|^\kappa}{u^{\kappa(1+\epsilon)/\alpha_1}} \\
&\leq \sum_{k=0}^\infty \frac{2^\kappa K^\kappa}{u^{\kappa(1+\epsilon)/\alpha_1}}=o(u^{-1})
\end{align}

\noindent as desired, where $K$ is again some constant dependent on $K_1$, $K_2$, and $K_3$ from Assumptions \ref{assump1}, \ref{assump2}, \ref{assump4}.

\end{proof}

 Now, considering $I$, we utilize Lemma \ref{hulemma}. We have the following, where again $\prescript{}{j} Q_i^s$ represents the $i$-th component of $\prescript{}{j} Q^s$:

\begin{align}
%\frac{\prescript{*}{2}{\bar{\bm \theta}}_i^{\prescript{}{1}T_u}}{\prescript{*}{1}{\bar{\bm \theta}}_i^{\prescript{}{1}T_u}}&=\frac{\int_0^{\prescript{}{1}T_u}|\prescript{}{2}Q_i^s| ds}{\int_0^{\prescript{}{1}T_u}|\prescript{}{1}Q_i^s| ds} \\
%&\leq \max_i \max_{0 \leq s \leq \prescript{}{1}T_u} \frac{|\prescript{}{2}Q_i^s|}{|\prescript{}{1}Q_i^s|} \\
%\Rightarrow \prescript{*}{2}{\bar{\bm \theta}}_i^{\prescript{}{1}T_u} &\leq \max_{0\leq k \leq \prescript{}{1}T_u} \frac{|\prescript{}{2}Q_i^k |}{|\prescript{}{1}Q_i^k |} \prescript{*}{1}{\bar{\bm \theta}}_i^{\prescript{}{1}T_u} \\
 \prescript{*}{2}{\bar{\bm \theta}}^{\prescript{}{1}T_u} &\leq  \max_i  \max_{0 \leq s \leq \prescript{}{1}T_u} \frac{|\prescript{}{2}Q_i^s|}{|\prescript{}{1}Q_i^s|}\prescript{*}{1}{\bar{\bm \theta}}^{\prescript{}{1}T_u} \\
&\leq \max_i \max_{0 \leq s \leq \prescript{}{1}T_u} \frac{|\prescript{}{2}Q_i^s|}{|\prescript{}{1}Q_i^s|} u^{(1+\epsilon)/\alpha_1} \\
&\leq u^{\frac{1}{\alpha_1}*(1+2\epsilon)} \label{Iupperbound}
\end{align}

\noindent Then, choosing $\epsilon$ sufficiently small, we have, by (\ref{Iupperbound}), Assumption \ref{assump5}, and assumption from earlier that $\alpha_2\leq\alpha_1$, that

\begin{align}
\bigg\{\big\{\prescript{*}{2}{\bar{\bm \theta}}^{\prescript{}{1}T_u}>\frac{u^{1/\alpha_2}}{2}\big\} \cap H_u(\epsilon)\bigg\} \subset \bigg\{ u^{\frac{1}{\alpha_1}(1+2\epsilon)}\geq \prescript{*}{2}{\bar{\bm \theta}}^{\prescript{}{1}T_u}>\frac{u^{1/\alpha_2}}{2}\bigg\}=\emptyset
\end{align}

\noindent for large enough $u$. Thus, by Lemma \ref{hulemma}, $I$ does not contribute.

Next, for $II$, since $\prescript{*}{2}{\bar{\bm \theta}}^{\prescript{}{1}T_u : \infty}$ is independent of $\{\prescript{}{1}T_u < \infty \}$, 

\begin{align}
&\lim_{u\to \infty} \mathbb{P} \bigg( \prescript{*}{2}{\bar{\bm \theta}}^{\prescript{}{1}T_u : \infty}> \frac{u^{1/\alpha_2}}{2} \bigg| \prescript{}{1}T_u <\infty\bigg) \\
&=\lim_{u\to \infty} \mathbb{P} \bigg( \prescript{*}{2}{\bar{\bm \theta}}^{\prescript{}{1}T_u : \infty}> \frac{u^{1/\alpha_2}}{2} \bigg)  \\
&\leq\lim_{u\to \infty} \mathbb{P} \bigg( \prescript{*}{2}{\bar{\bm \theta}}> \frac{u^{1/\alpha_2}}{2} \bigg). 
\end{align}

\noindent By (\ref{KGTstar}), $\mathbb{P} \bigg( \prescript{*}{2}{\bar{\bm \theta}}> \frac{u^{1/\alpha_2}}{2} \bigg)$ tends to zero for large enough $u$, which means that $II$ tends to zero, as desired.

Thus, we conclude that

\begin{align}
\lim_{u\to \infty} \mathbb{P}\big(  \prescript{*}{2}{\bar{\bm \theta}}>u^{1/\alpha_2}|\prescript{}{1}T_u <\infty \big)=0
\end{align}

\noindent and our result follows.

\end{proof}

\begin{cor}\label{prop2}

Assume Assumptions \ref{assump1}, \ref{assump2}, \ref{assump4}, and \ref{assump3} hold. Then, $\prescript{}{1}\rho_t$ and $\prescript{}{2}\rho_t$ are asymptotically uncorrelated.

\end{cor}

\begin{proof}

From the previous proofs on the convergence properties in the individual layers, we had that

\begin{align}
\prescript{}{j}{\bm \theta}_i^{k}&=\prescript{}{j}{\bm \theta}_i^{0}+2\epsilon \sum_{\ell=0}^{k-1} \xi(\ell\epsilon)\prescript{}{j}{\bf F}_i({\bm \theta}_i^\ell,{\bm \beta}_{\ell+1}) \\
\prescript{}{j}{\bar{\bm \theta}}_i^t&=\prescript{}{j}{\bar{\bm \theta}}_i^0+2\int_0^t \xi(s) \prescript{}{j}{\bf G}({\bar{\bm \theta}}_i^s;\prescript{}{2}\rho_s)ds \label{corrsub}
\end{align}

\noindent with the the first line being the SGD dynamics and second line being the nonlinear dynamics of the PDE of layer $j=1,2$, where 

\begin{align}
\prescript{}{1}{\bf F}_i({\bm \theta};{\bm \beta}_k)&=\big(y_k-{\hat y}({\bf x}_k;{\bm \theta})\big)\nabla_{\prescript{}{1}{\bm \theta}_i} \prescript{}{1}\sigma_*({\bf x}_k,\prescript{}{1}{\bm \theta}), \hspace{3.5cm} \prescript{}{1}{\bm \theta}=(\prescript{}{1}{\bm \theta}_i)_{i\leq \prescript{}{1}N} \in \mathbb{R}^{\prescript{}{1} D\times \prescript{}{1}N} \\
\prescript{}{1}{\bf G}({\bm \theta};\prescript{}{2}\rho) &= - \nabla_{\prescript{}{1}{\bm \theta}} \Psi_1({\bm \theta},\prescript{}{2}\rho)=-\nabla_{\prescript{}{1}{\bm \theta}} V(\prescript{}{1}{\bm \theta},\prescript{}{2}{\bm \theta})-\int \nabla_{\prescript{}{1}{\bm \theta}} U(\prescript{}{1}{\bm \theta},\prescript{}{2}{\bm \theta},\prescript{}{2}{\bm \theta}')\prescript{}{2}\rho(d\prescript{}{2}{\bm \theta}'), \hspace{1cm} \prescript{}{1}{\bm \theta} \in \mathbb{R}^{\prescript{}{1}D}\\
\prescript{}{2}{\bf F}_i({\bm \theta};{\bm \beta}_k)&=\big(y_k-{\hat y}({\bf x}_k;{\bm \theta})\big)\nabla_{\prescript{}{2}{\bm \theta}_i} \prescript{}{2}\sigma_*(\prescript{}{1}\sigma_*({\bf x}_k,\prescript{}{1}{\bm \theta}),\prescript{}{2}{\bm \theta}_i ), \hspace{2cm} \prescript{}{2}{\bm \theta}=(\prescript{}{2}{\bm \theta}_i)_{i\leq \prescript{}{2}N} \in \mathbb{R}^{\prescript{}{2} D\times \prescript{}{2}N} \\
\prescript{}{2}{\bf G}({\bm \theta};\prescript{}{2}\rho) &= - \nabla_{\prescript{}{2}{\bm \theta}} \Psi_2({\bm \theta},\prescript{}{2}\rho)=-\nabla_{\prescript{}{2}{\bm \theta}} V(\prescript{}{1}{\bm \theta},\prescript{}{2}{\bm \theta})-\int \nabla_{\prescript{}{2}{\bm \theta}} U(\prescript{}{1}{\bm \theta},\prescript{}{2}{\bm \theta},\prescript{}{2}{\bm \theta}')\prescript{}{2}\rho(d\prescript{}{2}{\bm \theta}'), \hspace{1cm} \prescript{}{2}{\bm \theta} \in \mathbb{R}^{\prescript{}{2}D}.
\end{align}

We will work with the nonlinear dynamics of the PDE, since in the proofs of Theorems \ref{layer2convergence} and \ref{layer1convergence}, it was shown that the difference in the SGD and nonlinear dynamics is bounded. To show that two distributions $X$ and $Y$ are uncorrelated, we need to show that $\mathbb{E}[XY]=\mathbb{E}[X]\mathbb{E}[Y]$. Thus, our goal is showing that

\begin{align}
\mathbb{E}[\prescript{}{1}{\bar{\bm \theta}}_i^t\prescript{}{2}{\bar{\bm \theta}}_i^t ] =\mathbb{E}[\prescript{}{1}{\bar{\bm \theta}}_i^t] \mathbb{E}[\prescript{}{2}{\bar{\bm \theta}}_i^t]
\end{align}

\noindent since the $\prescript{}{j}{\bar{\bm \theta}}_i^t$ are i.i.d. within each layer $j=1,2$. Substituting from (\ref{corrsub}), we have

\begin{align}
\mathbb{E}[\prescript{}{1}{\bar{\bm \theta}}_i^t\prescript{}{2}{\bar{\bm \theta}}_i^t ] &= \mathbb{E}\bigg[\big\{\prescript{}{1}{\bar{\bm \theta}}_i^0+2\int_0^t \xi(s) \prescript{}{1}{\bf G}({\bar{\bm \theta}}_i^s;\prescript{}{2}\rho_s)ds \big\}\big\{\prescript{}{2}{\bar{\bm \theta}}_i^0+2\int_0^t \xi(s) \prescript{}{2}{\bf G}({\bar{\bm \theta}}_i^s;\prescript{}{2}\rho_s)ds  \big\}  \bigg] \\
&=\mathbb{E}\bigg[ \prescript{}{1}{\bar{\bm \theta}}_i^0\prescript{}{2}{\bar{\bm \theta}}_i^0 +2\prescript{}{1}{\bar{\bm \theta}}_i^0\int_0^t \xi(s) \prescript{}{2}{\bf G}({\bar{\bm \theta}}_i^s;\prescript{}{2}\rho_s)ds +2\prescript{}{2}{\bar{\bm \theta}}_i^0\int_0^t \xi(s) \prescript{}{1}{\bf G}({\bar{\bm \theta}}_i^s;\prescript{}{2}\rho_s)ds \nonumber \\
&\hspace{1cm} + 4\int_0^t \xi(s) \prescript{}{1}{\bf G}({\bar{\bm \theta}}_i^s;\prescript{}{2}\rho_s)ds\int_0^t \xi(s) \prescript{}{2}{\bf G}({\bar{\bm \theta}}_i^s;\prescript{}{2}\rho_s)ds \bigg] \\
&=\prescript{}{1}{\bar{\bm \theta}}_i^0\prescript{}{2}{\bar{\bm \theta}}_i^0+2\prescript{}{1}{\bar{\bm \theta}}_i^0\mathbb{E}\bigg[\int_0^t \xi(s) \prescript{}{2}{\bf G}({\bar{\bm \theta}}_i^s;\prescript{}{2}\rho_s)ds\bigg] +2\prescript{}{2}{\bar{\bm \theta}}_i^0\mathbb{E}\bigg[\int_0^t \xi(s) \prescript{}{1}{\bf G}({\bar{\bm \theta}}_i^s;\prescript{}{2}\rho_s)ds\bigg] \nonumber \\
&\hspace{1cm} + 4\mathbb{E}\bigg[\int_0^t \xi(s) \prescript{}{1}{\bf G}({\bar{\bm \theta}}_i^s;\prescript{}{2}\rho_s)ds\int_0^t \xi(s) \prescript{}{2}{\bf G}({\bar{\bm \theta}}_i^s;\prescript{}{2}\rho_s)ds \bigg]
\end{align}

\noindent If we can show

\begin{align}
\mathbb{E}\bigg[\int_0^t \xi(s) \prescript{}{1}{\bf G}({\bar{\bm \theta}}_i^s;\prescript{}{2}\rho_s)ds\int_0^t \xi(s) \prescript{}{2}{\bf G}({\bar{\bm \theta}}_i^s;\prescript{}{2}\rho_s)ds\bigg] =\mathbb{E}\bigg[\int_0^t \xi(s) \prescript{}{1}{\bf G}({\bar{\bm \theta}}_i^s;\prescript{}{2}\rho_s)ds \bigg] \mathbb{E}\bigg[\int_0^t \xi(s) \prescript{}{2}{\bf G}({\bar{\bm \theta}}_i^s;\prescript{}{2}\rho_s)ds \bigg]
\end{align}

\noindent then we are done. By assumption, the integrals $\int_0^t \xi(s) \prescript{}{1}{\bf G}({\bar{\bm \theta}}_i^s;\prescript{}{2}\rho_s)ds$ and $\int_0^t \xi(s) \prescript{}{2}{\bf G}({\bar{\bm \theta}}_i^s;\prescript{}{2}\rho_s)ds$ are bounded and convergent. Then, by linearity,

\begin{align}
\mathbb{E}\bigg[\int_0^t \xi(s) \prescript{}{1}{\bf G}({\bar{\bm \theta}}_i^s;\prescript{}{2}\rho_s)ds&\int_0^t \xi(s) \prescript{}{2}{\bf G}({\bar{\bm \theta}}_i^s;\prescript{}{2}\rho_s)ds\bigg]\\
&=\mathbb{E}\bigg[\int_0^t \bigg(\int_0^t \xi(s) \prescript{}{1}{\bf G}({\bar{\bm \theta}}_i^s;\prescript{}{2}\rho_s)ds \bigg)\xi(w) \prescript{}{2}{\bf G}({\bar{\bm \theta}}_i^w;\prescript{}{2}\rho_w)dw \bigg] \\
&=\mathbb{E}\bigg[\int_0^t \int_0^t \xi(s) \prescript{}{1}{\bf G}({\bar{\bm \theta}}_i^s;\prescript{}{2}\rho_s)\xi(w) \prescript{}{2}{\bf G}({\bar{\bm \theta}}_i^w;\prescript{}{2}\rho_w)dsdw \bigg] \\
&=\int\int\int_0^t \int_0^t \xi(s) \prescript{}{1}{\bf G}({\bar{\bm \theta}}_i^s;\prescript{}{2}\rho_s)\xi(w) \prescript{}{2}{\bf G}({\bar{\bm \theta}}_i^w;\prescript{}{2}\rho_w)dsdw \prescript{}{1,2}\rho_t d\prescript{}{1}{\bar{\bm \theta}}_i^td\prescript{}{2}{\bar{\bm \theta}}_i^t
\end{align}

\noindent Under Theorem \ref{independence}, we can separate the density $\prescript{}{1,2}\rho_t$ and are left with

\begin{align}
&\int\int\int_0^t \int_0^t \xi(s) \prescript{}{1}{\bf G}({\bar{\bm \theta}}_i^s;\prescript{}{2}\rho_s)\xi(w) \prescript{}{2}{\bf G}({\bar{\bm \theta}}_i^w;\prescript{}{2}\rho_w)dsdw \prescript{}{1}\rho_t\prescript{}{2}\rho_t d\prescript{}{1}{\bar{\bm \theta}}_i^td\prescript{}{2}{\bar{\bm \theta}}_i^t \\
&= \int \int_0^t \xi(s) \prescript{}{1}{\bf G}({\bar{\bm \theta}}_i^s;\prescript{}{2}\rho_s)ds \prescript{}{1}\rho_t d\prescript{}{1}{\bar{\bm \theta}}_i^t \int \int_0^t \xi(w) \prescript{}{2}{\bf G}({\bar{\bm \theta}}_i^w;\prescript{}{2}\rho_w) dw d\prescript{}{2}{\bar{\bm \theta}}_i^t  \\
&=\mathbb{E}\bigg[\int_0^t \xi(s) \prescript{}{1}{\bf G}({\bar{\bm \theta}}_i^s;\prescript{}{2}\rho_s)ds \bigg] \mathbb{E}\bigg[\int_0^t \xi(s) \prescript{}{2}{\bf G}({\bar{\bm \theta}}_i^s;\prescript{}{2}\rho_s)ds \bigg]
\end{align}

\noindent as desired by utilizing Fubini's Theorem.
%https://math.stackexchange.com/questions/1603784/derivative-of-expected-value-with-respect-to-parameter-in-both-pdf-and-expectati
\end{proof}

%%%%%%%%%%%%%%%%%%%%%%%%%%%%%%%%%%%%%%%%%
\subsection{Centered Isotropic Gaussians}

In this section, we detail a way to simulate the PDE dynamics through evolution equations.

\subsubsection{Statics} \label{statics}
%similar to section 8.1 from original paper
We consider centered isotropic Gaussians, i.e. assume the joint law of $(y,{\bf x})$ follows with probability $1/2$, $y=1$, ${\bf x} \sim N({\bm 0},(1+\Delta)^2 {\bf I}_d)$ and with probability $1/2$, $y=-1$, ${\bf x} \sim N({\bm 0},(1-\Delta)^2 {\bf I}_d)$. We will assume $0<\Delta<1$, and choose $\prescript{}{i}\sigma_*({\bf z};\prescript{}{i} {\bm \theta}_i)=\prescript{}{i}\sigma(\langle {\bf z}, \prescript{}{i} {\bf w}_i\rangle)$ for $i=1,2$ for some activation function $\sigma$. 

Through this section, set $\tau_{\pm} =(1\pm \Delta)$ and $q_\pm (r_1,r_2)= q (\tau_+r_1,\tau_\pm r_2)$. Also, assume that $x\mapsto \sigma(x)$ is bounded, non-decreasing, Lipschitz continuous and its weak derivative $x\mapsto \sigma'(x)$ is Lipschitz in a neighborhood of 0.

Since the distribution of ${\bf x}$ is invariant under rotations for each of the two classes, the functions 
\begin{align}
V({\bf w}_1,{\bf w}_2)&=v(\lVert{\bf w}_1 \rVert_2,\lVert{\bf w}_2 \rVert_2),\\
U({\bf w}_1,{\bf w}_2,{\bf w}_3)&=u_0(\lVert{\bf w}_1 \rVert_2,\lVert{\bf w}_2 \rVert_2,\lVert{\bf w}_3\rVert_2, \langle {\bf w}_2,{\bf w}_3\rangle)
\end{align}

\noindent are as well, with $ \langle {\bf w}_2,{\bf w}_3\rangle$ being the inner product of ${\bf w}_2$ and ${\bf w}_3$. These take the form
\begin{align}
v(r_1,r_2)&=-\frac{1}{2}q(\tau_+r_1,\tau_+r_2)+\frac{1}{2}q(\tau_+ r_1,\tau_- r_2), \indent q(t_1,t_2)=\mathbb{E} \{\sigma_2 (t_2\sigma_1(t_1 G)) \} \\
u_0(r_1,r_2,r_3,r_2r_3\cos \alpha) &= \frac{1}{2}\mathbb{E}\bigg\{ \sigma_2(\tau_+ r_2 \sigma_1(\tau_+ r_1 G_1))\sigma_2(\tau_+ r_3 \sigma_1(\tau_+ r_1 G_2))\bigg\} \nonumber \\
&\indent + \frac{1}{2}\mathbb{E}\bigg\{ \sigma_2(\tau_- r_2 \sigma_1(\tau_+ r_1 G_1))\sigma_2(\tau_+ r_3 \sigma_1(\tau_+ r_1 G_2))\bigg\}
\end{align}

\noindent with the expectations being in respect to standard normals $G,G_1,G_2 \sim N(0,1)$ with $(G_1,G_2)$ jointly Gaussian and $\mathbb{E} \{G_1G_2\}=\cos \alpha$.

To minimize $R({\prescript{}{2} \rho})$, we restrict ourselves to distributions that are invariant under rotations. For any probability distribution $\rho$ on $\mathbb{R}^d$, we can define its symmetrization $\rho_s(Q)\equiv \int \rho ({\bf R} Q)\mu_{\text{Haar}} (d{\bf R})$ for any Borel set $Q\subseteq \mathbb{R}^d$, where $\mu_{\text{Haar}}$ is the Haar measure over the group of orthogonal rotations. Since ${\prescript{}{2} \rho} \to R({\prescript{}{2} \rho})$ is convex, $R({\prescript{}{2} \rho_s}) \leq R({\prescript{}{2} \rho})$.

Denote by $\prescript{}{i} {\bar \rho}$ the probability distribution of $\|{\bf w}_i \|_2$ when ${\bf w}_i \sim \prescript{}{i}\rho$ for $i=1,2$. In addition, let ${\bar R}_d (\prescript{}{2} {\bar \rho})$ denote the resulting risk. Then, we have
\begin{align}
{\bar R}_d (\prescript{}{2}{\bar \rho}) &=1+ 2\int v(r_1,r_2 )\prescript{}{2} {\bar \rho}(dr_2) + \int u_d (r_1,r_2,r_3) \prescript{}{2} {\bar \rho} (dr_2) \prescript{}{2} {\bar \rho}(dr_3)\\
u_d(r_1,r_2,r_3)&=\mathbb{E}[u_0(r_1,r_2,r_3,r_2r_3\cos\Theta)]
\end{align}

\noindent where $\Theta \sim (1/Z_d)\sin^{d-2} \theta \cdot {\bm 1}\{\theta \in [0,\pi]\}d\theta$. As $d\to \infty$, $\lim_{d\to \infty} u_d(r_1,r_2,r_3 )=u_\infty (r_1,r_2,r_3)$ (uniformly over compact sets), with 
\begin{align}
u_\infty (r_1,r_2,r_3)=\frac{1}{2}\bigg[ q(\tau_+r_1,\tau_+ r_2) q(\tau_+r_1,\tau_+ r_3)+q(\tau_+r_1,\tau_- r_2)q(\tau_+r_1,\tau_+ r_3) \bigg].
\end{align}

\noindent The risk function converges as

\begin{align}
\label{convergeRisk}
{\bar R}_\infty (\prescript{}{2}{\bar \rho}) = \frac{1}{2}\bigg(1-\int q(\tau_+ r_1,\tau_+ r_2)\prescript{}{2}{\bar \rho}(dr_2) \bigg)^2+ \frac{1}{2}\bigg(1+\int q(\tau_+ r_1,\tau_- r_2)\prescript{}{2}{\bar \rho}(dr_2) \bigg)^2
\end{align}

We also define

\begin{align}
\psi_d(r_1,r_2) &=v(r_1,r_2)+\int u_d(r_1,r_2,r_2')\prescript{}{2}{\bar \rho}(dr_2).
\end{align}

\noindent As $d\to \infty$, these expressions simplify down to 

\begin{align}
\label{psi2}
\psi_\infty (r_1,r_2;\prescript{}{2}{\bar \rho})&=\lambda_+ (\prescript{}{2} {\bar \rho})\cdot q_+ (r_1,r_2)+\lambda_-(\prescript{}{2}{\bar \rho})\cdot q_-(r_1,r_2) \\
\lambda_+ (\prescript{}{2} {\bar \rho}) &=\frac{1}{2} [\langle q_+, \prescript{}{2}{\bar \rho}\rangle -1 ] \\
\lambda_- (\prescript{}{2} {\bar \rho}) &=\frac{1}{2} [ \langle q_-,\prescript{}{2}{\bar \rho}\rangle +1]
\end{align}

\noindent where $\langle f, \mu \rangle=\langle \mu, f \rangle=\int f d\mu$. 

\subsubsection{Empirical Validation of Distributional Dynamics}

Since simulating the PDE in the joint case for general $d$ is computationally intensive, we consider the case where $d=\infty$. The risk is then given by (\ref{convergeRisk}), as

\begin{align}
{\bar R}_\infty (\prescript{}{2}{\bar \rho}) = \frac{1}{2}\bigg(1-\int q(\tau_+ r_1,\tau_+ r_2)\prescript{}{2}{\bar \rho}(dr_2) \bigg)^2+ \frac{1}{2}\bigg(1+\int q(\tau_+ r_1,\tau_- r_2)\prescript{}{2}{\bar \rho}(dr_2) \bigg)^2
\end{align}

\noindent where $q(t_1,t_2)=\mathbb{E} \{\sigma_2 (t_2\sigma_1(t_1 G))\}$, $G\sim N(0,1)$. In addition, from (\ref{psi2}), we had that

\begin{align}
 \psi_\infty (r_1,r_2;\prescript{}{2}{\bar \rho})&=\lambda_+ (\prescript{}{2} {\bar \rho})\cdot q_+ (r_1,r_2)+\lambda_-(\prescript{}{2}{\bar \rho})\cdot q_-(r_1,r_2) \\
\lambda_+ (\prescript{}{2} {\bar \rho}) &=\frac{1}{2} [\langle q_+, \prescript{}{2}{\bar \rho}\rangle -1 ] \\
\lambda_- (\prescript{}{2} {\bar \rho}) &=\frac{1}{2} [ \langle q_-,\prescript{}{2}{\bar \rho}\rangle +1].
\end{align}

\noindent The PDE is then given by 

\begin{align}
\partial_t \prescript{}{1,2}{\bar \rho}_t = 2\xi(t)\bigg(\partial_{\bf r} [\prescript{}{1,2}{\bar \rho}_t \partial_{\bf r}\psi_\infty (r_1,r_2; \prescript{}{2}{\bar \rho}_t) ] \bigg) 
\end{align}

\noindent where ${\bf r}=(r_1,r_2)$. The solution to the PDE is approximated, at all times $t$, by the multiple-deltas ansatz

\begin{align}
\prescript{}{1}{\bar \rho}_t &=\frac{1}{J} \sum_{i=1}^J \delta_{r_{1,i}(t)} \\
\prescript{}{2}{\bar \rho}_t &=\frac{1}{J} \sum_{i=1}^J \delta_{r_{2,i}(t)} 
\end{align}

\noindent where $J\in \mathbb{N}$ is a pre-chosen parameter. For any fixed $J$, if the PDE is initialized at $\prescript{}{1}{\bar \rho}_0, \prescript{}{2}{\bar \rho}_0$ taking the above form, then for any $t\geq 0$, $\prescript{}{i} {\bar \rho}_t$ remains in the above form for $i=1,2$. So, for any smooth test function $f: \mathbb{R} \to \mathbb{R}$ with compact support, where $m=1,2$ depending on which layer we are in, we have

\begin{align}
\frac{1}{J} \sum_{i=1}^J f'(r_{m,i}(t))r_{m,i}'(t)&=\partial_t \langle f,\prescript{}{m} {\bar \rho}_t \rangle =-2\xi(t) \langle f', \prescript{}{m}{\bar \rho}_t \partial_{r_m}  \psi_\infty (r_1,r_2;\prescript{}{2}{\bar \rho}_t)\rangle \\
&= -2\xi (t) \frac{1}{J} \sum_{i=1}^J f'(r_{m,i}(t))\partial_{r_m}  \psi_\infty (r_{1,i}(t),r_{2,i}(t);\prescript{}{2} {\bar \rho}_t).
\end{align}

\noindent Under this ansatz, we write ${\bar R}_\infty (\prescript{}{2}{\bar \rho}_t)={\bar R}_{\infty,J} ({\bf r}_1(t),{\bf r}_2(t))$, where ${\bf r}_m(t)=(r_{m,1}(t),\dots,r_{m,J}(t) )^T$ and 

\begin{align}
{\bar R}_{\infty,J} ({\bf r}_1,{\bf r}_2)=\frac{1}{2} \bigg(1-\frac{1}{J}\sum_{i=1}^J q_+(r_{1,i},r_{2,i}) \bigg)^2+\frac{1}{2} \bigg(1+\frac{1}{J}\sum_{i=1}^J q_-(r_{1,i},r_{2,i}) \bigg)^2.
\end{align}

We can see that $\partial_{r_m} \psi_\infty (r_{1,i}(t),r_{2,i}(t);\prescript{}{2} {\bar \rho}_t) =\frac{J}{2} (\nabla_{{\bf r}_m} {\bar R}_{\infty,J}({\bf r}_1(t),{\bf r}_2(t)) )_i $ for $m=1,2$. Therefore, we have the following system:

\begin{align}
\frac{d}{dt} {\bf r}_1(t)=-J\xi(t) \nabla_{{\bf r}_1} {\bar R}_{\infty,J} ({\bf r}_1(t),{\bf r}_2(t) ) \\
\frac{d}{dt} {\bf r}_2(t)=-J\xi(t) \nabla_{{\bf r}_2} {\bar R}_{\infty,J} ({\bf r}_1(t),{\bf r}_2(t) ).
\end{align}

\noindent Under the multiple-deltas ansatz, we can numerically simulate the PDE using the above evolution equation of ${\bf r}_1(t)$ and ${\bf r}_1(t)$. So, given ${\bf r}_1(t)$ and ${\bf r}_2(t)$, we approximate ${\bf r}_1(t+\delta t)$ and ${\bf r}_2(t+\delta t)$ for small displacement $\delta t$ as

\begin{align}
{\bf r}_1(t+\delta) &\approx {\bf r}_1(t)-J\xi(t) \nabla_{{\bf r}_1}{\bar R}_{\infty,J} ({\bf r}_1(t),{\bf r}_2(t))\delta t \label{DD1}\\
{\bf r}_2(t+\delta) &\approx {\bf r}_2(t)-J\xi(t) \nabla_{{\bf r}_2}{\bar R}_{\infty,J} ({\bf r}_1(t),{\bf r}_2(t))\delta t. \label{DD2}
\end{align}
%%%%%%%%%%%%%%

\section{Exploratory Applied Work}

\subsection{Background}

Every year, insurance companies and their policyholders lose billions of dollars due to car insurance fraud. Such examples of fraudulent activities include staged accidents, counterfeit air bags, and towing scams. As one of the largest such insurance companies, American Family has developed a Fraud Program internally to combat and identify fraudulent activity. Its purpose is to effectively mitigate insurance fraud using Machine Learning in an automated fashion to optimize manual process and intervention. In order to do this, they seek to create a comprehensive fraud detection solution, using a variety of internal and external data as well as software frameworks, in order to identify and mitigate fraud. The hope is that fraud solutions produced can more accurately detect as well as adapt to fraudulent activities resulting in increased profitability and efficiency. All in all, the measurable gains can be established as improved fraud referral quality and mitigation rate, and an increase in dollars mitigated.

\begin{figure}[h]
\caption{Table summarizing internal data set features from American Family.}
\label{SummaryFeatures}
\centering
\includegraphics[scale=0.4]{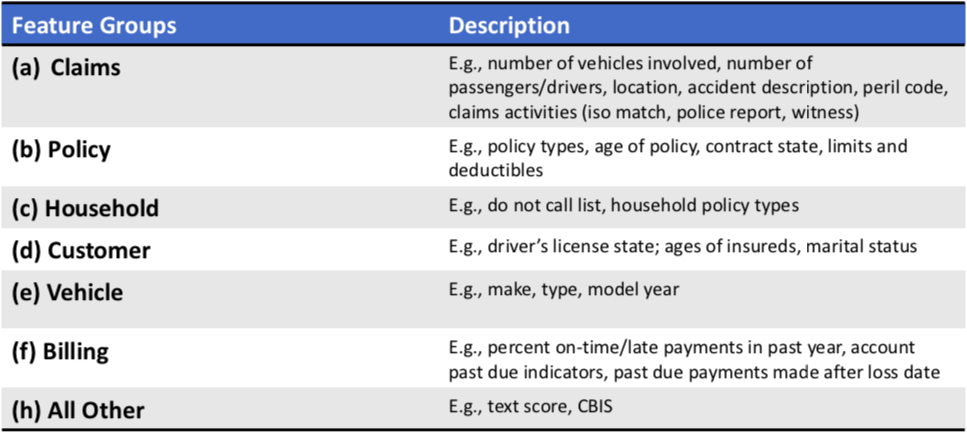}
\end{figure}

In partnership with this fraud program, the American Family data scientist has provided us with a large-scale data set. It contains around 4.5 million cases from 2012-2015. The data scientist split the data set into a training set, validation set, and testing set at a ratio of $50\%$, $20\%$, and $30\%$. There are four main indicator variables that the data scientist is interested in predicting, namely ``referral$\_$ind", ``siu$\_$ind", ``mitigation$\_$ind", and ``assignment$\_$ind". These will be expanded upon in the next paragraph when discussing the claims process. The data is sourced and integrated together from a variety of sources. Internal data sets provide information on the customer and the customer's claim, policy, household, vehicle, and billing preferences. This is expanded on in Figure \ref{SummaryFeatures}. External data sets are sourced from the National Insurance Crime Bureau and Insurance Services Office, and information on customer banking transaction frauds and their social media are obtained from companies such as CoreLogic and Networked Insights. There are around 255 possible independent variables, which present a complete and comprehensive view of the customer and their claim, to be used in prediction of the four indicators mentioned earlier.

\begin{figure}[h]
\caption{Workflow of the Special Investigation Unit (SIU).}
\label{Workflow}
\centering
\includegraphics[scale=0.5]{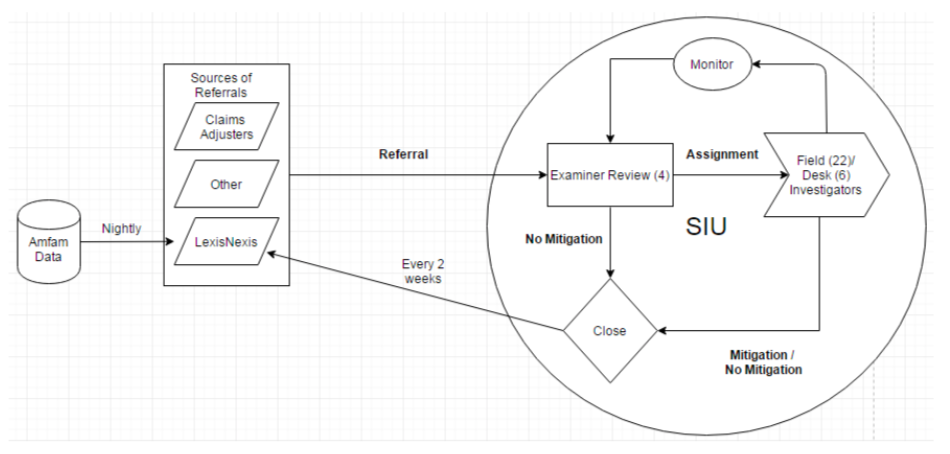}
\end{figure}

The process of reviewing claims goes through the Special Investigation Unit (SIU). Data from American Family is given to three sources of referrals. The two main ones are  Claim Adjusters within American Family and LexisNexis, an external corporation that provides risk management services. If these sources believe that a claim could be fraudulent, they refer the claim to SIU (referral$\_$ind). Once referred to SIU, the referred claim is given to an examiner to review. Once reviewed, if the claim is deemed to be necessary for investigation, the examiner assigns the claim to either a field or desk investigator (assignment$\_$ind). If this is not the case, the claim is not mitigated (mitigation means the payment was reduced or not paid), and the case is closed. Otherwise, if the claim is received by the investigators, they decide whether to continue to monitor the case, or whether to mitigate or not mitigate the case. The ``siu$\_$ind" indicator variable indicates if the claim was mitigated (payment reduced or not paid) or the claim was questionable or submitted to National Insurance Crime Bureau (NICB), while the ``mitigation$\_$ind" indicator variable indicates if the claim was just mitigated or not. This process is presented visually in Figure \ref{Workflow}. 

Currently, all of these steps are done by hand, without any sort of implemented statistical help. This of course lends itself to fraudulent claims slipping through the cracks, and in a much smaller subset of cases, possible oversight from the examiners or investigators. For this project, we hope to increase the accuracy of prediction on the ``siu$\_$ind" indicator variable by a significant amount using tuned Random Forest and Neural Network methods, and will compare the accuracy of the methods using Area under ROC Curve and Type II error rate.

\subsection{Preliminary Analysis}

Throughout the rest of the analysis, we will be using the "h2o" package in $R$. In general, H2O is an open source platform that allows businesses to deploy AI and deep learning to solve complex problems. With in-memory compression, it is capable of handling billions of rows of data in-memory using just a small cluster. It includes many common machine learning algorithms, such as generalized linear models, principal components analysis, deep learning, and Random Forest. With seamless integration into $R$ through the "h2o" package, it allows for a much easier way of performing analysis on the large data sets provided than standard $R$ packages such as "randomForest".

\begin{figure}[h]
\caption{Proportion of Missing Data in Training Set.}
\label{MissingData}
\centering
\includegraphics[scale=0.5]{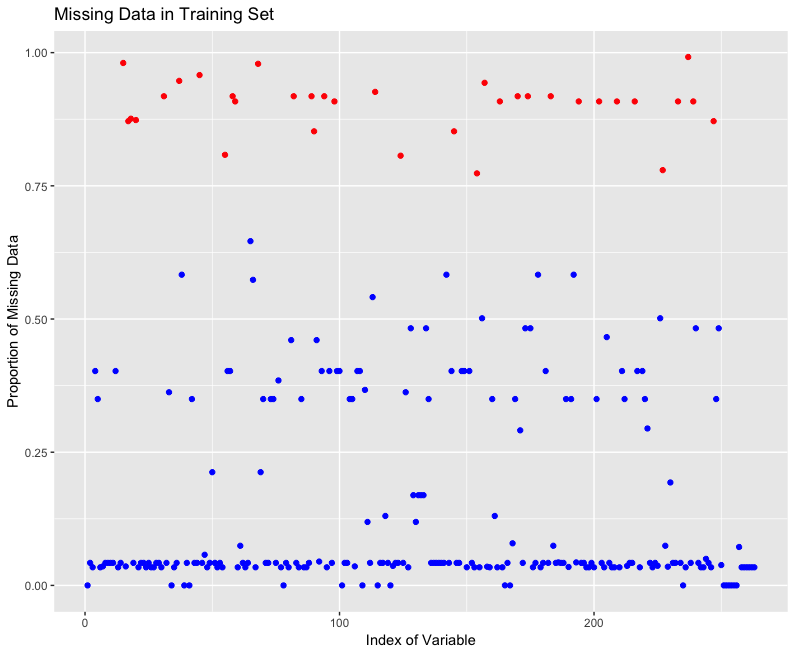}
\end{figure}

Performing a cursory look through the training data, we first check to see how much missing data there is in the training set. We provide these plots in Figure \ref{MissingData}. From the graph, we see that about $10-15\%$ of the predictors are missing more than $75\%$ of cases and $20-25\%$ of the predictors are missing between $25\%$ and $75\%$ of cases. Typically, in building a generalized linear model such as logistic regression, we would use some type of mean imputation or algorithm such as MICE to create ``substitute" values for these missing values or throw out cases with missing values. However, problems arise in either of these approaches. If we just throw out cases with missing values, we are needlessly throwing out relevant information, and many cases have at least one missing value. Adapting algorithms to handle observations with missing data is important, as rarely do we have many instances of perfect cases with all relevant predictors having observed values. On the other hand, if we use mean imputation, it could drastically affect the relationship between predictors, and cause a reduction of variance since we are essentially scaling everything towards the mean. In addition, performing multiple imputation such as MICE on large data sets is very time-intensive, and as new data is read in, we would have to perform the imputation algorithm over and over again, which could take just as long or longer than just performing the fraudulent detection by hand. While the imputation methods assume the data is missing at random, we will instead interpret the data as missing for a reason. This allows us to treat missing values as containing information. They will then be assigned their own category within each variable (treated as ``NA" category). 

\subsection{Neural Networks}

A neural network is a collection of neurons arranged in layers, consisting of the input layer, hidden layer, and output layer. The input layer consists of our predictor variables. Each predictor variable is assigned a neuron in the input layer, and represents a feature/piece of information of our data. The hidden layers lie between the input and output layers. Each hidden layer consists of a set number of nodes. Each neuron in a hidden layer receives information from the previous layer's nodes, multiplies it by some weight, and adds a bias to it. Then some pre-specified non-linear activation function is applied to that value, and the resulting value is used in the next hidden layer. The output layer brings together all the information from the last layer; in our case, it will be one neuron as we are interested in classifying whether a case is fraudulent or not (binary classification). To train for these weights and biases involved in connecting the neurons, the neural network starts with random weights and biases, and trains for optimal values by reducing the cost/loss function of the neural network using backpropogation and gradient descent. 

Neural networks are flexible since there are so many parameters one can tune for them. This may also be considered a disadvantage too, since optimizing so many parameters can be very tedious and require many computations. For us, we will stick with only a few parameters, namely mini batch size, input and hidden layer dropout ratio, epochs, max w2, learning rate, l1 regularization, and l2 regularization. Scaling of these parameters should allow us to make adjustments in order to balance how well the network fits the training data vs. how well it fits the validation and testing sets, so that we can avoid overfitting in the training set and maximize Area under ROC curve in the validation and testing sets.

\subsection{PDE Simulation of Two-Layer Network (from Section 11 of \cite{MMN2018})}

\subsubsection{Piecewise Linear Activation Function}
\label{Piecewise}

In this section, the activation $\sigma_*({\bf x};{\bm \theta}_i)=\sigma(\langle {\bf w}_i, {\bf x}\rangle )$ is used, where $\sigma(t)=s_1$ if $t\leq t_1$, $\sigma(t)=s_2$ if $t\geq t_2$, and $\sigma(t)$ is interpolated linearly for $t\in (t_1,t_2)$. In the simulations, $t_1=0.5$, $t_2=1.5$, $s_1=-2.5$, and $s_2=7.5$. 

Simulation of the PDE for general $d$ is computationally intensive. So, we only consider $d=\infty$. In the case of $d=\infty$, the risk can be given by

\begin{align}
{\bar R}_{\infty}({\bar \rho})=\frac{1}{2} \bigg( 1-\int q_+ (r){\bar \rho}(dr)\bigg)^2+\frac{1}{2}\bigg( 1+ \int q_-(r){\bar \rho}(dr)\bigg)^2,
\end{align}

where $q_\pm (t)=\mathbb{E}\{\sigma((1\pm \Delta)tG)\}$, $G\sim N(0,1)$. In addition, 

\begin{align}
\psi_\infty(r;{\bar \rho})=\frac{1}{2} [\langle q_+,{\bar \rho}\rangle-1]q_+(r)+\frac{1}{2}[\langle q_-,{\bar \rho}\rangle +1]q_-(r).
\end{align}

The PDE will then be $\partial_t {\bar \rho}_t=2\xi(t)\partial_r[{\bar \rho}_t \partial_r \psi_\infty(r;{\bar \rho}_t)]$. The solution to the PDE is approximated by the multiple-deltas anstz ${\bar \rho}_t=\frac{1}{J} \sum_{i=1}^J \delta_{r_i(t)}$ at al times $t$, where $J\in \mathbb{N}$ is a pre-chosen parameter. If we fix $J$ and the PDE is initialized at ${\bar \rho}_0$ taking the above form, then for any $t\geq 0$, ${\bar \rho}_t$ remains in the above form. Then for any smooth test function $f:\mathbb{R}\to \mathbb{R}$ with compact support, 

\begin{align}
\frac{1}{J} \sum_{i=1}^J f'(r_i(t))r_i'(t)&=\partial_t\langle f, {\bar \rho}_t\rangle=-2xi(t)\langle f',{\bar \rho}_t\partial_r\psi_\infty(r;{\bar \rho}_t)\langle \\
&= -2\xi(t) \frac{1}{J}\sum_{i=1}^J f'(r_i(t))\partial_r \psi_\infty (r_i(t);{\bar \rho}_t).
\end{align}

Under this, we write ${\bar R}_\infty({\bar \rho}_t)={\bar R}_{\infty,J}({\bf r}(t))$, where ${\bf r}(t)=(r_1(t),\dots,r_J(t))^T$, and 

\begin{align}
{\bar R}_{\infty, J}({\bf r})=\frac{1}{2} \bigg( 1-\frac{1}{J} \sum_{i=1}^J q_+(r_i)\bigg)^2+\frac{1}{2}\bigg(1+\frac{1}{J}\sum_{i=1}^J q_-(r_i)\bigg)^2.
\end{align}

Since $\partial_r\psi_\infty(r_i(t);{\bar \rho}_t)=(J/2)(\nabla {\bar R}_{\infty, J}({\bf r}(t)))_i$, we obtain $\frac{d}{dt} {\bf r}(t)=-J\xi(t)\nabla {\bar R}_{\infty, J}({\bf r}(t))$. Therefore, under the multiple-deltas ansatz, we can simulate numerically the PDE using the above evolution equation of ${\bf r}(t)$. So, given ${\bf r}(t)$, we can approximate ${\bf r}(t+\delta t)$ for some small displacement $\delta t$ by

\begin{align}
{\bf r}(t+\delta t) \approx {\bf r}(t)-J\xi(t) \nabla {\bar R}_{\infty,J}({\bf r}(t))\delta t.
\end{align}

\subsubsection{ReLU Activation}

The activation $\sigma_* ({\bf x}; {\bm \theta})=a\max(\langle {\bf w},{\bf x} \rangle + b,0)$, with ${\bm \theta}=({\bf w}, a,b) \in \mathbb{R}^{d+2}$. 

Consider $s_0=\gamma d$ for some $\gamma \in (0,1)$. For simplicity, the limit $d\to \infty$ is considered. For ${\bm \theta}\sim \rho$, let ${\bar \rho}$ be the joint distribution of the four parameters ${\bf r}=(a,b,r_1=\|{\bf w}_{1:s_0}\|_2,r_2=\|{\bf w}_{(s_0+1):d} \|_2)$, where ${\bf w}_{i:j}=(w_i,\dots,w_j)^T$. In the limit $d\to \infty$, the risk $R(\rho)={\bar R}_\infty({\bar \rho})$, where

\begin{align}
{\bar R}_\infty ({\bar \rho})=\frac{1}{2} \bigg(1-\int a q_+ (r_1,r_2,b){\bar \rho}(d{\bf r})\bigg)^2+\frac{1}{2}\bigg(1+\int a q_-(r_1,r_2,b){\bar \rho}(d{\bf r})\bigg)^2, \\
q_\pm (r_1,r_2,b)=b\Phi\bigg(\frac{b}{\sqrt{(1\pm \Delta)^2 r_1^2+r_2^2}}\bigg)+\sqrt{(1\pm \Delta)^2 r_1^2+r_2^2}\phi\bigg(\frac{b}{\sqrt{(1\pm \Delta)^2r_1^2+r_2^2}}\bigg),
\end{align}

with $\phi(x)=\exp(-x^2/2)/\sqrt{2\pi}$ and $\Phi(x)=\int_{-\infty}^x \phi(t) dt$. Assuming that the solution to the PDE can be approximated at all times $t$ by the multiple-deltas ansatz ${\bar \rho}_t=\frac{1}{J} \sum_{i=1}^J \delta_{{\bf r}_i(t)}$, where $J\in \mathbb{N}$ is a pre-chosen parameter, and ${\bf r}_i(t)=(a_i(t), b_i(t), r_{1,i}(t),r_{2,i}(t))$. We are then left with the evolution equation

\begin{align}
\frac{d}{dt} {\bf r}_i(t)=-J \xi(t) \nabla_i {\bar R}_{\infty, J} ({\bf r}_1(t),\dots,{\bf r}_J(t)),
\end{align}

for $i=1,\dots,J$, where ${\bar R}_{\infty,J} ({\bf r}_1(t),\dots,{\bf r}_J(t))={\bar R}_\infty({\bar \rho}_t)$ under the ansatz, and $\nabla_i$ denotes the gradient of ${\bar R}_{\infty,J}({\bar r}_1,\dots,{\bar r}_J)$ w.r.t. ${\bf r}_i$. More explicitly, 

\begin{align}
{\bar R}_{\infty,J} ({\bf r}_1,\dots,{\bf r}_J)=\frac{1}{2} \bigg(1-\frac{1}{J} \sum_{i=1}^J a_i q_+ (r_{1,i},r_{2,i},b_i)\bigg)^2+\frac{1}{2}\bigg(1+\frac{1}{J} \sum_{i=1}^J a_i q_0(r_{1,i},r_{2,i},b_i)\bigg)^2.
\end{align}

Given ${\bf r}_i(t)$, we can approximate ${\bf r}_i(t+\delta t)$ for some small displacement of $\delta t$ by

\begin{align}
{\bf r}_i (t+\delta t)\approx {\bf r}_i(t)-J \xi (t) \nabla_i {\bar R}_{\infty,J}({\bf r}_1,\dots,{\bf r}_J)\delta t.
\end{align}

\subsection{Distribution of Weights}
\subsubsection{AmFam}

\begin{figure}[h]
\caption{Evolution of four parameters from ReLU activation for SGD and PDE algorithms varying $\Delta$. The x-axis represents the iteration, where 1-9 represents iterations 1-9, 10-19 represents iterations (10,20,30,40,50,60,70,80,90,100), 20-28 represents iterations (200, 300, 400, 500, 600, 700,800,900,1000), and so on.}
\label{SGDvsPDEReLU}
\centering
\includegraphics[scale=0.4]{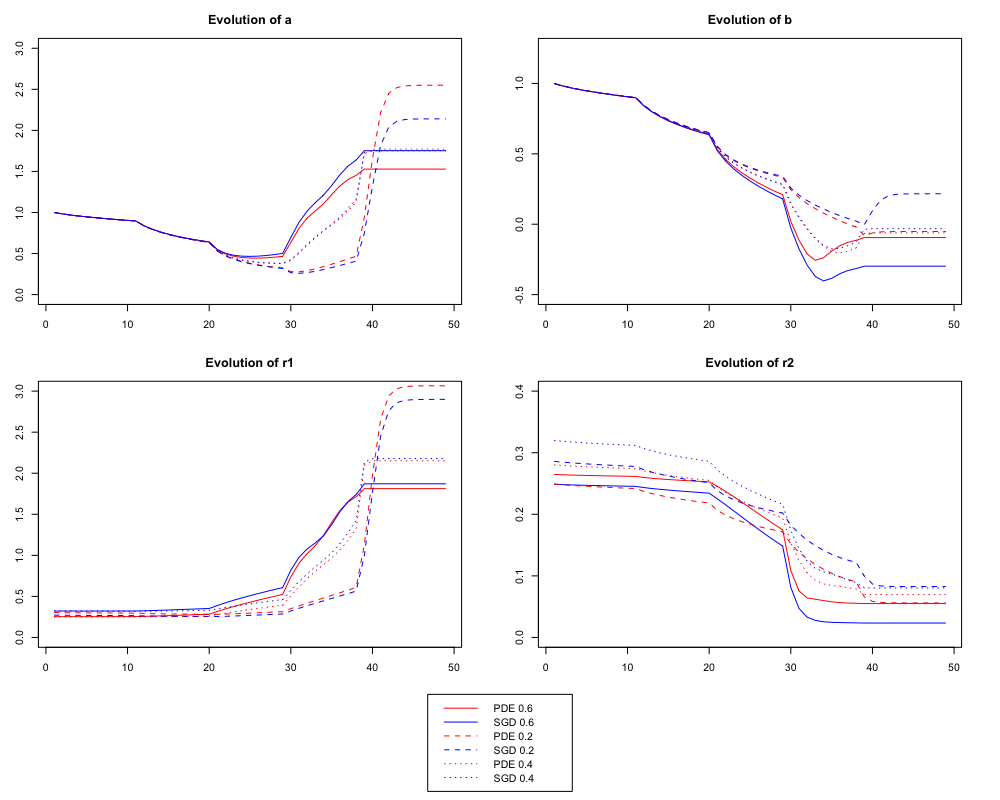}
\end{figure}

We first look at the closeness of the SGD and PDE under ReLU for the AmFam data to see how well the PDE simulation does for real world data.  First, we set $d=252$, the number of columns in the training set. We then set $s=120$, $N=200$, $\epsilon=2*10^{-4}$, $\xi(t)=t^{-1/4}$, $J=100$, and each progressive step/iteration of the algorithm to be $1*10^{-3}$. 

\begin{figure}[h]
\caption{Second layer weight distribution observed on two-layer network (250,250) ran on AmFam Data, and weight distributions derived from PDE simulation using Piecewise Linear Activation Function for various choices of weights from second layer.}
\label{AmFam2Layer}
\centering
\includegraphics[scale=0.5]{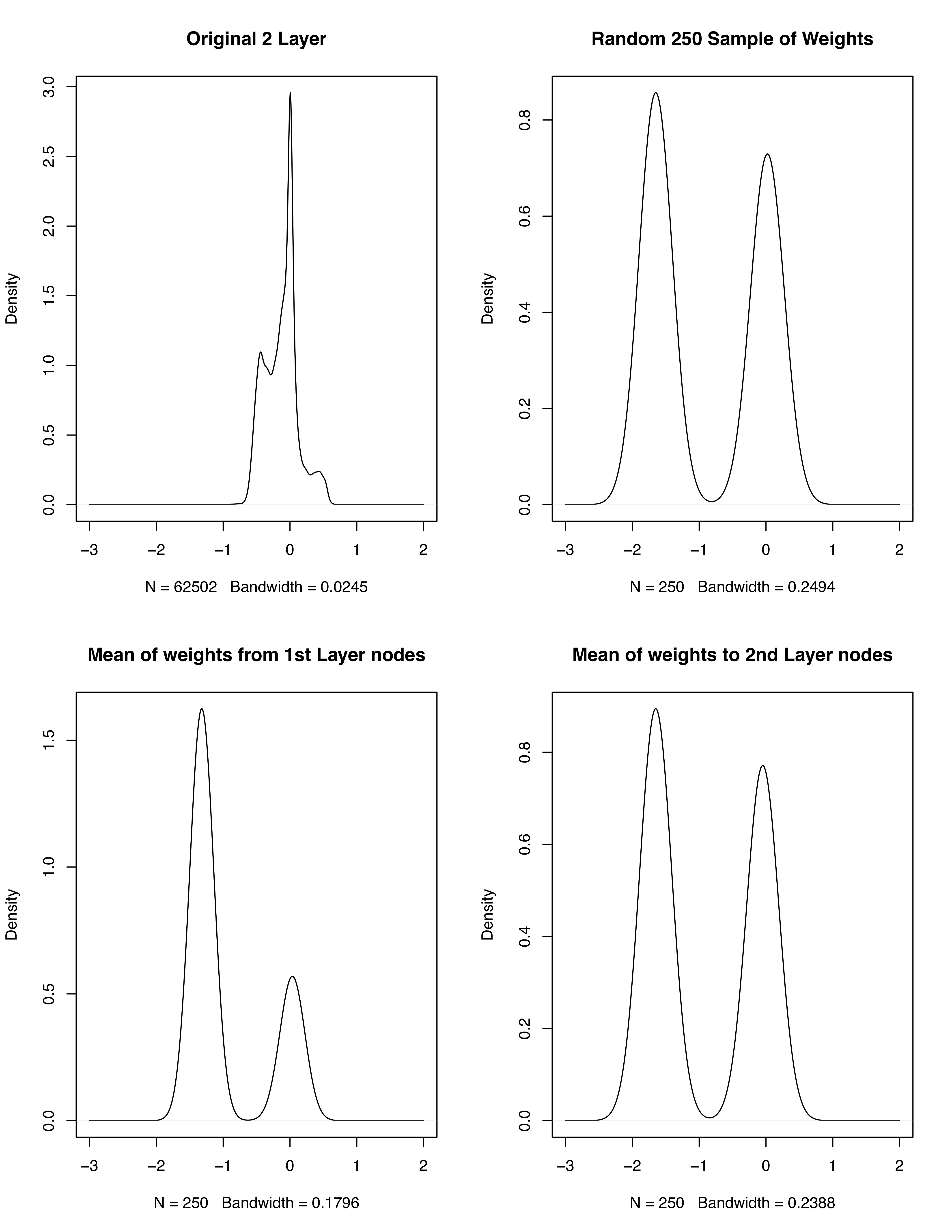}
\end{figure}

\begin{figure}[h!]
\caption{Bimodal Nature observed in various layers of three-layer networks run on AmFam Data.}
\label{Bimodal3Layer}
\centering
\includegraphics[scale=0.5]{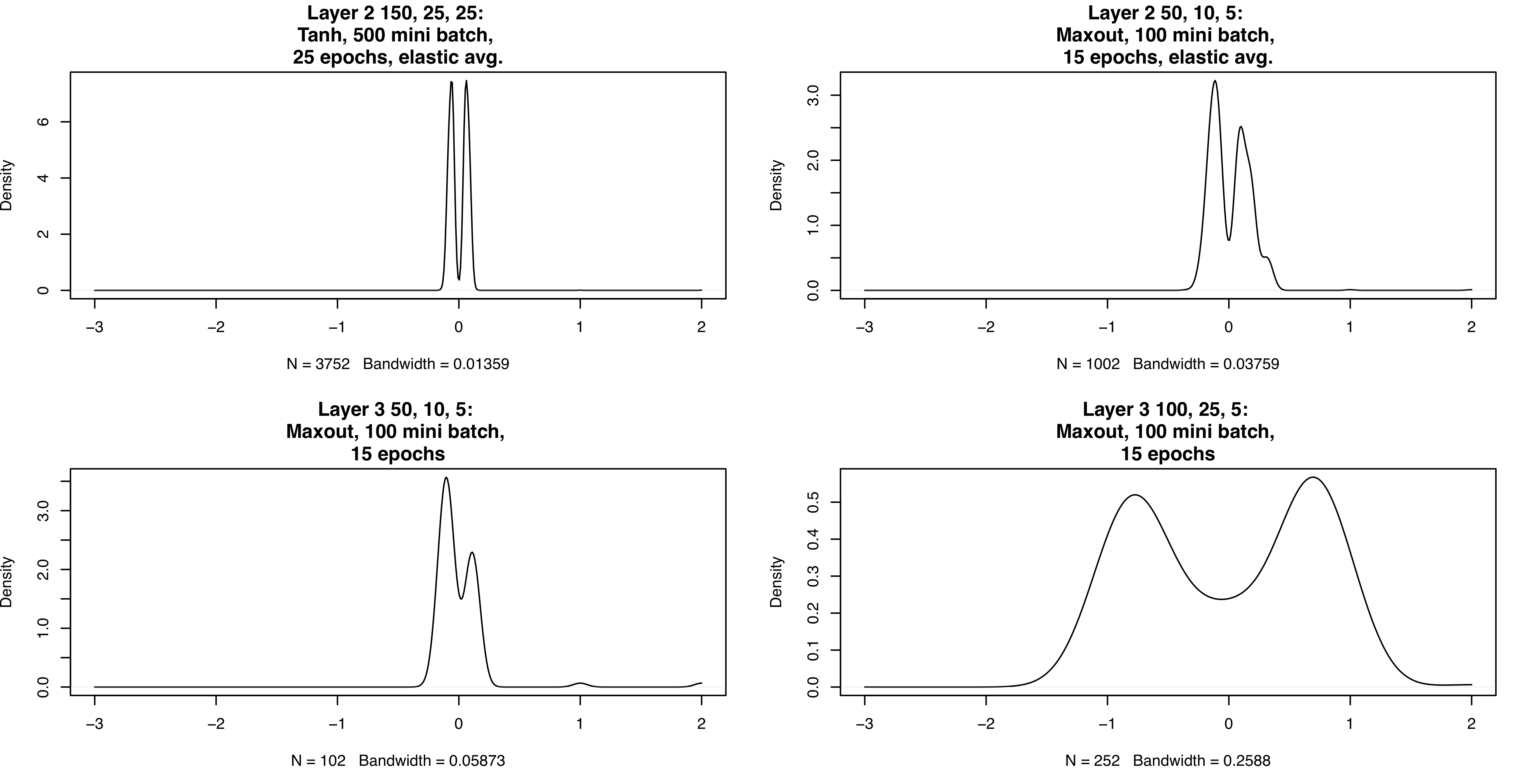}
\caption{Bimodal Nature observed in various layers of four-layer networks run on AmFam Data.}
\label{Bimodal4Layer}
%\centering
\includegraphics[scale=0.5]{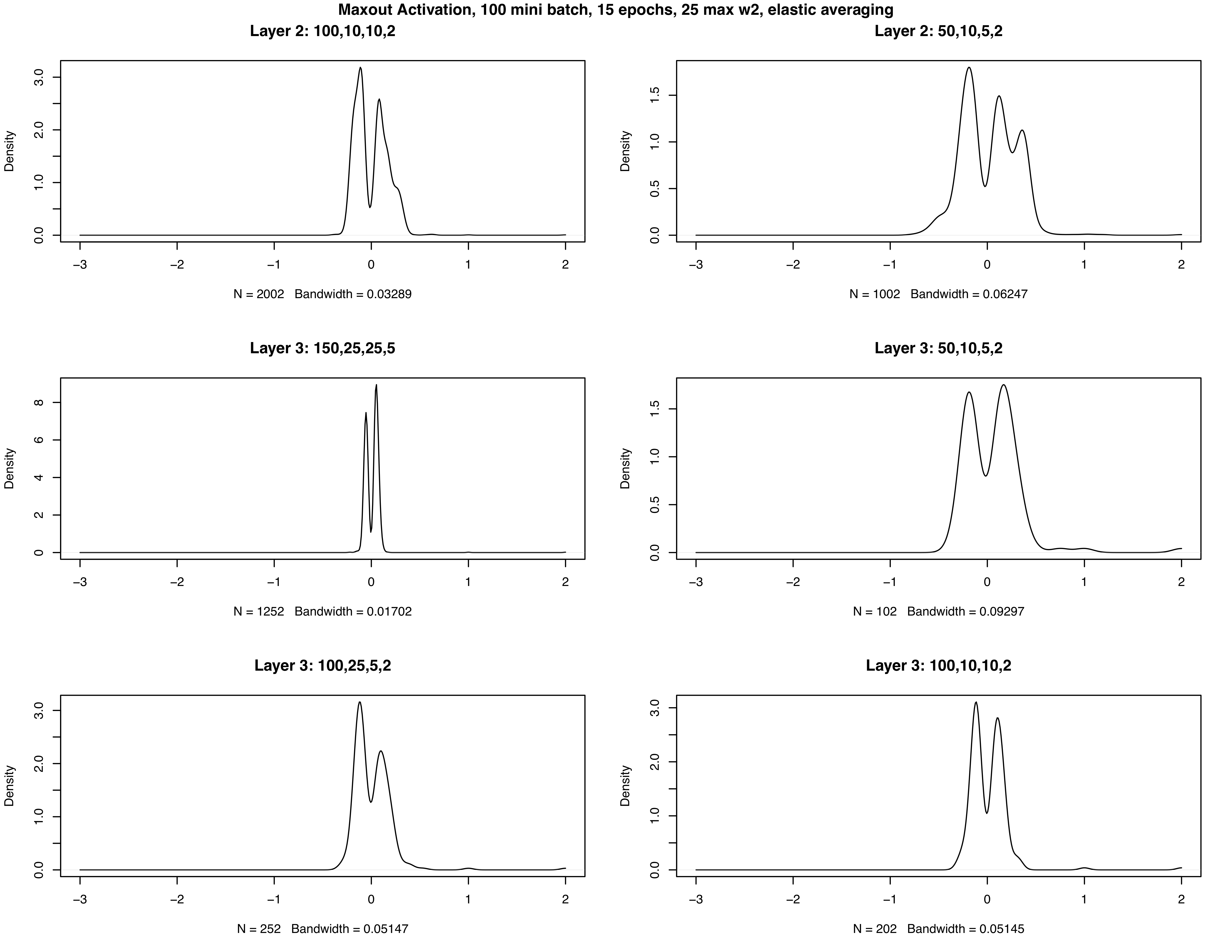}
\end{figure}

We then create an association matrix $\Sigma$ of all the variables in the train set by taking every 500th data point in the train set and applying the function mixed$\_$assoc, which allows us to calculate a pairwise association between all variables in a data-frame. We then subset $\Sigma$ as $\Sigma_1=\Sigma_{1:s,1:s}$ and $\Sigma_2=\Sigma_{(s+1):d,(s+1):d}$, i.e. the association matrices of the first $s$ variables and last $d-s$ variables.

We use the make.positive.definite function to convert each of the $\Sigma_i$ into a positive definite matrix $\Sigma_i^+$, and then use the cor2cov function to convert $\Sigma_i^+$ into a covariance matrix ${\bm \Sigma}_i^+$. We then initialize $r_{j,i}(0)=\|Z_{j,i}\|_2$, where $(Z_{1,i})_{i\leq N} \sim_{i.i.d.} N(0, {\bm \Sigma}_1^+)$ and $(Z_{2,i})_{i\leq N} \sim_{i.i.d.} N(0, {\bm \Sigma}_2^+)$ indepedently, along with $a_i(0)=1$, $b_i(0)=1$. We sequentially compute ${\bf r}_i(t)$ for $5*10^5$ iterations.

In Figure \ref{SGDvsPDEReLU}, for evolution of $a$, we compute $\frac{1}{N} \sum_{i=1}^N a_i$ for the SGD and $\frac{1}{J} \sum_{i=1}^J a_i(t)$ for the PDE. We do similar calculates for $b$, $r_1$, and $r_2$. We also vary $\Delta$ to see how the parameters change. For the most part, it appears that there is a good match in the evolutions of the parameters between the SDG and PDE across varying $\Delta$.

We next look at the weight distributions produced from the PDE. We first fit a simple neural network with two hidden layers, each layer containing 250 nodes to the AmFam data. We focus our attention on the weights from the second layer. In Figure \ref{AmFam2Layer}, we see the top left figure is the density of the weights produced from the 2nd layer. The top right is sampling 250 of the weights produced from the second layer and running them through the PDE simulation using the Piecewise Linear Activation Function. The bottom two are looking at the means of weights from the 250 nodes (either averaging across weights coming from the same node or going to the same node) and running them through the PDE simulation using the Piecewise Linear Activation Function. It looks like while in the original 2 layer density it is centered around 0 and seems relatively unimodal, in the other 3 density plots, there is a tendency towards a bimodal density distribution.

We also found that the bimodalness arises in weight distributions of different layers in various neural networks run on the AmFam data, as shown in Figures \ref{Bimodal3Layer} and \ref{Bimodal4Layer}.

\subsubsection{Theoretical}

In addition to seeing how the PDE affects the AmFam data, we also try to create a ``true" theoretical distribution through simulation. To do this, we will use the interpolation activation function with $s_1=-2.5$, $s_2=7.5$, $t_1=0.5$, $t_2=1.5$, $J=100$, $N=2000$, $\Delta=0.8$, and $d=250$, with an increment size of $1*10^{-5}$ and $\xi(t)=1$. 

We create a target vector $y$ of size $N$, where $y_i=1$ with probability $1/2$ and $0$ with probability $1/2$. Next, we create our covariance matrix, ${\bm \Sigma}$ of size $d$ by $d$, where the diagonal entries are $(1+\Delta)^2$ and the off diagonals are $0.001$. We then sample $x\sim N(0,{\bf \Sigma})$. We next create ${\bf \Sigma}_2$ of size $d$ by $d$, where the diagonal entries are $(\Delta^2/d)$, and the off diagonals are $0.001$. We then initialize $r_i(0)=\|Z_i\|_2$, where $(Z_i)_{i\leq J} \sim_{i.i.d.} N(0, {\bf \Sigma}_2)$. We sequentially compute the value of ${\bf r}(t)$ using ${\bf r}(t+\delta t) \approx {\bf r}(t)-J\xi(t) \nabla {\bar R}_{\infty,J}({\bf r}(t))\delta t$ from Section \ref{Piecewise}.

Once we have run ${\bf r}(t)$ through the PDE approximation, we then set them as the initial weights in a standard neural network in $R$ using the neuralnet package, with a single hidden layer of 100 nodes and then two hidden layers, each with 100 nodes. Using $y$ and $x$ initialized from above, we run the two neural networks and extract the weights from the output layer. 

\begin{figure}[h]
\caption{Evolution of weight distribution under PDE approximation with ${\bf r}(0)$ taken from extracted weights of single hidden layer neural network with 100 nodes. }
\label{PDEEvolution1Layer}
\centering
\includegraphics[scale=0.4]{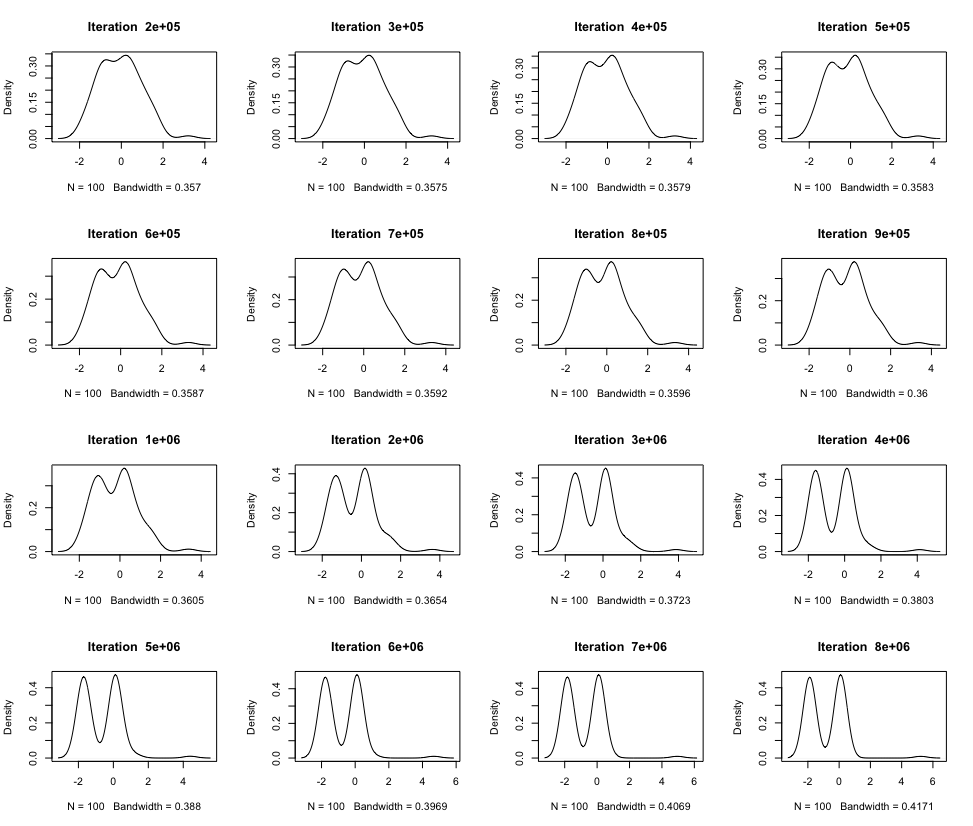}
\end{figure}

\begin{figure}[h]
\caption{Weight Distributions after running PDE approximation on extracted weights from neural networks with one hidden layer of 100 nodes and two hidden layers of 100 nodes each, respectively.}
\label{TwoNetworksPDE}
\centering
\includegraphics[scale=0.4]{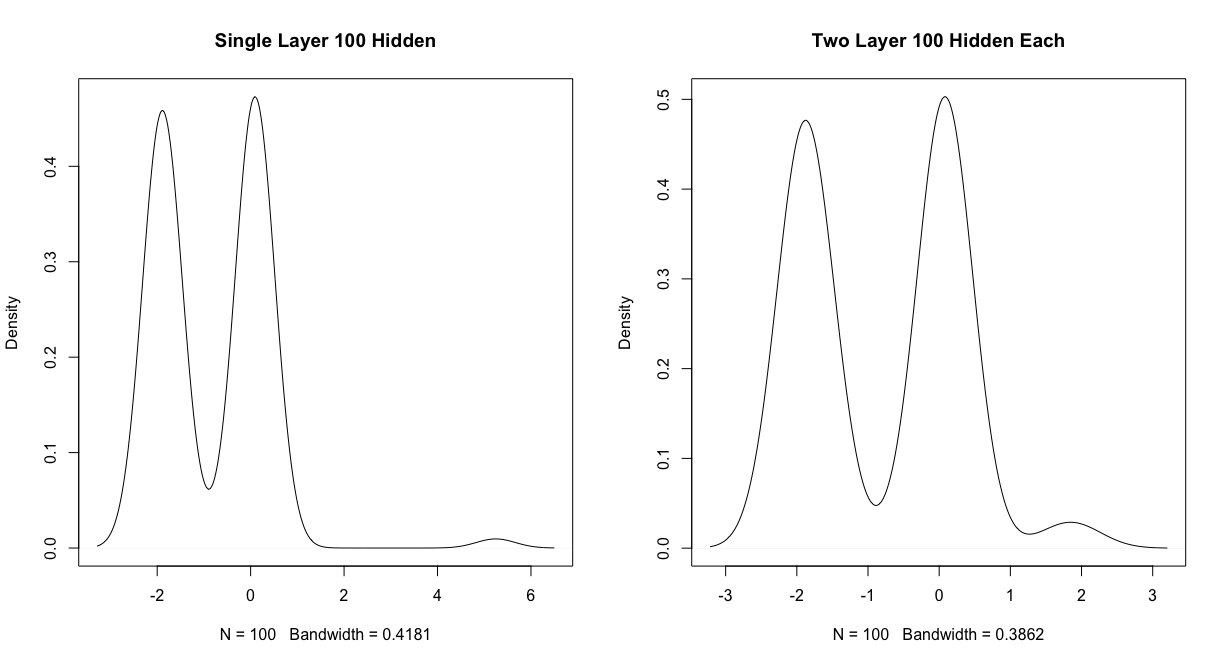}
\end{figure}

We then set these extracted weights as ${\bf r}(0)$ and run them through the PDE approximation. We observe the transformation of ${\bf r}(t)$ in Figure \ref{PDEEvolution1Layer} and the weight distributions produced from the single hidden layer neural network and two hidden layer neural network after running the extracted weights through the PDE approximation in Figure \ref{TwoNetworksPDE}. 

\subsection{Analysis of Distribution of Weights}

To compare the distributions produced from neural networks on the AmFam data to the theoretical distributions produced from the PDE approximation, we utilize two classical distribution distance metrics, the Kullback-Leibler divergence (KL) and L1 distance. The PDE approximation provides theoretical mathematical models to analyze the bimodality phenomenon, and we want to see which fit neural networks get closest to/achieve that bimodal phenonemon. Iterating through each of the neural networks trained on the AmFam data, we compute the KL and L1 distance measures using the ``true" theoreticals produced from the PDE approximation as the base distribution (we will refer to distribution A as the true theoretical that utilized a single hidden layer of 100 nodes, and distribution B as the other).  Once the KL distances are computed for the two distributions, we will rank them from lowest to highest with respect to distribution A and distribution B, and then average their rankings. This averaged rank is stored in the variable kl.avg, and the same is done for the L1 distances, their average being stored in the variable l1.avg.

\begin{figure}[h!]
\caption{Variance explained from MCA.}
\label{VarianceExplainedMCA}
\centering
\includegraphics[scale=0.45]{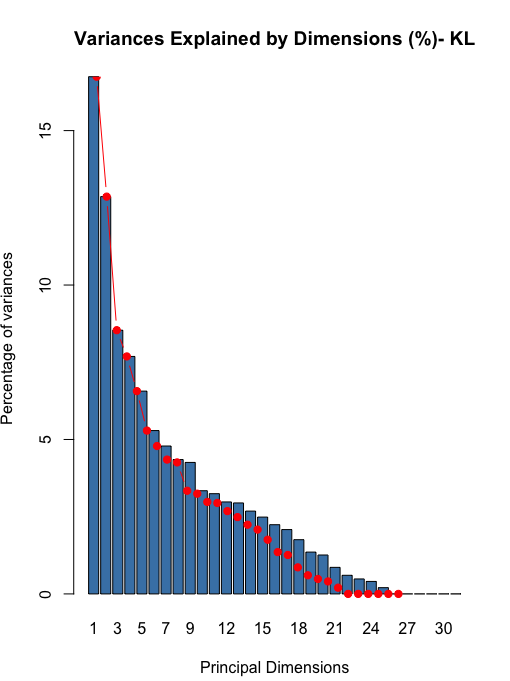}
\end{figure}

\begin{figure}[hbtp]
\caption{MCA Analysis plotted with KL divergence variable.}
\label{MCAAnalysisKL}
\centering
\includegraphics[scale=0.4]{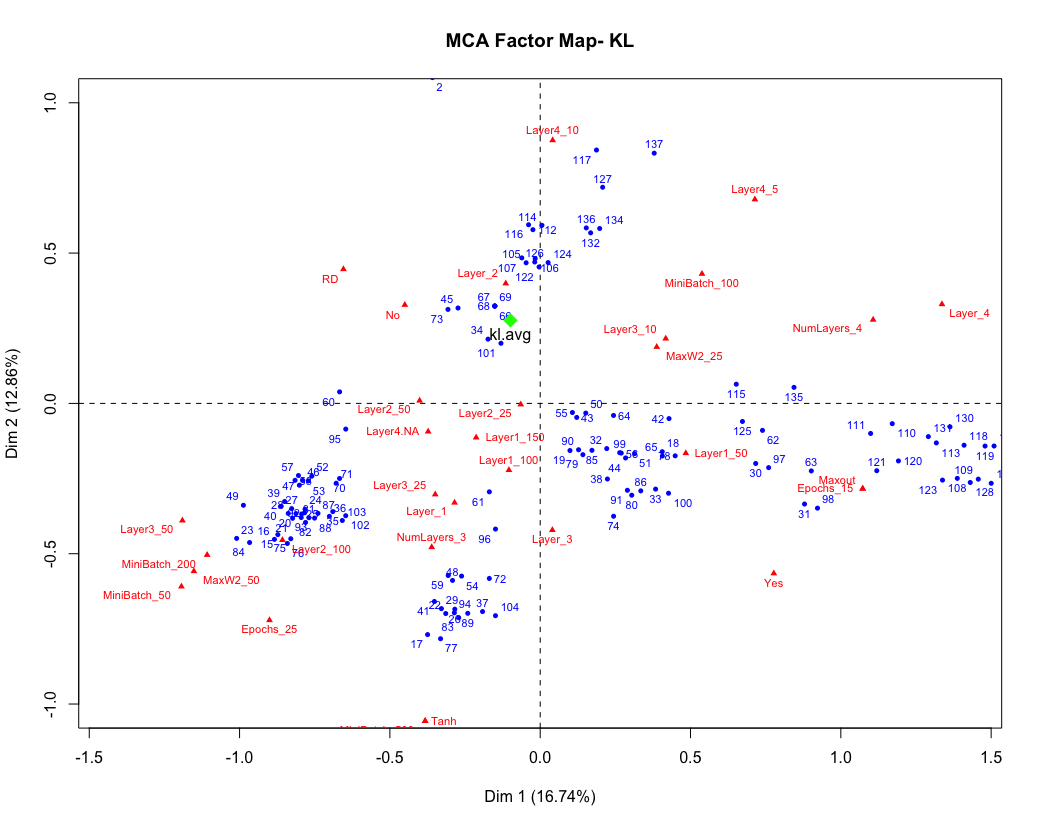}
\caption{MCA Analysis plotted with L1 distance variable.}
\label{MCAAnalysisL1}
%\centering
\includegraphics[scale=0.4]{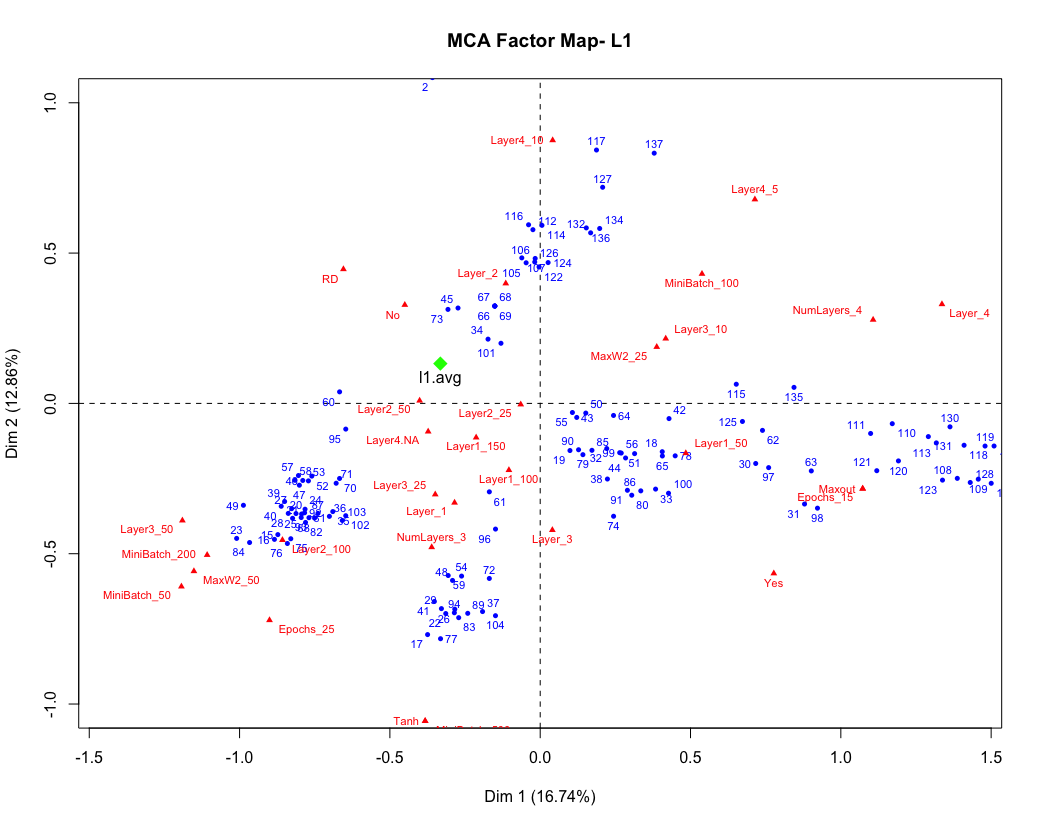}
\end{figure}

Multiple Correspondence Analysis (MCA) is a generalization of principal component analysis when the variables to be analyzed are mainly categorical instead of quantitative. It allows us to identify which attributes of the neural network contribute the most to reducing the distance metrics. In Figures \ref{MCAAnalysisKL} and \ref{MCAAnalysisL1}, we produce plots from the MCA based on a reduction to the first two dimensions/components. On these plots, we also graph the KL divergence and L1 distance variables based on their values of the first two dimensions/components produced from the MCA. Looking at the factor maps, the red dots represent variables, while the blue dots represent the individual neural networks. By plotting the kl.avg and l1.avg variables on the factor map, we can roughly see which neural networks and attributes achieve the lower values of those variables. 

Using a weighted average based on dimension contribution from the first five dimensions (seen in Figure \ref{VarianceExplainedMCA}), we find that for both of the KL and L1 distance variables, the MCA suggests that a three layer network with Layer 1 around 100-150 nodes, Layer 2 around 25-50 nodes, and Layer 3 around 25 nodes with the Rectifier activation, mini batch of 100, and max w2 of 25 seems to do best, with the second layer having the distribution most similar to distribution A and distribution B.

\end{document}